\documentclass[12pt]{article}
\usepackage{newtxtext}

\usepackage[tmargin=1.5in,bmargin=1in,lmargin=1.5in,rmargin=1.25in]{geometry}
\usepackage[ruled,vlined]{algorithm2e}
\usepackage{amsmath,amssymb}

\usepackage{enumerate}
\usepackage{graphicx}
\usepackage{caption}
\usepackage{subcaption}
\graphicspath{{./figures/}}

\usepackage[colorinlistoftodos,textsize=footnotesize]{todonotes}

\newcommand{\rr}[1]{#1}
\newcommand{\rb}[1]{#1}
\newcommand{\new}[1]{#1}
\newcommand{\rR}[1]{#1}

\usepackage{booktabs}
\usepackage{threeparttable}

\usepackage{lineno}
\modulolinenumbers[1]

\usepackage{hyperref}
\usepackage{listings}
\AtBeginDocument{\addtocontents{toc}{\protect\setlength{\parskip}{0pt}}}

\usepackage{titlesec}

\titleformat*{\section}{\large\bfseries\sffamily}
\titleformat*{\subsection}{\normalsize\bfseries\sffamily}
\titleformat*{\subsubsection}{\normalsize\bfseries\sffamily}

\providecommand{\keywords}[1]{\textbf{\textsf{Keywords:}} #1}

\usepackage{etoolbox}

\makeatletter
\patchcmd{\@maketitle}{\LARGE}{\bfseries\sffamily\large}{}{}
\makeatother

\title{SPINN: Sparse, Physics-based, and \rR{partially} Interpretable Neural Networks for PDEs\footnote{Author names listed alphabetically. Both authors contributed equally to the work.}}
\author{\textsf{Amuthan A. Ramabathiran}$^{1,2}$\footnote{Email address: \texttt{amuthan@aero.iitb.ac.in}} \and \textsf{Prabhu Ramachandran}$^{1,2}$\footnote{Email address: \texttt{prabhu@aero.iitb.ac.in}}}
\date{%
	$^1${\small Department of Aerospace Engineering, Indian Institute of Technology Bombay, Mumbai - 400076, Maharashtra, India.}\\[2ex]%
	$^2${\small Center for Machine Intelligence and Data Science (CMINDS), Indian Institute of Technology Bombay, Mumbai - 400076, Maharashtra, India.}\\[2ex]%
	\today
}

\begin{document}
\maketitle

\begin{quote}
\section*{Abstract}
We introduce a class of Sparse, Physics-based, and \rR{partially} Interpretable Neural Networks (SPINN) for solving ordinary and partial differential equations \new{(PDEs)}. By reinterpreting a traditional meshless representation of solutions of PDEs we develop a class of sparse neural network architectures that are \rR{partially} interpretable. The SPINN model we propose here serves as a seamless bridge between two extreme modeling tools for PDEs, \new{namely} dense neural network based methods \new{like Physics Informed Neural Networks (PINNs)} and traditional mesh-free numerical methods, thereby providing a novel means to develop a new class of hybrid algorithms that build on the best of both these viewpoints. A unique feature of the SPINN model that distinguishes it from other neural network based approximations proposed earlier is that it is (i) interpretable\rR{, in a particular sense made precise in the work}, and (ii) sparse in the sense that it has much \new{fewer connections than typical dense neural networks used for PDEs.} Further, the SPINN algorithm implicitly encodes mesh adaptivity and is able to handle discontinuities in the solutions. In addition, we demonstrate that Fourier series representations can \new{also} be expressed as a special class of SPINN and propose generalized neural network analogues of Fourier representations. We illustrate the utility of the proposed method with a variety of examples involving ordinary differential equations, elliptic, parabolic, hyperbolic and nonlinear partial differential equations, and an example in fluid dynamics.
\end{quote}

\keywords{Physics-based Neural Networks, Sparse Neural Networks, Interpretable Machine Learning, Partial Differential Equations, Meshless methods, Numerical Methods for PDEs}


\section{Introduction}
There has been a flurry of activity in the recent past on the application of machine learning algorithms to solve Partial Differential Equations (PDEs). Unlike traditional methods like the finite element, finite volume, finite difference, and mesh-free methods, Deep Neural Network (DNN) based methods like Physics Informed Neural Networks (PINN) \cite{RPK2019} and the Deep-Ritz method \cite{EYu2018} circumvent the need for traditional mesh-based representations and instead use a DNN to approximate solutions to PDEs. The idea of using DNNs to solve PDEs is not new \cite{LLF97, LLP2000}, but their usage has exploded in the recent past. A non-exhaustive list of other approaches to apply deep learning techniques to solving PDEs include \cite{BN2018, SiKo2018, HJE2018, LLMXD2018, SAGNGHZR2020, WZ2020, LCX2020, CCLL2020, WXZZ2020pre, DS2020, lu2021deepxde, LTPGC2021}. A drawback with such DNN based techniques, apart from their marked inefficiency in comparison with traditional mesh based methods for lower dimensional PDEs, is the fact that they are difficult to interpret and involve many arbitrary choices related to the network architecture.

In this work, we propose a new class of Sparse, Physics-based, and \rR{partially} Interpretable Neural Network (SPINN) architectures to solve PDEs that are both interpretable\rR{, in a sense to be made precise later,} and efficient. DNNs have been studied in the context of meshless methods in works such as \cite{HHM2020, WZ2020}, and connections between DNNs and meshless methods have been noted in works like \cite{EMW20, CGPPT20}. The key idea behind SPINN, which distinguishes it from other works cited above, is the observation that certain meshless approximations can be directly transcribed into a sparse DNN. We demonstrate that such a simple re-expression of meshless approximations as sparse DNNs allows us to bridge the emerging field of scientific machine learning and the well established methods of traditional scientific computing.

To set the stage for introducing SPINN, we note that a connection between ReLU DNNs and piecewise linear finite element approximations was proved in \cite{HLXZ2020}. This shows that basis functions with compact support can be represented as a DNN. We generalize this to represent kernel functions in meshless methods as DNNs. We use this to construct the SPINN architecture, which is a new and \rR{partially} interpretable DNN. This is significant in light of the notorious interpretability problem that attends the use of DNNs. We point out that once the DNN is replaced by a sparse equivalent network based on the aforementioned reinterpretation of meshless methods, we use an approach similar to PINN in modeling the boundary conditions and the loss function directly using the PDE. A further novelty of our method is that it naturally suggests neural network analogues of commonly used transforms such as the Fourier and wavelet transforms. We illustrate how Fourier decomposition can be accomplished using special sparse architectures in one dimension, and suggest natural neural network generalizations that go beyond traditional transformations.

\rr{The interpretation of deep neural networks in general, and PINN in particular, as a basis decomposition is not new. For instance, the authors in \cite{CGPPT20} propose new training algorithms for deep neural networks based on this adaptive basis interpretation, and show applications to PINN too. In \cite{FO19}, the authors use this interpretation to develop greedy algorithms for function approximation using neural networks. We remark that while the interpretation of DNNs via basis functions is not new, what distinguishes SPINN is that fact that we take advantage of this interpretation to provide a new class of sparse network architectures that generalize traditional meshless methods along the lines of deep neural networks. Further, the basis functions that we employ in SPINN are shifted and scaled versions of a single \emph{parent} basis function, unlike a collection of distinct basis functions that one would get from using DNNs. We clarify this issue further in the discussion on interpretability.}

\rr{We further distinguish SPINN from other methods in the literature like \cite{WANG2021113938, YRK20, LPYWVJ21} that first extract system specific features from the inputs, use these as inputs to a DNN, and finally extract the desired solution from the outputs of this DNN. In contrast, the primary point of departure of the method we propose is that the inputs are retained as such, but domain specific fetures are abstracted into kernels in a manner that the overall network architecture is much sparser than fully connected DNNs. We present many examples to illustrate this in the sequel.}

One of the primary advantages of SPINN is that it serves as a natural bridge between traditional meshless methods and methods that use DNNs, including both of these as special cases. This permits us to develop solution methodologies for PDEs that are based on neural networks and yet are \rb{potentially} computationally competitive with traditional methods. Further, the fact that our proposed method is \rR{partially} interpretable opens up new roads for traditional computational methods to be enhanced using insights from DNNs, and vice versa. For instance, we also present in this work a new class of implicit finite difference and neural network based solution schemes for time dependent partial differential equations. We thus point out the unique role of SPINN in unifying both traditional and modern numerical methods, while at the same time generalizing them. Further, the fact that SPINN generalizes meshless methods using DNNs facilitates differentiable programming, which is difficult to implement with traditional methods.

The rest of the paper is structured as follows: we begin with an introduction of the SPINN architecture by highlighting the exact relation between certain meshless representations and DNNs. We also show how Fourier representations of functions can be handled in the same framework. We then present a variety of examples involving ordinary and partial differential equations to illustrate the method. We conclude with a discussion of the key ideas presented in this work, along with directions for future investigations. Additional details about some of the simulations are presented in the appendices. The code developed to implement SPINN along with an automation script that reproduces every result reported here can be found at \url{https://github.com/nn4pde/SPINN}.

\section{Methodology}
We present here details of the SPINN architecture in a systematic fashion starting with an illustration of how certain meshless methods using radial basis functions can be exactly represented as a deep neural network. We subsequently use this to develop the general SPINN framework that goes beyond traditional meshless methods, and highlight the precise sense in which SPINN is interpretable. We then provide various details about the way the SPINN architecture is used in solving PDEs. We also propose a generalization of SPINN based on Fourier decomposition.

\subsection{Meshless approximation using radial basis functions}
To set the stage for the introduction of Sparse Physics-based \rR{and partially} Interpretable Neural Networks (SPINNs), we focus on the problem of finding a solution $u:\Omega \subset \mathbb{R}^d \to \mathbb{R}$ ($d \ge 1$) of the partial differential equation $\mathcal{N}(x, u(x), \nabla u(x), \ldots) = 0$, $x \in \Omega$, with specified Dirichlet and Neumann boundary conditions on the domain boundary $\partial \Omega$. Among the many numerical techniques that have been developed to solve such equations we focus in particular on a class of meshless methods that approximate the solution of the differential equation in terms of Radial Basis Functions (RBFs): for every $x = (x^1, \ldots x^d) \in \Omega$,
\begin{equation} \label{eq:pde_meshless_approx}
u(x) = \sum_{i=1}^N U_i \; \varphi\left(\frac{\lVert x - X_i \rVert}{h_i}\right).
\end{equation}
The various terms in the approximation \eqref{eq:pde_meshless_approx} are to be understood as follows. $(X_i \in \mathbb{R}^d)_{i=1}^N$ represent nodes in the domain $\Omega$. $\varphi:\mathbb{R}\to \mathbb{R}$ represents an RBF kernel. The variables $(h_i \in \mathbb{R})_{i=1}^N$ are appropriately defined measures of width of the RBF kernels centered at the nodes $(X_i)$. Finally, the coefficients $(U_i \in \mathbb{R})_{i=1}^N$ represent the nodal weights associated with the basis functions centered at $(X_i)_{i=1}^N$. In the sequel we also consider a variant of the meshless approximation \eqref{eq:pde_meshless_approx} that enforces the partition of unity property; such an approximation takes the form
\begin{equation} \label{eq:pde_meshless_PoU}
u(x) = \left(\sum_{j=1}^N \varphi\left(\frac{\lVert x - X_j \rVert}{h_j}\right)\right)^{-1}\sum_{i=1}^N U_i  \; \varphi\left(\frac{\lVert x - X_i \rVert}{h_i}\right).
\end{equation}
It is noted that the partition of unity meshless representation \eqref{eq:pde_meshless_PoU} satisfies the important property that it can represent any constant function on $\Omega$ exactly.

\subsection{Meshless approximation reinterpreted as a sparse DNN}
\begin{figure}
\centering
\includegraphics[width=0.8\textwidth]{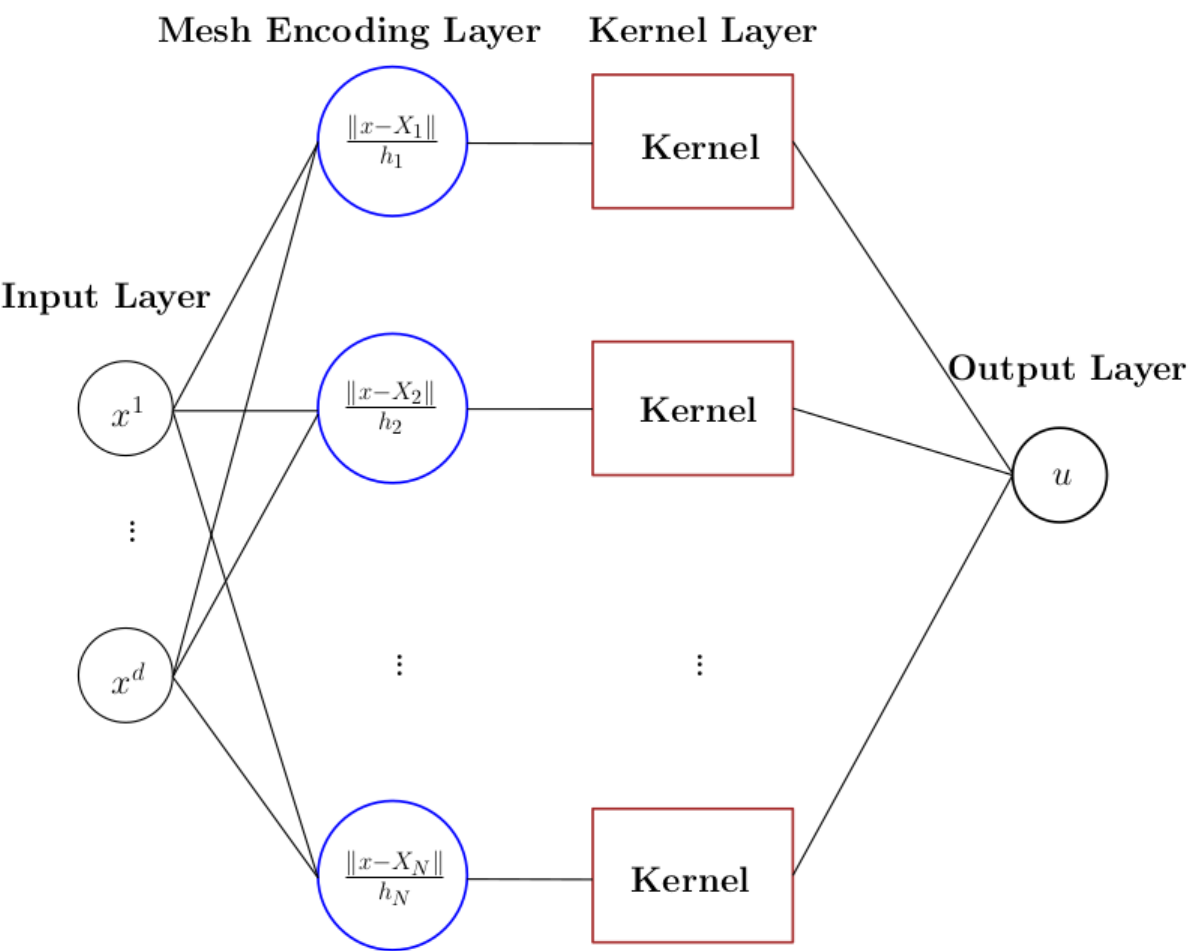}
\caption{Simplified SPINN architecture.}
\label{fig:meshless_nn_repr}
\end{figure}

The key idea behind SPINNs is the fact that meshless approximations like \eqref{eq:pde_meshless_approx} and \eqref{eq:pde_meshless_PoU} can be \emph{exactly} represented as specially structured sparse DNN. We first discuss how the meshless approximation \eqref{eq:pde_meshless_approx} can be written as a sparse DNN; the corresponding sparse DNN representation for the partition of unity approximation \eqref{eq:pde_meshless_PoU} is constructed analogously. The meshless approximation \eqref{eq:pde_meshless_approx} can be thought of as a DNN with an architecture as shown in Fig.~\ref{fig:meshless_nn_repr}.

We first transform the input  $x \in \Omega$ to the vector $(\lVert x - X_i\rVert/h_i)_{i=1}^N$ via hidden layers which we call the \emph{mesh encoding layer}, shown in blue in Fig.~\ref{fig:meshless_nn_repr}. In more detail, the mesh encoding layer first transforms the input $x \in \Omega$ to a hidden layer with $Nd$ neurons that have input weights
\begin{displaymath}
\left(\underbrace{\frac{1}{h_1}, \ldots, \frac{1}{h_1}}_{\text{$d$ terms}}, \underbrace{\frac{1}{h_2}, \ldots, \frac{1}{h_2}}_{\text{$d$ terms}}, \ldots, \underbrace{\frac{1}{h_N}, \ldots, \frac{1}{h_N}}_{\text{$d$ terms}} \right),
\end{displaymath}
and biases being the $Nd$ vector
\begin{displaymath}
(-X_1^1, \ldots, -X_1^d, -X_2^1, \ldots, -X_2^d, \ldots, -X_N^1, \ldots, -X_N^d),
\end{displaymath}
and with the function $\text{sqr}:\mathbb{R} \to \mathbb{R}$ defined as $\text{sqr}(z) = z^2$ as their activation functions. The output of the first hidden layer of the mesh encoding layer is thus the $Nd$-vector
\begin{displaymath}
\left(\frac{(x - X_1^1)^2}{h_1^2}, \ldots, \frac{(x - X_1^d)^2}{h_1^2}, \frac{(x - X_2^1)^2}{h_2^2}, \ldots, \frac{(x - X_2^d)^2}{h_2^2}, \ldots, \frac{(x - X_N^1)^2}{h_N^2}, \ldots, \frac{(x - X_N^d)^2}{h_N^2}\right).
\end{displaymath}
This is then transformed to another hidden layer consisting of $N$ neurons each of which takes $d$ inputs with weights $1$ and has the function $\text{sqrt}:\mathbb{R} \to \mathbb{R}$ defined as $\text{sqrt}(z) = \sqrt{z}$ as the activation function to produce the $N$-vector
\begin{displaymath}
\left(\frac{\lVert x - X_1\rVert}{h_1}, \frac{\lVert x - X_2\rVert}{h_2}, \ldots, \frac{\lVert x - X_N\rVert}{h_N}\right).
\end{displaymath}
This vector is then passed to the \emph{kernel layer}, shown in brown in Fig.~\ref{fig:meshless_nn_repr}, that consists of $N$ neurons with unit input weights and the the RBF kernel $\varphi$ as the activation function. The outputs of the kernel layer, which is the vector $(\varphi(\lVert x - X_i\rVert/h_i)_{i=1}^N$, is then linearly combined using weights $(U_i)$, which are the coefficients of the meshless approximation \eqref{eq:pde_meshless_approx}, to compute the final output $u(x)$ according to the ansatz \eqref{eq:pde_meshless_approx}. This demonstrates that the meshless ansatz \eqref{eq:pde_meshless_approx} is exactly representable as a DNN with a special architecture as described above. We wish to highlight two important aspects of this architecture: (i) it is sparse; the number of connections and trainable parameters of this network are much smaller than a DNN with the same number of hidden layers and neurons, and (ii) the trainable coefficients of this network, namely the vectors $(h_i)$, $(X_i)$ and $(U_i)$, are interpretable directly in terms of the meshless ansatz \eqref{eq:pde_meshless_approx}.

\begin{figure}
\centering
\includegraphics[width=0.9\textwidth]{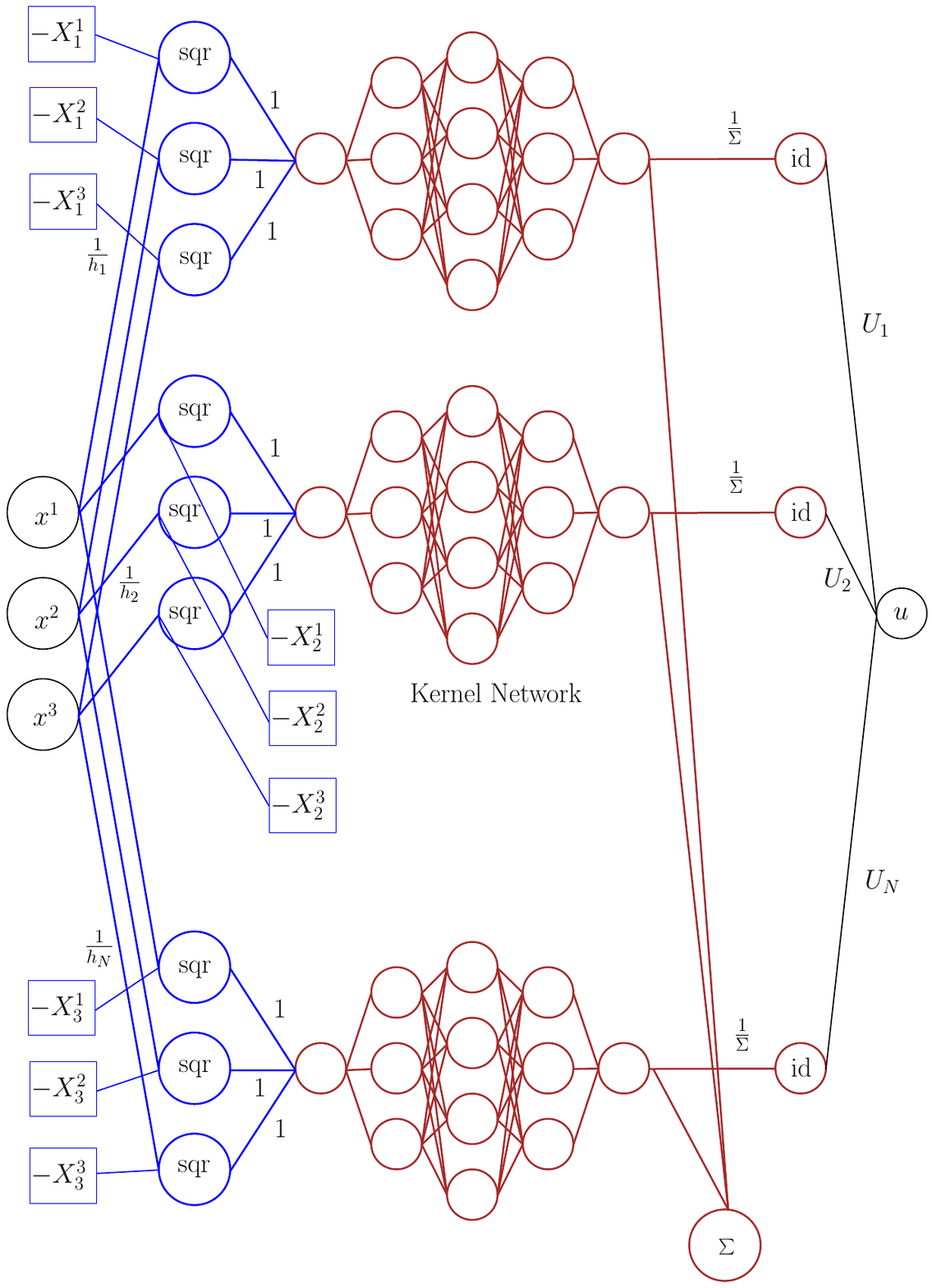}
\caption{A detailed view of SPINN with DNN kernel. \rr{Note that the kernel networks are exact copies of each other. They are shown separately to emphasize the sparsity of the connection between the mesh encoding and kernel layers.}}
\label{fig:meshless_nn_detailed}
\end{figure}

\subsection{SPINN architecture}
The foregoing discussion naturally motivates the introduction of generalized meshless approximations where the kernel is represented using a DNN. For instance, a partition of unity meshless approximation of the form \eqref{eq:pde_meshless_PoU} with the kernel replaced by a DNN is shown in Figure~\ref{fig:meshless_nn_detailed}. It can be seen that the mesh encoding layers, shown in blue in Figure~\ref{fig:meshless_nn_detailed}, are identical to that described earlier in the context of the DNN equivalent of \eqref{eq:pde_meshless_approx}. The primary difference is that instead of using an RBF kernel as the activation function in the kernel layer, a standard DNN with any differentiable activation function, shown in brown in Figure~\ref{fig:meshless_nn_detailed}, is used as the kernel; we call this the \emph{kernel network} in the sequel. It is worth pointing out that the \emph{same} kernel network is used for each of the outputs of the mesh encoding layer, in conformity with the meshless ansatz \eqref{eq:pde_meshless_PoU}. This drastically reduces the number of connections in the SPINN architecture in comparison with a DNN network having a similar architecture. We therefore see that the SPINN architecture proposed here is more efficient than conventional DNN based methods such as Deep Ritz \cite{EYu2018}, PINN \cite{RPK2019}, etc. Note also that the number of neurons in the mesh encoding layer is exactly the same as the number of nodes used in the meshless discretization. We thus have a physically inspired means to fix the size of the hidden layers in SPINN, unlike other DNN based approaches like PINN where the size of the hidden layers is arbitrary.  Further, except for the parameters of the kernel network, the remaining learnable parameters of the network are fully interpretable. In fact, once the SPINN model is trained, it is straightforward to extract the corresponding meshless ansatz. The parameters of the kernel network do not require to be interpreted since the kernel network itself can be thought of as a generalized RBF kernel.

To emphasize the interpretability of SPINN using as an example the architecture shown in Figure~\ref{fig:meshless_nn_detailed}, which is an instance of the meshless ansatz \eqref{eq:pde_meshless_PoU} with a kernel network in place of the RBF kernel $\varphi$. The interpretability of SPINN is to be understood as follows:

\begin{enumerate}[(i)]
\item The weights connecting the input layer to the first hidden layer encode the widths $(h_i)$ of the meshless kernels located at each node.
\item The biases of the first hidden layer, on the other hand, encode the coordinates $(X_i^1, \ldots, X_i^d)$ of the nodes, which are the centers of the meshless kernels.
\item The kernel network, which in this case is a DNN, stands for the RBF kernel $\varphi$ that the meshless ansatz uses.
\item Finally, the weights of the final connecting the output of the kernel networks to the output layer encodes the coefficients $(U_i)$ of the meshless ansatz.
\end{enumerate}

We thus see that the SPINN network is  interpretable \rR{in the sense outlined above}. In particular, given a SPINN model, it is straightforward to extract a meshless approximation using the procedure elaborated above. This provides a principled rationale for choosing the internal architecture of SPINN based on the corresponding meshless representation. This is in sharp contrast to methods like PINN, Deep Ritz, etc., where the internal architecture of the DNNs used is arbitrary. \rR{All the non-interpretability of SPINN is thus reduced to the non-interpretability of a \emph{much smaller} kernel network. It bears emphasis nevertheless that the kernel network itself is interpretable as a whole as a basis function.}

We also highlight the fact that since the positions of the nodes and the widths of the kernel associated with the nodes are trainable parameters of the network, the learning algorithm implicitly encodes mesh adaptivity as part of the training process. We note in particular that for problems with large gradients, the widths of the kernels naturally develop a multiscale hierarchy during training, as will be demonstrated in the \emph{Results} section.

\subsection{Kernel functions}
\begin{figure}[htpb]
\centering
\includegraphics[width=0.5\textwidth]{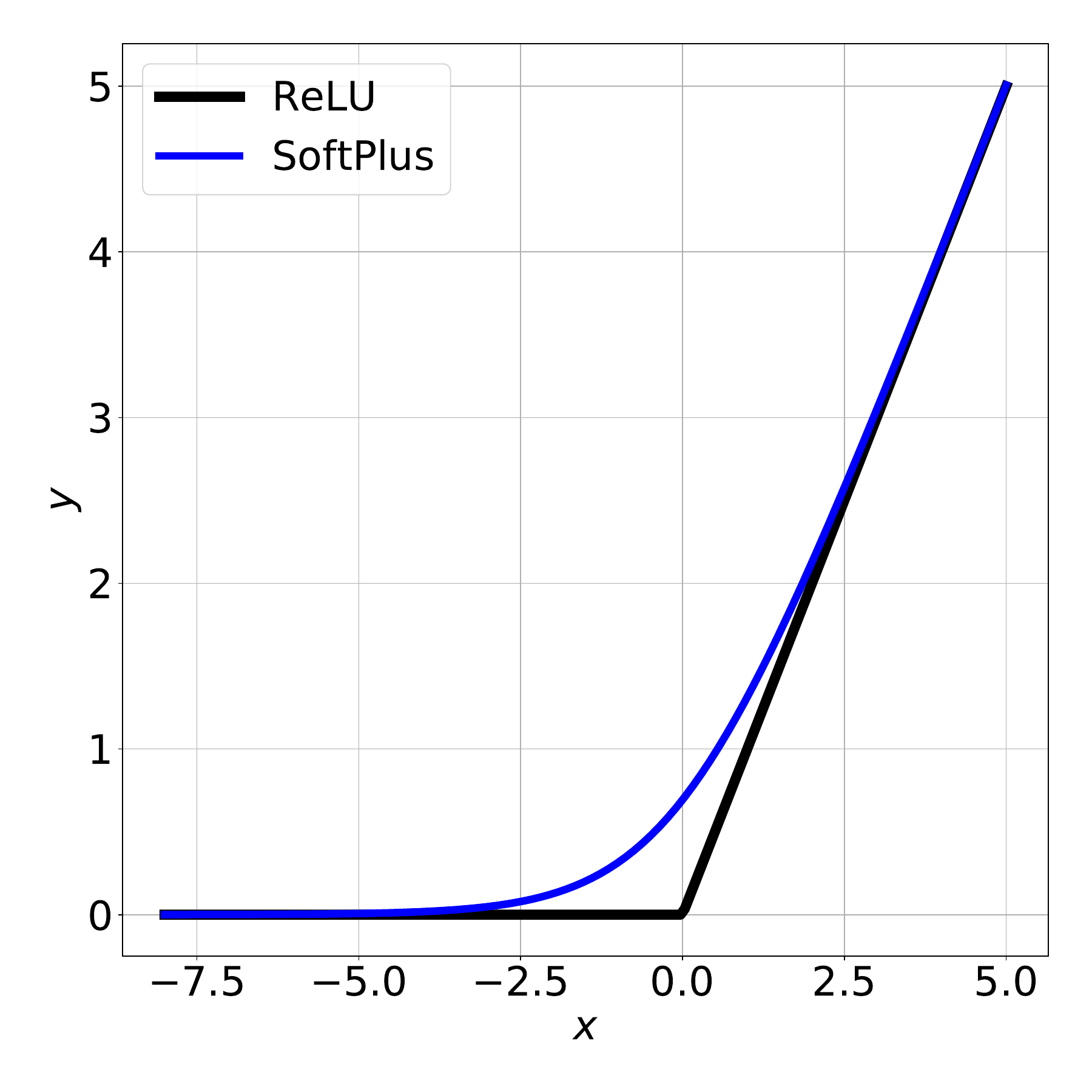}
\caption{Comparison of the softplus and ReLU functions.}
\label{fig:softplus_relu}
\end{figure}

\begin{figure}[htpb]
\begin{subfigure}{0.4\textwidth}
\centering
\includegraphics[width=\textwidth]{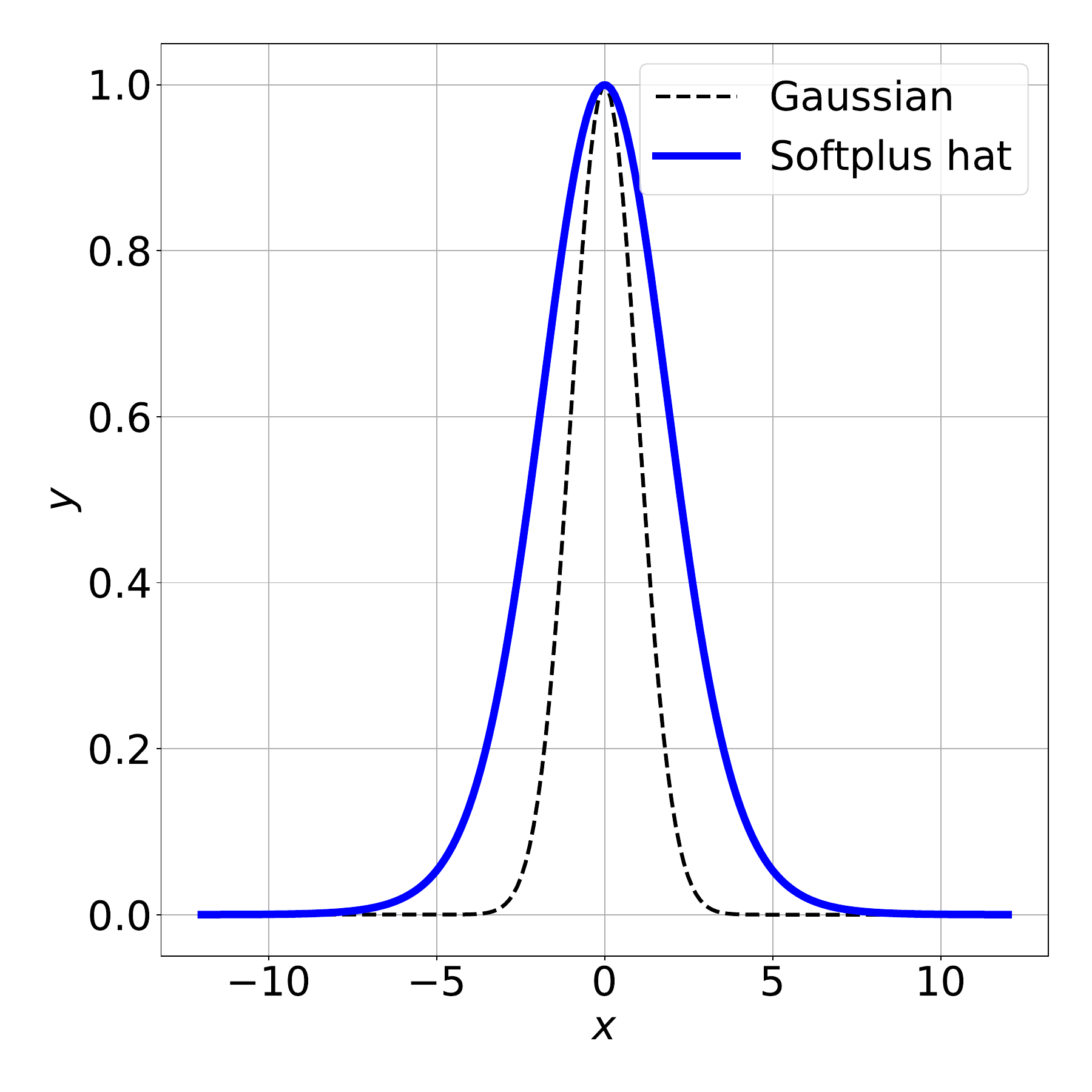}
\caption{Softplus hat function in 1D}
\label{fig:softplus_hat_1d}
\end{subfigure}
~
\begin{subfigure}{0.6\textwidth}
\centering
\includegraphics[width=\textwidth]{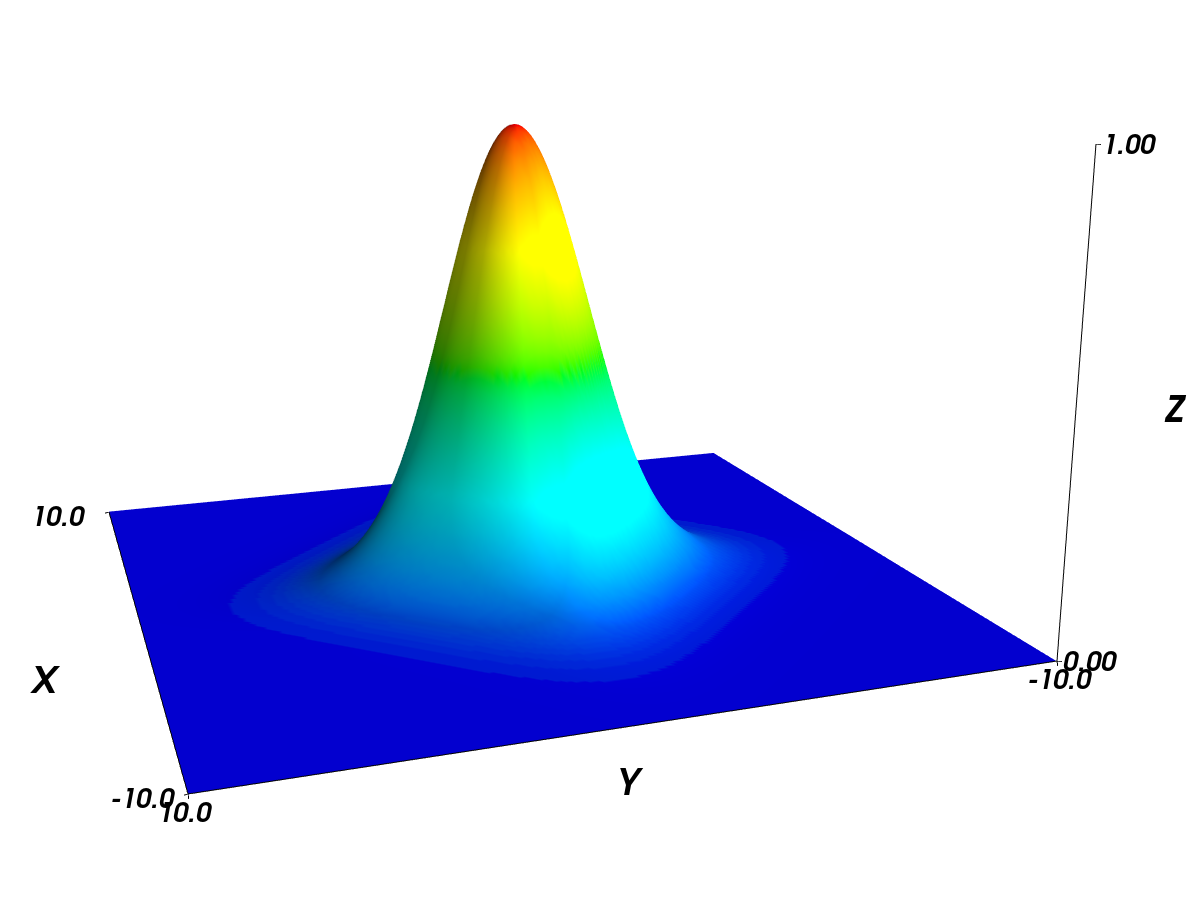}
\caption{Softplus hat function in 2D}
\label{fig:softplus_hat_2d}
\end{subfigure}
\caption{Softplus based kernel functions in 1 and 2 dimensions.}
\end{figure}

The restriction that the kernel is a radial basis function can be easily removed. It is well known \cite{HLXZ2020} that one-dimensional piecewise linear finite element basis functions can be written exactly in terms of the ReLU basis functions; details are provided in Appendix~\ref{app:relu_fem_1d}. However, ReLU functions are not differentiable, and this can pose problems for their use in ODEs and PDEs of order two or higher. Motivated by the connection between RELU activation functions and hat functions, we propose a new class of basis functions that are infinitely differentiable. We note first that the softplus function
\begin{equation} \label{eq:softplus}
\rho(x) = \log (1 + \exp x),
\end{equation}
provides a smooth approximation of the ReLU function, as shown in Figure~\ref{fig:softplus_relu}.
The softplus function can now be used to create a basis function with \emph{almost} compact support in a manner analogous to the construction of piecewise linear finite element basis functions using ReLU functions. Specifically, we note that the function
\begin{equation} \label{eq:hat_softplus_1d}
N(x) = \frac{1}{\rho(1)}\rho\left(1 + 2\log 2 - \rho(x) - \rho(-x)\right)
\end{equation}
resembles kernels used in meshless approximations; we call this the \emph{softplus hat kernel}. The constants are chosen such that $N(0) = 1$. A graph of the softplus hat function is shown in Figure~\ref{fig:softplus_hat_1d}.

What makes the softplus hat kernel interesting is the fact that it can be exactly represented as a two layer neural network. The input $x$ first feeds into a hidden layer with two neurons, with weights $(1, -1)$, bias $0$ and softplus activation function. The output of this hidden layer $(\rho(x), \rho(-x))$ is then linearly combined using a second hidden layer with one neuron with weights $(-1, -1)$, bias $1 + 2\log 2$ and softplus activation function. The output of this layer is therefore $\rho\left(1 + 2\log 2 - \rho(x) - \rho(-x)\right)$. Finally, it is straightforward to scale this by a factor of $1/\rho(1)$ to get the final output $N(x)$. We thus see that the softplus hat function is indeed transcribable exactly as a neural network.

\begin{figure}
\centering
\includegraphics[width=0.3\textwidth]{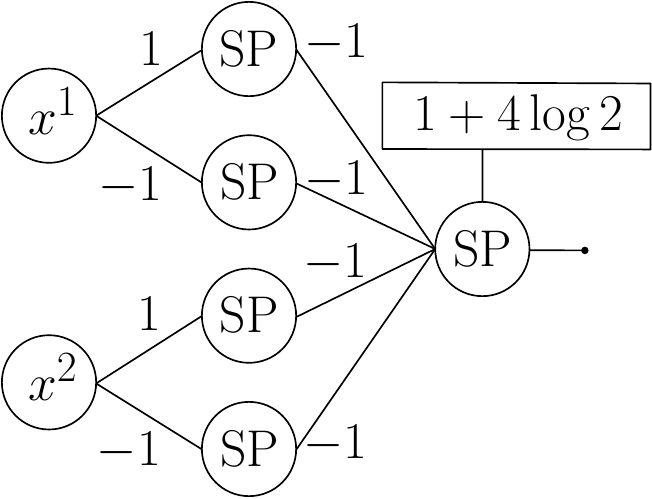}
\caption{Softplus hat kernel represented as a neural network.}
\label{fig:softplus_hat_nn}
\end{figure}

It is straightforward to generalize this to higher dimensions. For instance, a $d$ dimensional softplus hat kernel is given by
\begin{equation} \label{eq:hat_softplus_nd}
N(x^1, \ldots, x^d) = \frac{1}{\rho(1)}\rho\left(1 + 2d\log 2 - \sum_{k=1}^d (\rho(x^i) + \rho(-x^i))\right).
\end{equation}
We emphasize that the higher dimensional softplus hat kernel functions \eqref{eq:hat_softplus_nd} are also representable directly as a two layer neural network; this is illustrated for two dimensional softplus hat kernels in Figure~\ref{fig:softplus_hat_nn}. A graph of softplus hat kernel in two dimensions is shown in Figure~\ref{fig:softplus_hat_2d}. Higher dimensional softplus hat functions are also exactly represented by an equivalent neural network.

Though the softplus hat kernels do not have compact support, they approach zero quickly outside a small neighborhood of the central node, just like Gaussian kernels. Thus they are expected to have the same performance as Gaussian RBF kernels. This is indeed borne out by the numerical experiments, as will be demonstrated shortly.

We wish to point that the softplus hat kernel is new and has not been used before to the best of the our knowledge. With the choice of these softplus hat kernel functions, meshless approximations that go beyond radial basis functions are constructed along the same lines mentioned before. We present many examples using the softplus hat kernel in the Results section.

\subsection{Loss definition}
With this choice of architecture, solving the PDE, $\mathcal{N}(x, u, \nabla u, \ldots) = 0$, is easily accomplished using a collocation technique similar to the one used in PINNs \cite{RPK2019}. We also note in passing that for PDEs that are obtained as the Euler-Lagrange equations of a known functional, the loss function can be formulated using quadratures of the integral of the functional along with penalty terms to enforce boundary conditions, as is carried out in the Deep Ritz method \cite{EYu2018}.

To illustrate the loss functions used in training the SPINN models, consider the special case of a second order PDE of the form
\begin{displaymath}
\begin{split}
\mathcal{N}(x, u(x), \nabla u(x), \nabla^2 u(x)) = 0, \quad x \in \Omega,\\
u = u_0 \;\text{ on }\partial_0 \Omega, \quad \nabla u \cdot n = g_0 \;\text{ on } \partial \Omega \setminus \partial_0 \Omega,
\end{split}
\end{displaymath}
where $n$ is the outward unit normal to $\partial \Omega$. The generalization to other ODEs and PDEs is straightforward. The loss function for traing the SPINN network for this problem is chosen as
\begin{equation} \label{eq:spinn_loss}
\begin{split}
L((h_i), (X_i), (U_i)) &= w_i \sum_{i=1}^{M_i} \mathcal{N}(\xi_i, u(\xi_i), \nabla u(\xi_i), \nabla^2 u(\xi_i))\\
 &+ w_d \sum_{i=1}^{M_d} (u(\eta_i) - u_0(\eta_i))^2\\
 &+ w_n \sum_{i=1}^{M_n} \left(\nabla u(\zeta_i)\cdot n(\zeta_i) - g_0(\zeta_i)\right)^2.
\end{split}
\end{equation}
Here $w_i$, $w_d$ and $w_n$ are constants that enforce the loss in the interior of the domain, the Dirichlet boundary, and the Neumann boundary, respectively. The set of points $(\xi_i)_{i=1}^{M_i}$, $(\eta_i)_{i=1}^{M_d}$ and $(\zeta_i)_{i=1}^{M_n}$ in the interior of $\Omega$, and on the Dirichlet and Neumann boundaries on $\Omega$, respectively, are the sampling points where the loss is evaluated. \rb{The sampling points are chosen to be uniformly distributed points inside the domain and on its boundary. For regular domains, we choose a fine uniform grid and identify the vertices of the grid as the interior and boundary sampling points. For irregular domains, we generate a fine triangulation and choose the vertices of the triangulation as the sampling points. Note that the precise mode of generation of the sampling points is not as important as the fact that every region of the domain is sufficiently represented by the sampling points.  We choose a uniformly spaced distribution of the sampling points in the examples, but more general distributions that are tailor made for specific problems can be easily incorporated in the current framework.}

In cases where a variational form
\begin{equation} \label{eq:weak_form}
I(u) = \int_{\Omega} f(x, u(x), \nabla u(x))\,dx
\end{equation}
is available for the PDE, the interior loss can alternatively be defined directly in terms of an appropriate quadrature approximation of the integral in \eqref{eq:weak_form}. Dirichlet boundary conditions are imposed using penalty terms as in the strong form collocation case described earlier. Neumann boundary conditions, on the other hand, are directly integrated into the definition of the integral loss functional.

\subsection{Time dependent PDEs}
We present two different approaches for time dependent PDEs using SPINN. The first employs a space-time grid of dimension $d + 1$ and uses exactly the same ideas presented above to solve a time dependent PDE. \rr{This is similar to the spacetime DNN used in works like \cite{RPK2019}.} Alternatively, a hybrid finite difference and SPINN method, which we call FD-SPINN, can be employed which performs time marching using conventional finite difference methods and performs spatial discretization at each time step using the SPINN architecture. It is worth mentioning that both explicit and implicit finite difference schemes are subsumed in FD-SPINN. \rr{This can also be generalized to multi-step methods as is done in works like \cite{RPK18multistep}.} Both the space-time and the FD-SPINN methods will be illustrated later on.

We illustrate an implicit FD-SPINN algorithm in the special case of the 1D heat equation $u_t = u_{xx}$; a similar formalism applies for other time dependent PDEs. Choosing a time step $\Delta t$, we denote by $(u^k(x))_{k=1}^{N_t}$ the approximations to the solution $u(x, k\Delta t)$; here $N_t$ is chosen such that $N_t \Delta t \simeq T$. The following first order implicit time difference scheme is used in this work:
\begin{displaymath}
\frac{u^{n+1}(x) - u^n(x)}{\Delta t} = \frac{d^2 u^{n+1}(x)}{d x^2}, \quad n = 1, \ldots, N_t.
\end{displaymath}
\rb{It is to be noted that the parameters of the network are \emph{not} re-initialized at each time step. Rather, the converged parameters from the previous time step are used as the initial values of the parameters for the current time step.} We wish to emphasize that the spatial derivatives are computed \emph{exactly} using automatic differentiation since the spatial approximation uses SPINN. Thus this implicit scheme is different in comparison to traditional time marching schemes. The loss in the interior of the domain $[0,L]$ is computed as the squared residue of the foregoing equation. The second algorithm uses SPINN for both space and time discretization. The implementation of the space-time SPINN solution is similar to the implementation of second order PDEs described earlier.

\subsection{Fourier-SPINN}
Another advantage of the SPINN architecture is that it suggests natural generalizations of familiar decompositions of functions. To make this precise, consider the Fourier expansion of a function $u:[a,b] \to \mathbb{R}$, namely
\begin{displaymath}
u(x) = a_0 + \sum_{k=1}^{\infty} a_k \cos k\omega x + \sum_{k=1}^{\infty} b_k \sin k\omega x,
\end{displaymath}
where $\omega = 2\pi/(b - a)$. It is straightforward to reinterpret this as a neural network with one hidden layer that has the sinusoidal functions as the activation functions of the neurons. The input $x$ is transformed using a hidden layer with $2N$ neurons to the scaled inputs
\begin{displaymath}
(x, 2x, \ldots, Nx, x, 2x, \ldots, Nx) \in \mathbb{R}^{2N}.
\end{displaymath}
The activation function of the $2N$ neurons in the hidden layer are chosen as
\begin{displaymath}
(\underbrace{\cos \omega z, \cos \omega z, \ldots, \cos \omega z}_{\text{$N$ terms}}, \underbrace{\sin \omega z, \sin \omega z, \ldots, \sin \omega z}_{\text{$N$ terms}}).
\end{displaymath}
The biases of the hidden layer are uniformly set to zero. The output of this hidden layer is then linearly combined to produce the output
\begin{displaymath}
u(x) = U_0 + \sum_{k=1}^N U_k \cos k\omega x + \sum_{k=1}^N V_k \sin k\omega x,
\end{displaymath}
which is just the Fourier representation of $u$. The Fourier representation is then used in conjunction with an appropriately defined loss function to solve a given differential equation for $u$. It bears emphasis that the weights and biases of this neural network are fully interpretable as in the case of the meshless approximation discussed earlier.

\rr{We wish to point out that the specific version of Fourier-SPINN that we propose here is different from other Fourier based methods like \cite{WANG2021113938, LPYWVJ21}, where the authors extract Fourier features of the inputs, which are then fed as inputs to a DNN. In contrast, the method we propose here directly transcribes a Fourier decomposition as a neural network architecture, thereby suggesting many generalizations. It is possible however to combine both these approaches and study hybrid variants; we do not pursue those directions in this paper.}

We also note in passing that wavelet transforms, see for instance \cite{WA94}, could also be represented using SPINN. As a natural generalization one could replace the sinusoidal functions by a DNN thereby providing a neural network generalization of the Fourier transform. Investigations of such extensions are planned for future work. \new{We remark that the examples studied using SPINN presented later already show multiresolution capabilities in the specific manner in which it adapts the nodes and kernel widths.}

\subsection{Details of Implementation}
The SPINN architecture proposed in this work is easily implementable using standard NN libraries. We provide PyTorch~\cite{pytorch} implementations of all the examples considered in this work at \url{https://github.com/nn4pde/SPINN}.

We present here certain details of the implementation. We classify the nodes associated with the SPINN model as fixed and free nodes. Fixed nodes are typically used on the (Dirichlet) boundaries. Free nodes, on the other hand, are used inside the domain of definition of the problem. Both fixed and free nodes are designed to have variable kernel widths, but the free nodes are also free to move both inside and outside the domain. A separate set of sampling points on the interior and boundary of the domain are also used. These are the points where the interior and boundary losses are evaluated using the SPINN model. We provide options for both full sampling and random sampling. In random sampling, a random subset of an initially generated set of sampling points is chosen for each iteration of the loss minimization algorithm. \rb{It is to be noted that what we call as random sampling is the same as mini-batching in the machine learning literature. The weights and biases of both the hidden and output layers are set to zero by default.} This implies in particular that when using full sampling and either the Gaussian or softplus hat kernels, the SPINN algorithm is fully deterministic. Thus there are only two sources of stochasticity in the current implementation of SPINN: (i) randomness due to initialization of the DNN kernel weights and biases when DNN kernels are used, and (ii) randomness due to sampling of the interior sampling points when random sampling is used. We do not sample on the boundary and use full boundary sampling always. This is because of the fact that boundary conditions are more delicate to impose in the current framework. The optimization of the SPINN models is carried out using any of the well known optimizers. All the examples presented in this work were carried out using the Adam optimizer \cite{kingma2014} implemented in PyTorch.

\section{Results}
We now present solutions of a variety of ordinary and partial differential equations using SPINN. We implement the SPINN architecture using PyTorch~\cite{pytorch}, and the code is available at \url{https://github.com/nn4pde/SPINN}. All our results are automated using \verb|automan|~\cite{automan:2018} for easy reproducibility. \rb{We have provided details of various parameters used to arrive at the solutions. However, we wish to point out that our entire code including all the examples shown here are available at the aforementioned Github repository, a perusal of the \texttt{automate.py} file should provide full details of the various parameters used. All our simulations are thus \emph{fully reproducible}.} \rr{We also point out that the error values for \emph{each of the simulations} across training steps are automatically recorded and archived in the \texttt{outputs/} folder when the codes are run using the procedure outlined in the repository. Unless stated otherwise, all simulations use full sampling.  When random sampling is used, the relevant details are specified explicitly.}

\subsection{Ordinary differential equations}
To validate the SPINN method, we first consider ordinary differential equations (ODEs) with different boundary conditions. \new{We study a few ODEs primarily to illustrate various aspects of the SPINN algorithm in a simpler setting before moving on to PDEs.} Except for the cases involving variational SPINN, strong form collocation is used to compute the loss function in all examples shown here.

\subsubsection{ODE with Dirichlet boundary conditions}
\begin{figure}
\centering
\begin{subfigure}{0.3\textwidth}
\includegraphics[width=\textwidth]{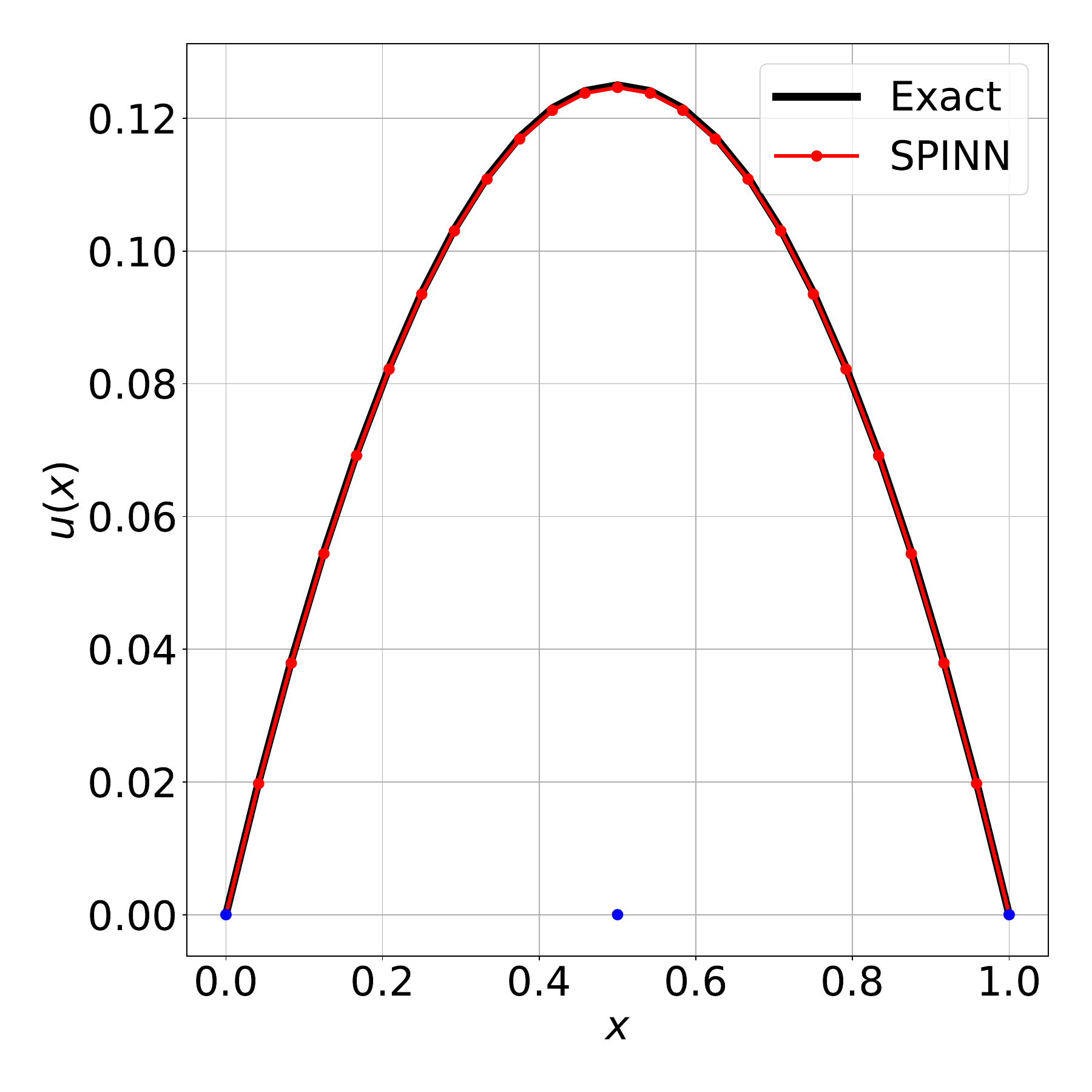}
\caption{$n = 1$}
\label{fig:ode1_n_1}
\end{subfigure}
~
\begin{subfigure}{0.3\textwidth}
\includegraphics[width=\textwidth]{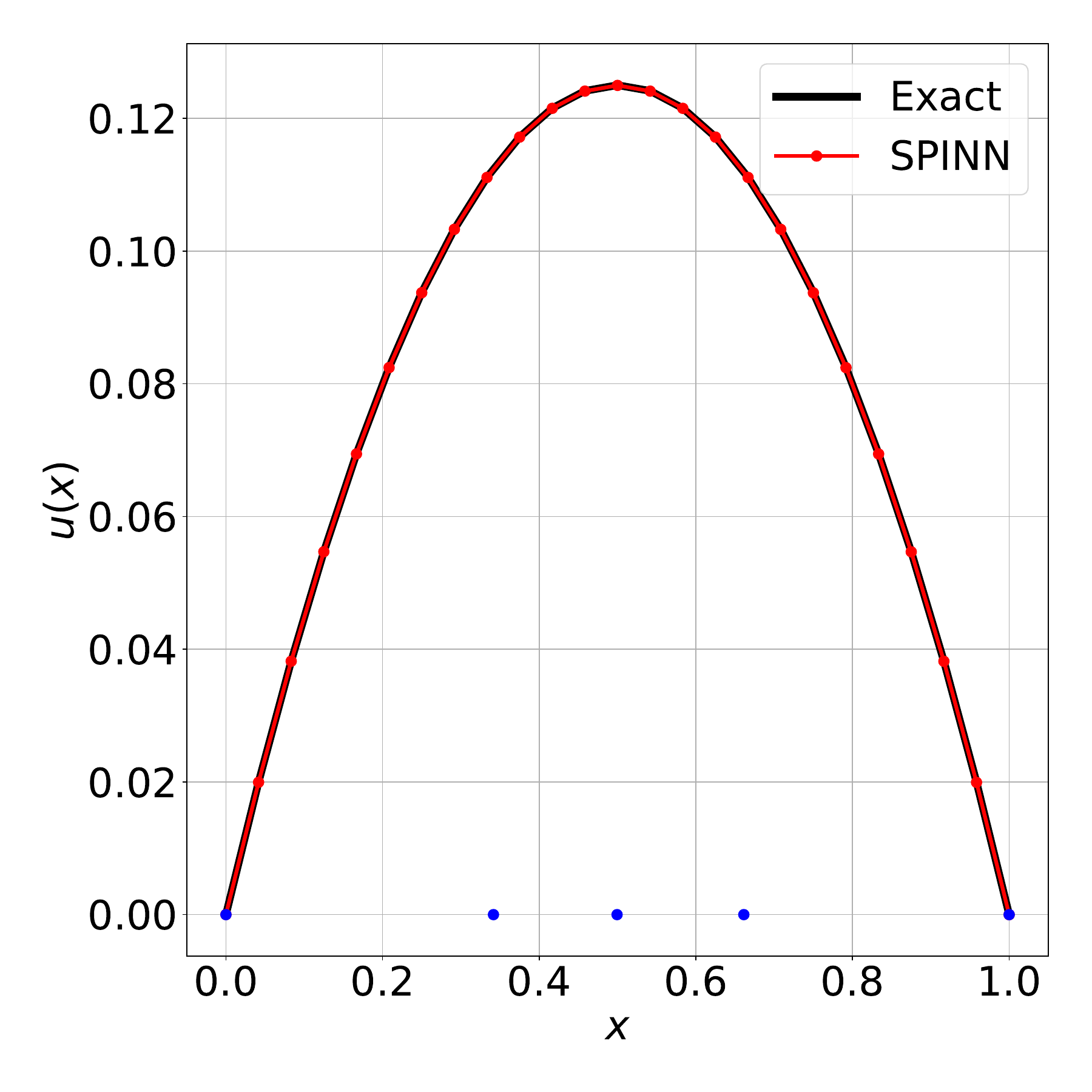}
\caption{$n = 3$}
\label{fig:ode1_n_3}
\end{subfigure}
~
\begin{subfigure}{0.3\textwidth}
\includegraphics[width=\textwidth]{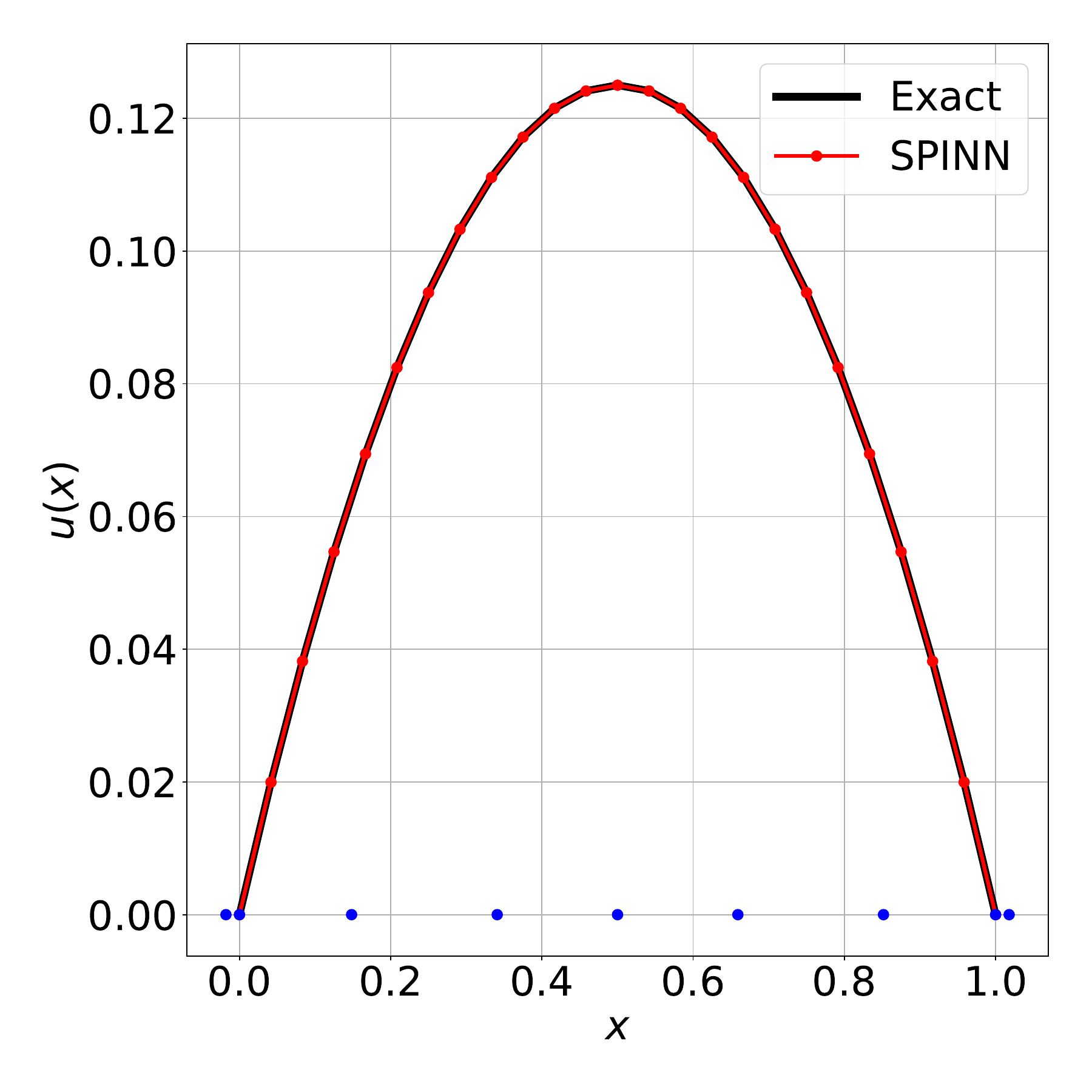}
\caption{$n = 7$}
\label{fig:ode1_n_7}
\end{subfigure}
\caption{Solution of the ODE \eqref{eq:ode1} using SPINN with Gaussian kernel using $1$, $3$ and $7$ interior nodes. The nodal positions learnt by SPINN are shown as blue circles along the $x$ axis.}
\label{fig:spinn_ode_1}
\end{figure}

\begin{figure}
\centering
\begin{subfigure}{0.3\textwidth}
\includegraphics[width=\textwidth]{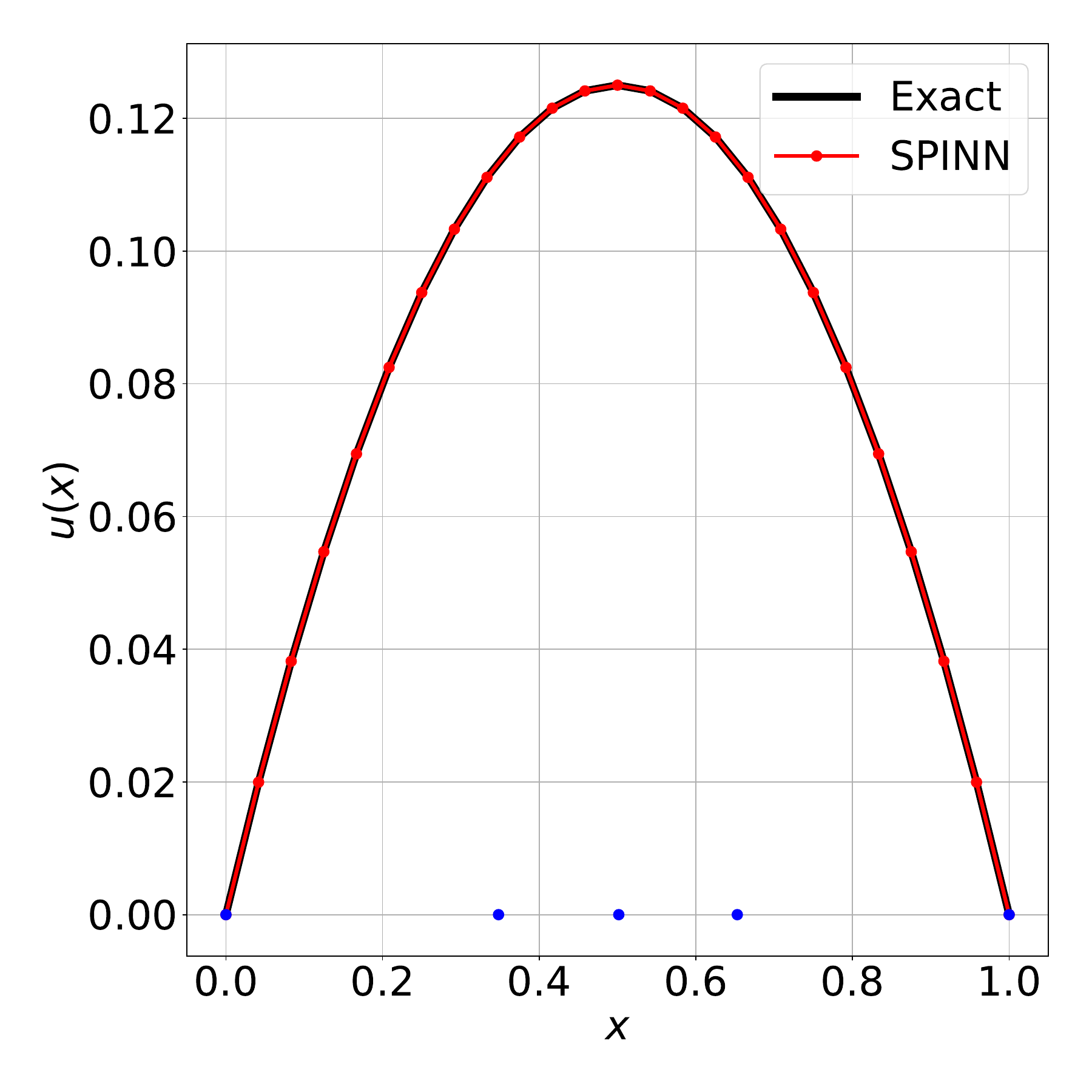}
\caption{Gaussian kernel}
\label{fig:ode1_n_3_gaussian}
\end{subfigure}
~
\begin{subfigure}{0.3\textwidth}
\includegraphics[width=\textwidth]{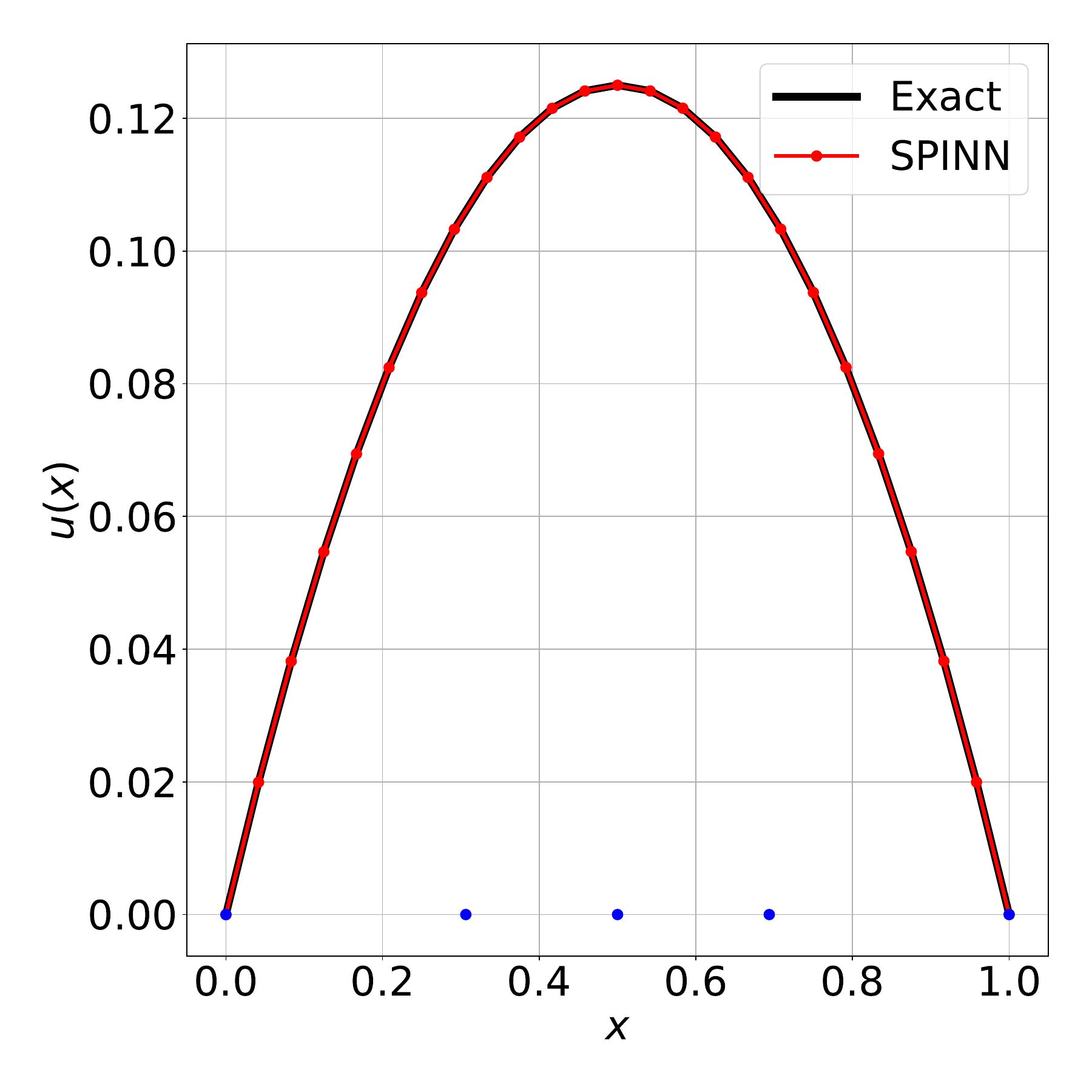}
\caption{Softplus hat}
\label{fig:ode1_n_3_softplus}
\end{subfigure}
~
\begin{subfigure}{0.3\textwidth}
\includegraphics[width=\textwidth]{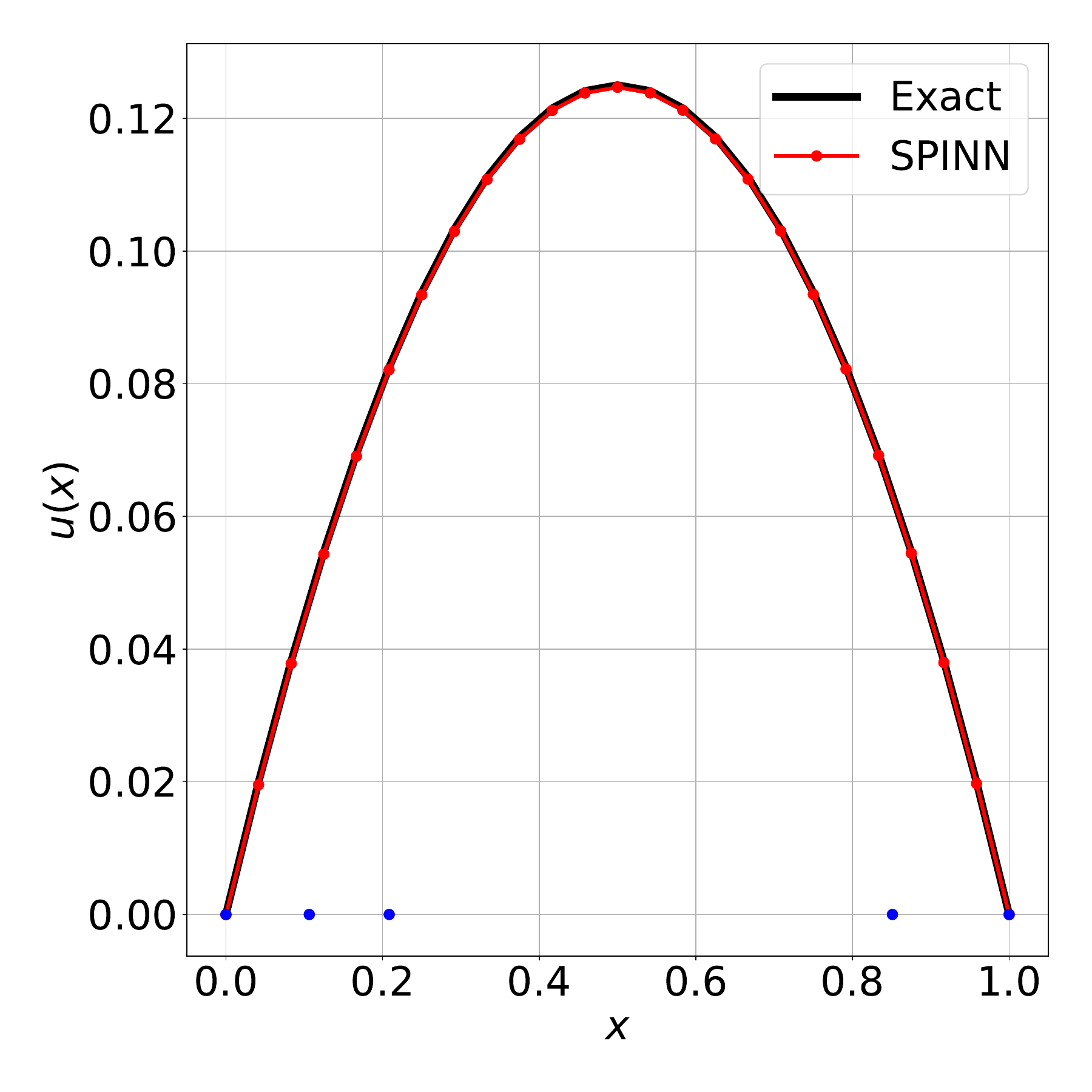}
\caption{Neural network}
\label{fig:ode1_n_3_kernel}
\end{subfigure}
\caption{Solution of the ODE \eqref{eq:ode1} using SPINN with different kernels. All simulations use $n = 3$ interior nodes. The nodal positions learnt by SPINN are shown as blue circles along the $x$ axis.}
\label{fig:spinn_ode_1_kernels}
\end{figure}

\begin{figure}
\centering
\includegraphics[width=0.5\textwidth]{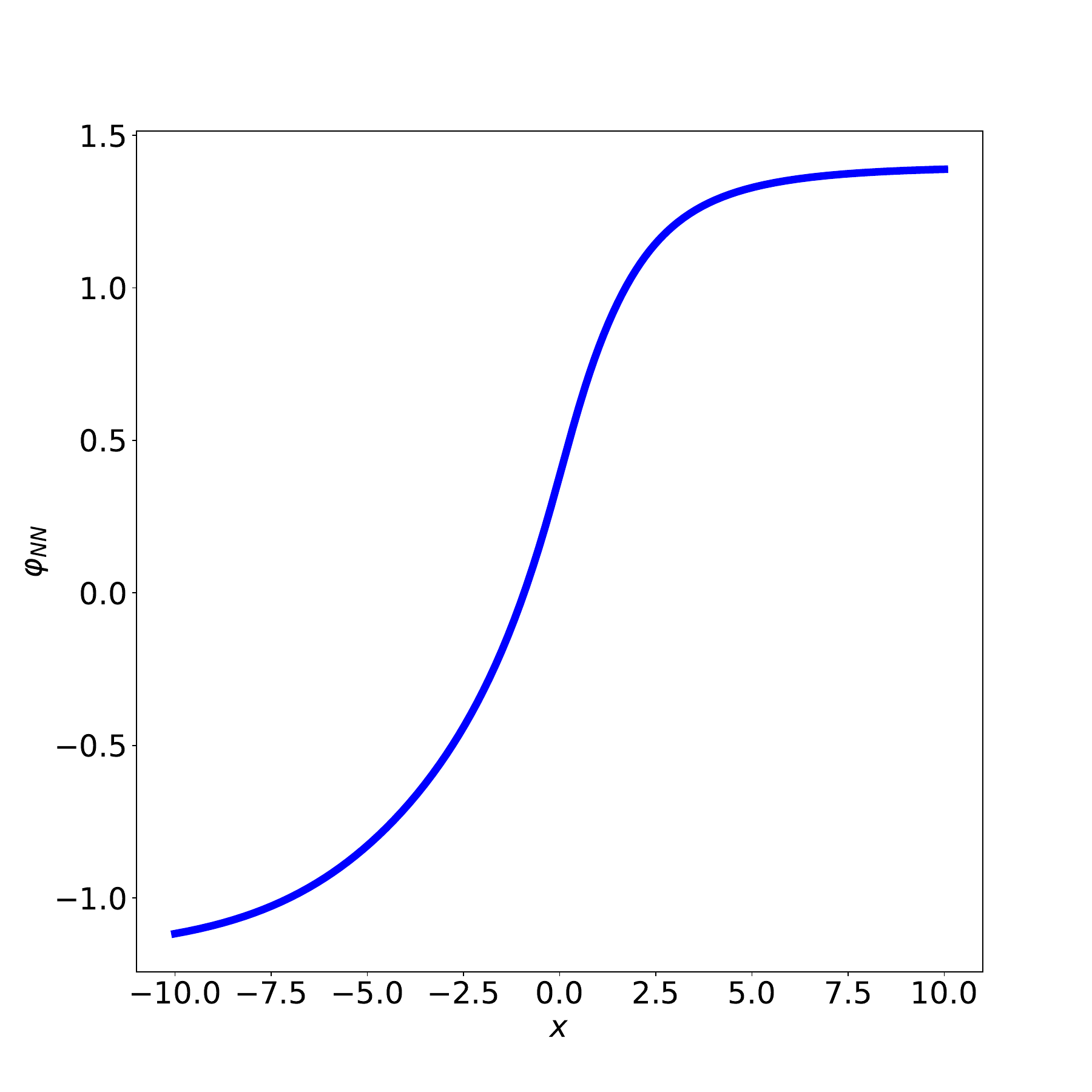}
\caption{Basis learnt by SPINN with neural network kernel. Note that the basis is asymmetric. This is reflected in the aysmmetric placement of the nodes in Figure~\ref{fig:ode1_n_3_kernel}.}
\label{fig:ode1_kernel}
\end{figure}

\new{To begin with we consider the ODE
\begin{equation} \label{eq:ode1}
\begin{split}
\frac{d^2 u(x)}{dx^2} + 1 = 0, &\quad x \in (0,1),\\
u(0) = u(1) &= 0.
\end{split}
\end{equation}
The exact solution for the ODE \eqref{eq:ode1} is easily computed as
\begin{equation} \label{eq:ode1_exact}
u(x) = \frac{1}{2}x(1 - x).
\end{equation}}
\rb{The solution computed using SPINN with Gaussian kernel, $n=1,3,7$ uniformly placed internal nodes, two fixed boundary nodes on the left and right boundaries $x = 0$ and $x = 1$ respectively is shown in Figure~\ref{fig:spinn_ode_1}. The nodal positions learnt by the SPINN algorithm are also shown therein. For all these simulations $n_S = 6n$ sampling points were used to compute the interior loss, while two boundary nodes, one at each boundary, were used to compute the boundary loss. Both the nodes and sampling points were initially uniformly distributed over the domain. As can be seen from Figure~\ref{fig:spinn_ode_1} the nodes cluster towards regions where the solution gradient is higher. A learning rate of $10^{-4}$ was used in all these simulations.} \new{For the case when $n=3$, we present the solutions obtained by using the Gaussian, Softplus and NN kernels in Figure~\ref{fig:spinn_ode_1_kernels}.} \rb{The kernel network was chosen as a fully connected NN with 2 hidden layers with 5 neurons each. We note that this is a much smaller network than what is typically employed in PINNs. The parameters of the kernel network are initialized randomly, unlike the other two kernels where all parameters are set to zero initially.} \new{Curiously, while both the Gaussian and softplus hat kernels show symmetric node placements, the node distribution for the case when the kernel is chosen as a neural network is asymmetric. To understand this better we show in Figure~\ref{fig:ode1_kernel} the kernel function that SPINN learns. It is seen that the kernel learnt by SPINN in this case is actually asymmetric, unlike the Gaussian and softplus hat kernels. This causes a corresponding asymmetry in the nodal positions observed in Fig.~\ref{fig:ode1_n_3_kernel}.}

\begin{figure}
\centering
\begin{subfigure}{0.3\textwidth}
\includegraphics[width=\textwidth]{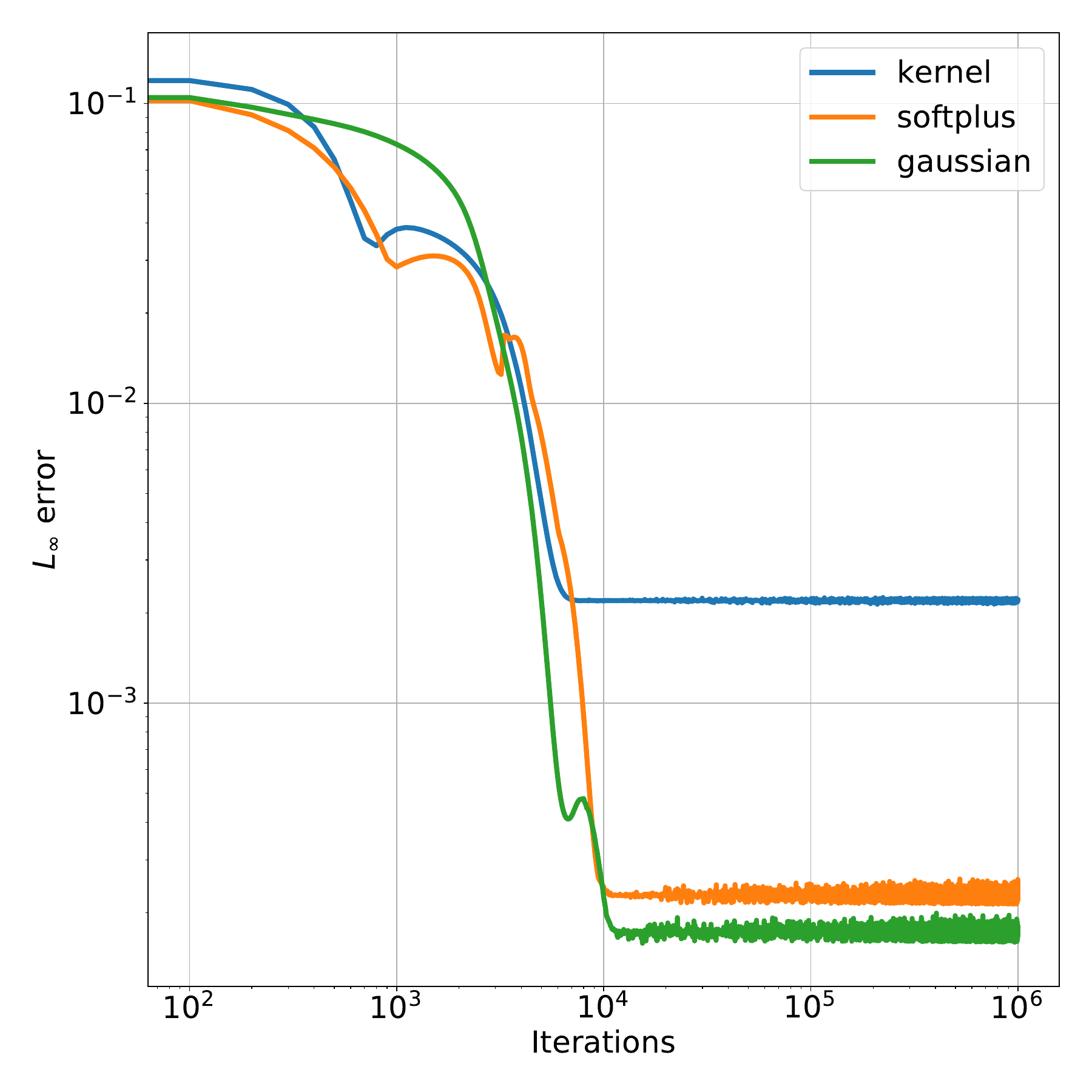}
\caption{$n=1$}
\label{fig:ode1_n_1_error}
\end{subfigure}
~
\begin{subfigure}{0.3\textwidth}
\includegraphics[width=\textwidth]{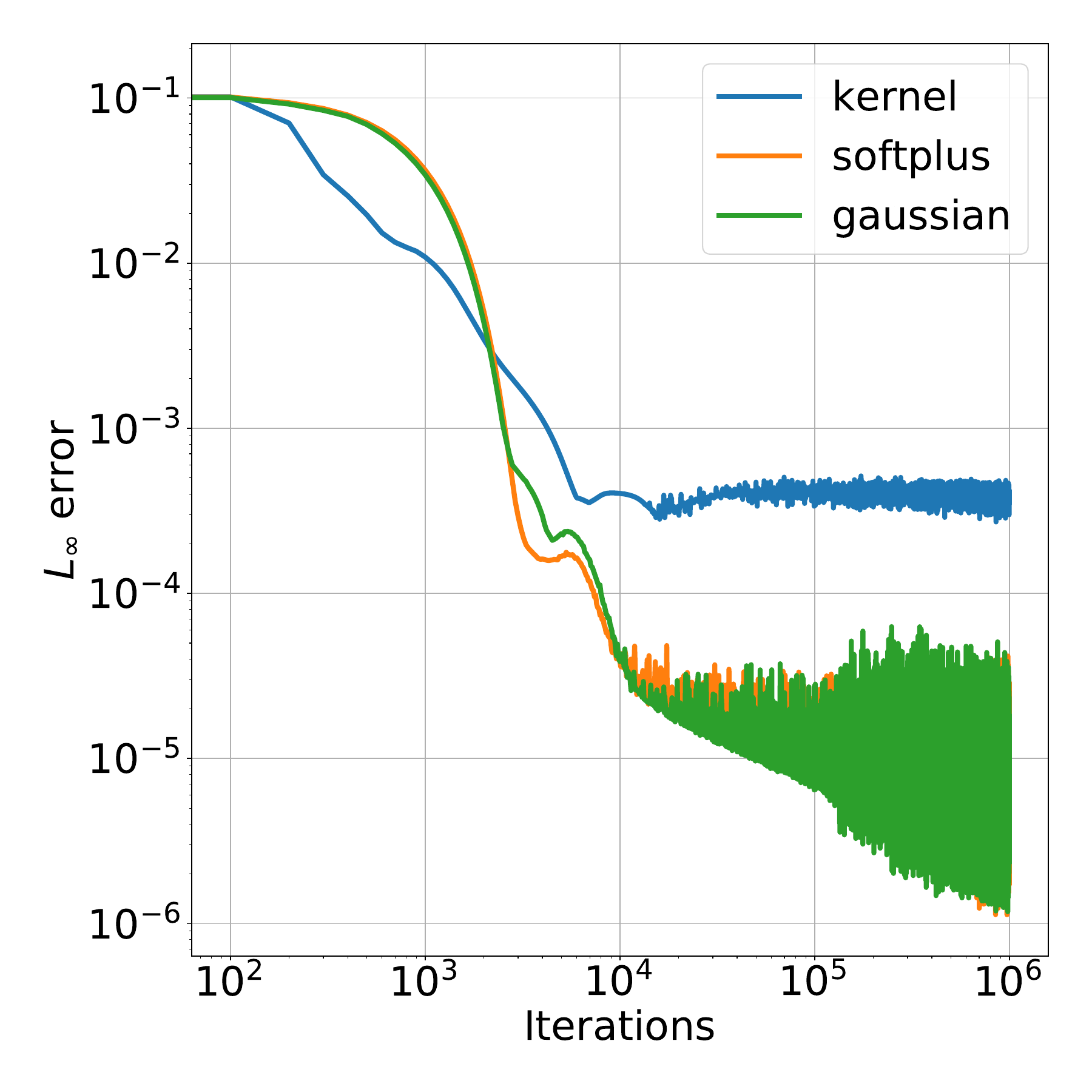}
\caption{$n=3$}
\label{fig:ode1_n_3_error}
\end{subfigure}
~
\begin{subfigure}{0.3\textwidth}
\includegraphics[width=\textwidth]{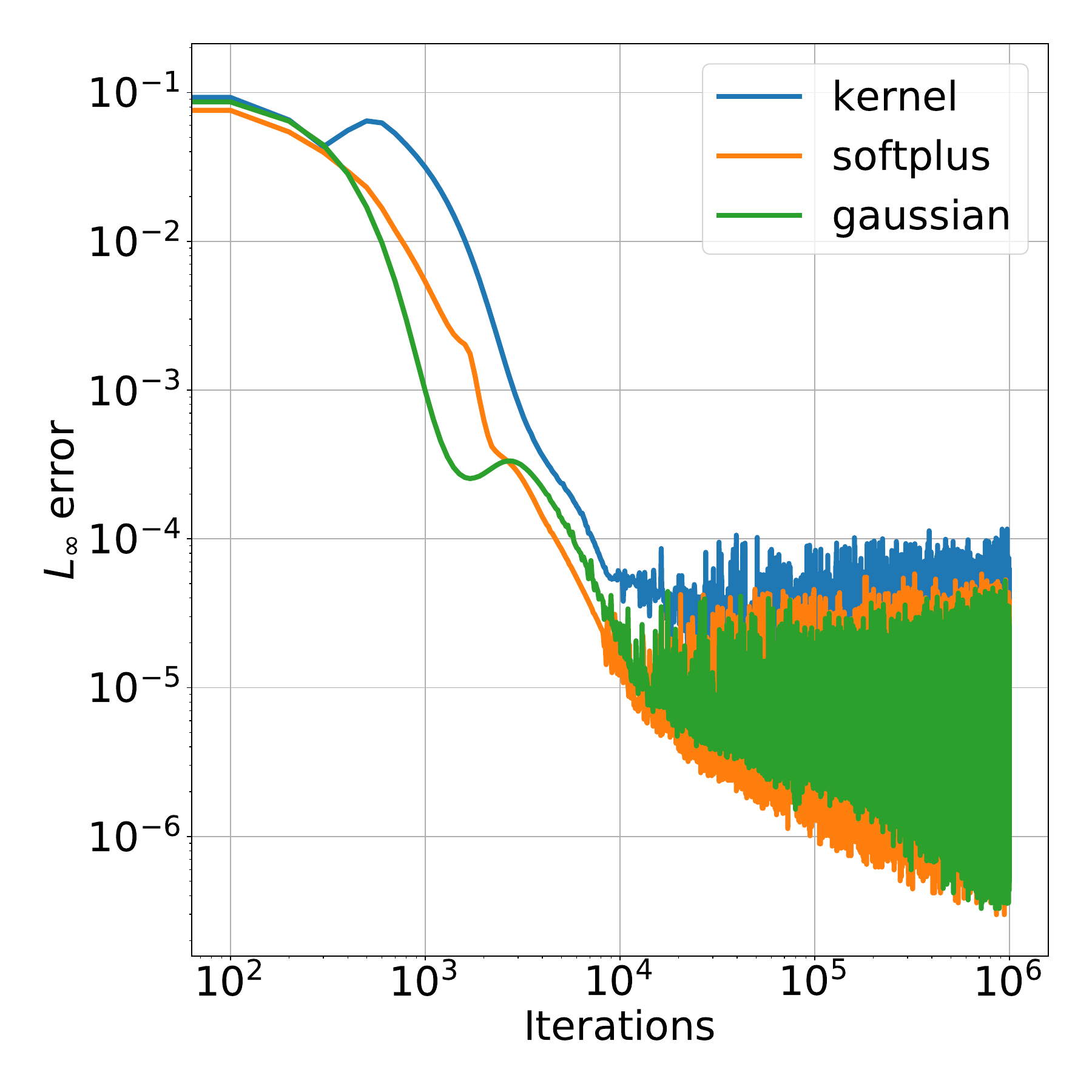}
\caption{$n=7$}
\label{fig:ode1_n_7_error}
\end{subfigure}
\caption{Evolution of $L_\infty$ error during training for ODE \eqref{eq:ode1} using SPINN with different kernels for $n=1,3,7$ nodes. \rb{The simulations were performed for $10^6$ training steps to study the long term error behavior.}}
\label{fig:spinn_ode_1_errors}
\end{figure}

\rr{We discuss the accuracy of the SPINN algorithm for this problem next. The evolution of the $L_\infty$ error during the training of the SPINN model with $n=3$ internal nodes and different kernels is shown in Figure~\ref{fig:spinn_ode_1_errors}.} \rb{All simulations shown in this figure were carried out for $10^5$ training steps to study the long term behavior of the  $L_\infty$ errors in each of the cases. We find that the Gaussian kernel has the best performance in general, with the softplus hat kernel having a similar performance. The neural network kernel, however, does not perform well for smaller number of internal nodes. We note that the errors for the $n=3$ and $n=7$ cases dip below $10^{-4}$, which is satisfactory considering the fact that these simulations employ singe precision floating point numbers.}

For all the kernels, it is observed that the error plot shows \new{three} distinct regimes. Focusing on Figure~\ref{fig:ode1_n_1_error}, for instance, an examination of the intermediate solutions during the iteration reveals that the first phase characterized by high loss and slow convergence correlates with the SPINN model minimizing the interior loss. The second phase in Figure~\ref{fig:ode1_n_1} characterized by a rapid drop in the error correlates with the SPINN model learning the boundary conditions of the ODE. \new{The final phase is characterized by the error saturating and exhibiting oscillations due to the inability of the algorithm to perform better given the various aspects of the discretization.}

\subsubsection{ODE with large gradients}
\begin{figure}
\centering
\begin{subfigure}{0.3\textwidth}
\includegraphics[width=\textwidth]{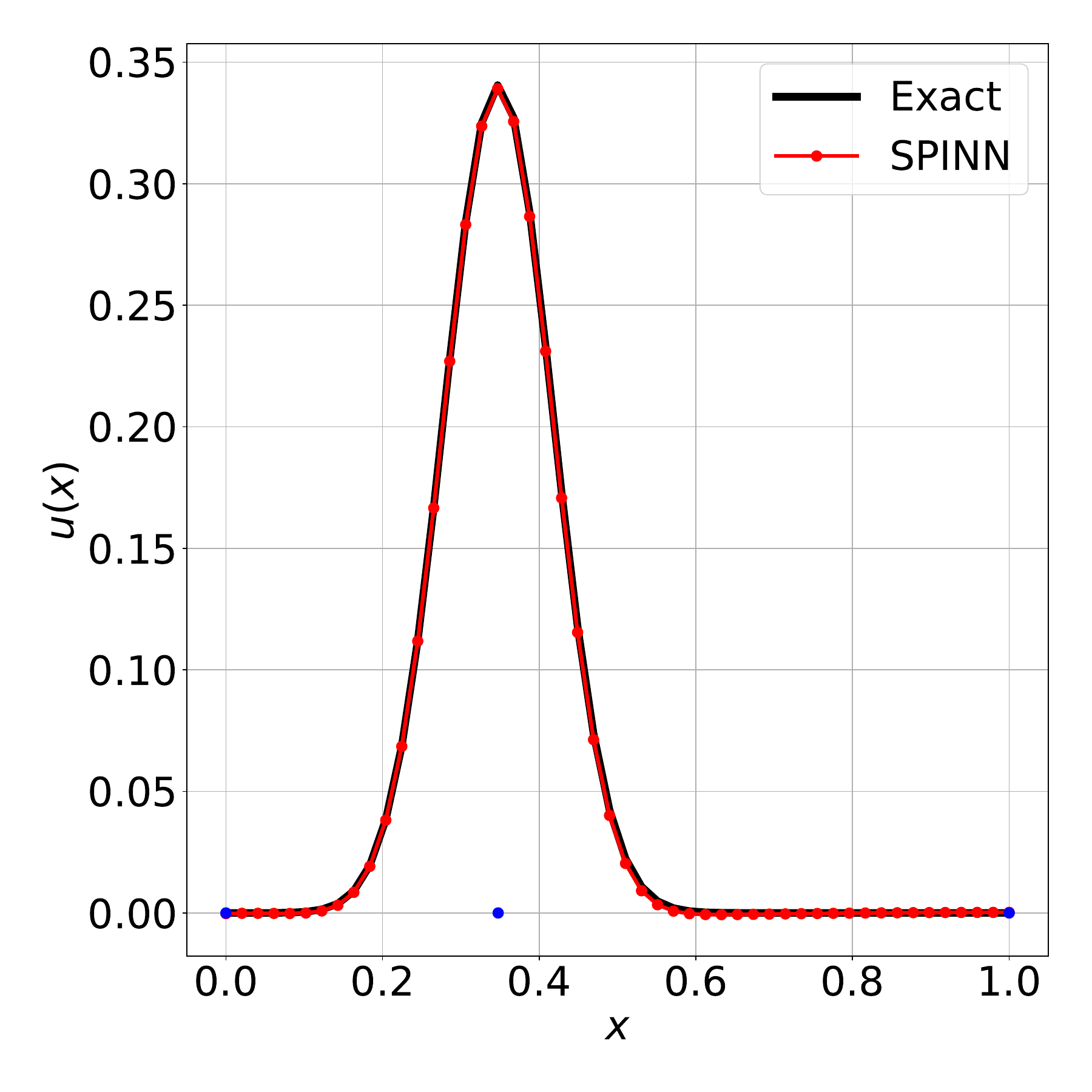}
\caption{$n = 1$}
\label{fig:ode3_n_1}
\end{subfigure}
~
\begin{subfigure}{0.3\textwidth}
\includegraphics[width=\textwidth]{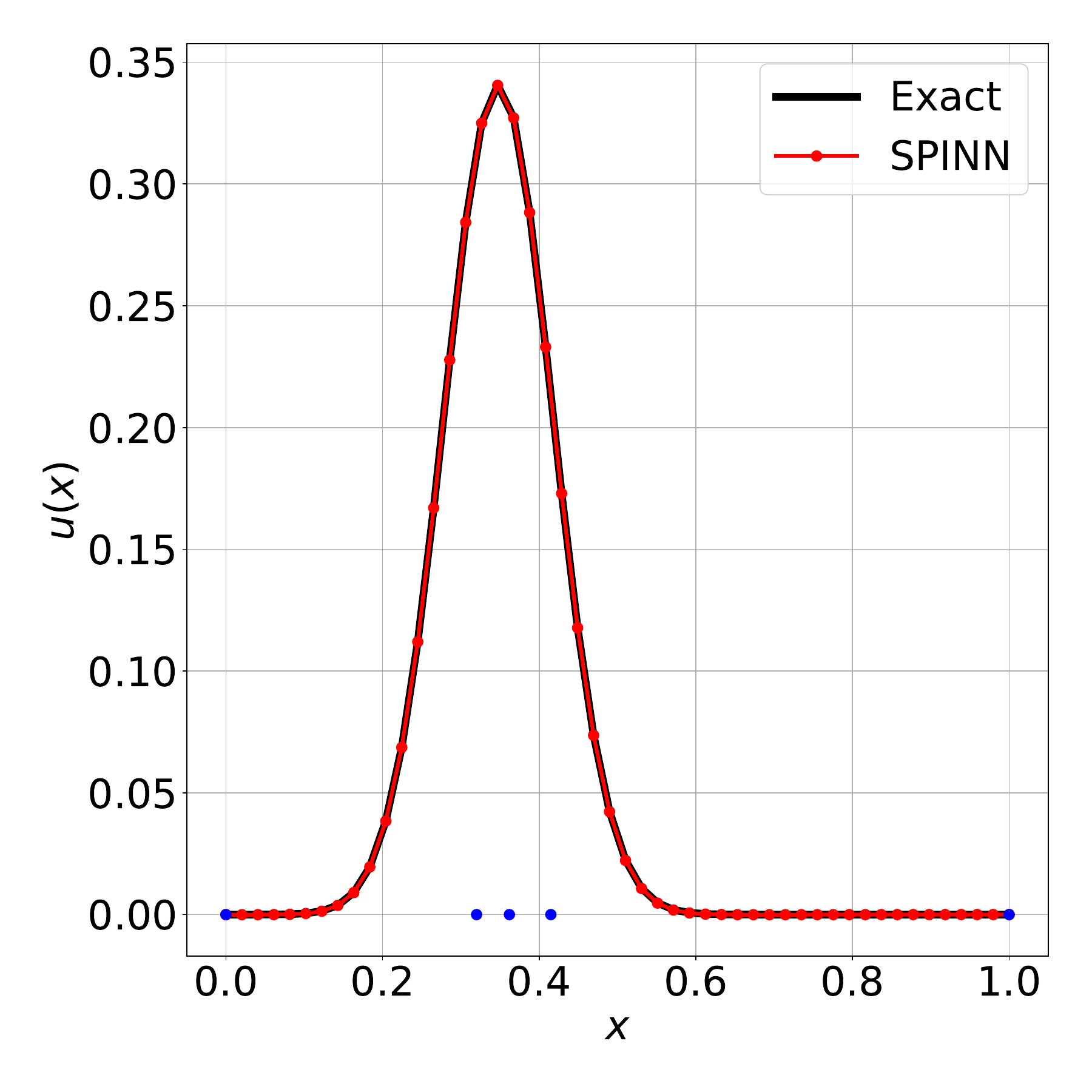}
\caption{$n = 3$}
\label{fig:ode3_n_3}
\end{subfigure}
~
\begin{subfigure}{0.3\textwidth}
\includegraphics[width=\textwidth]{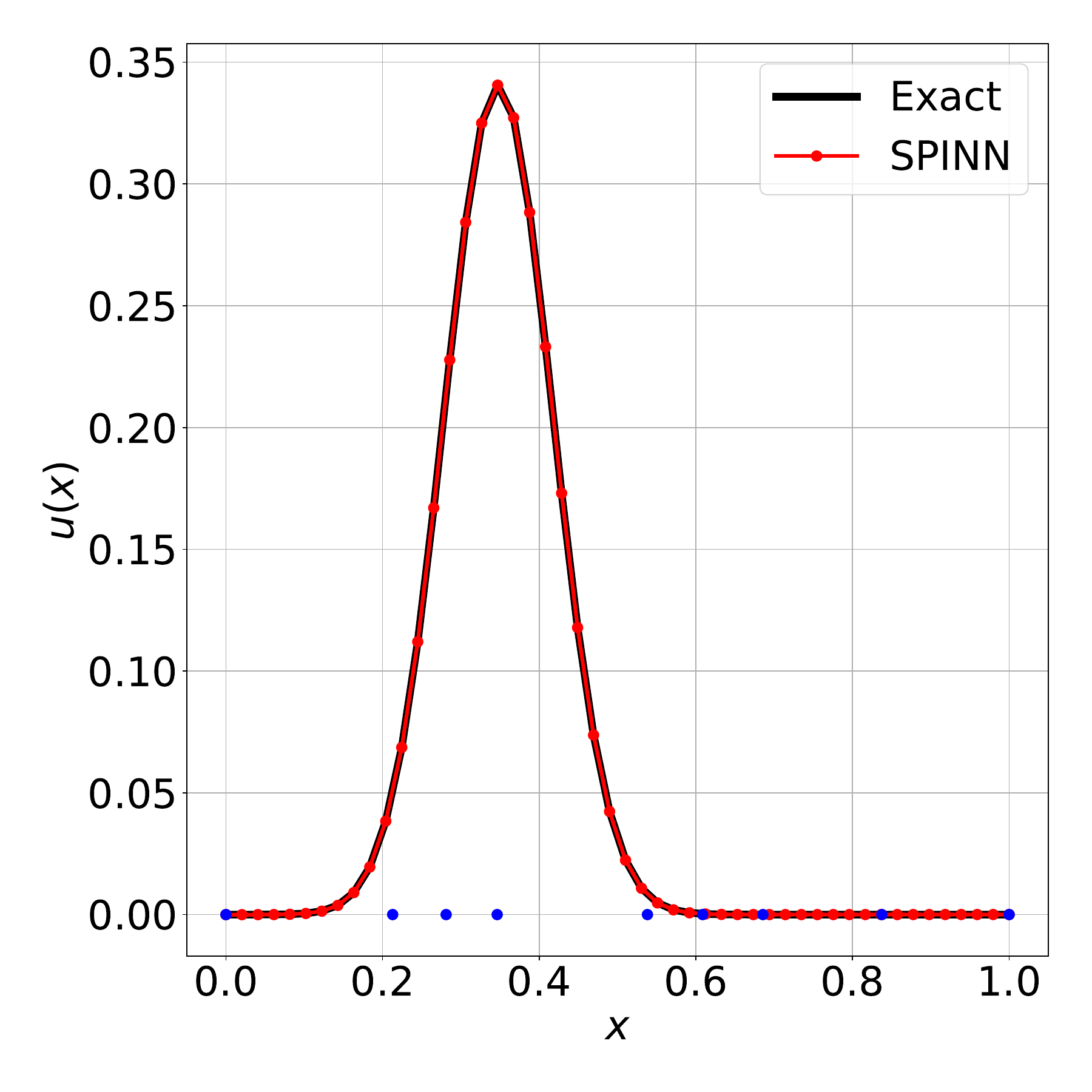}
\caption{$n = 7$}
\label{fig:ode3_n_7}
\end{subfigure}
\caption{Solution of ODE \eqref{eq:ode3} using SPINN with Gaussian kernel and $n=1,3,7$ interior nodes. The nodal positions learnt by SPINN are shown as blue circles along the $x$ axis.}
\label{fig:spinn_ode_3}
\end{figure}

\begin{figure}
\begin{subfigure}{0.48\textwidth}
\centering
\includegraphics[width=\textwidth]{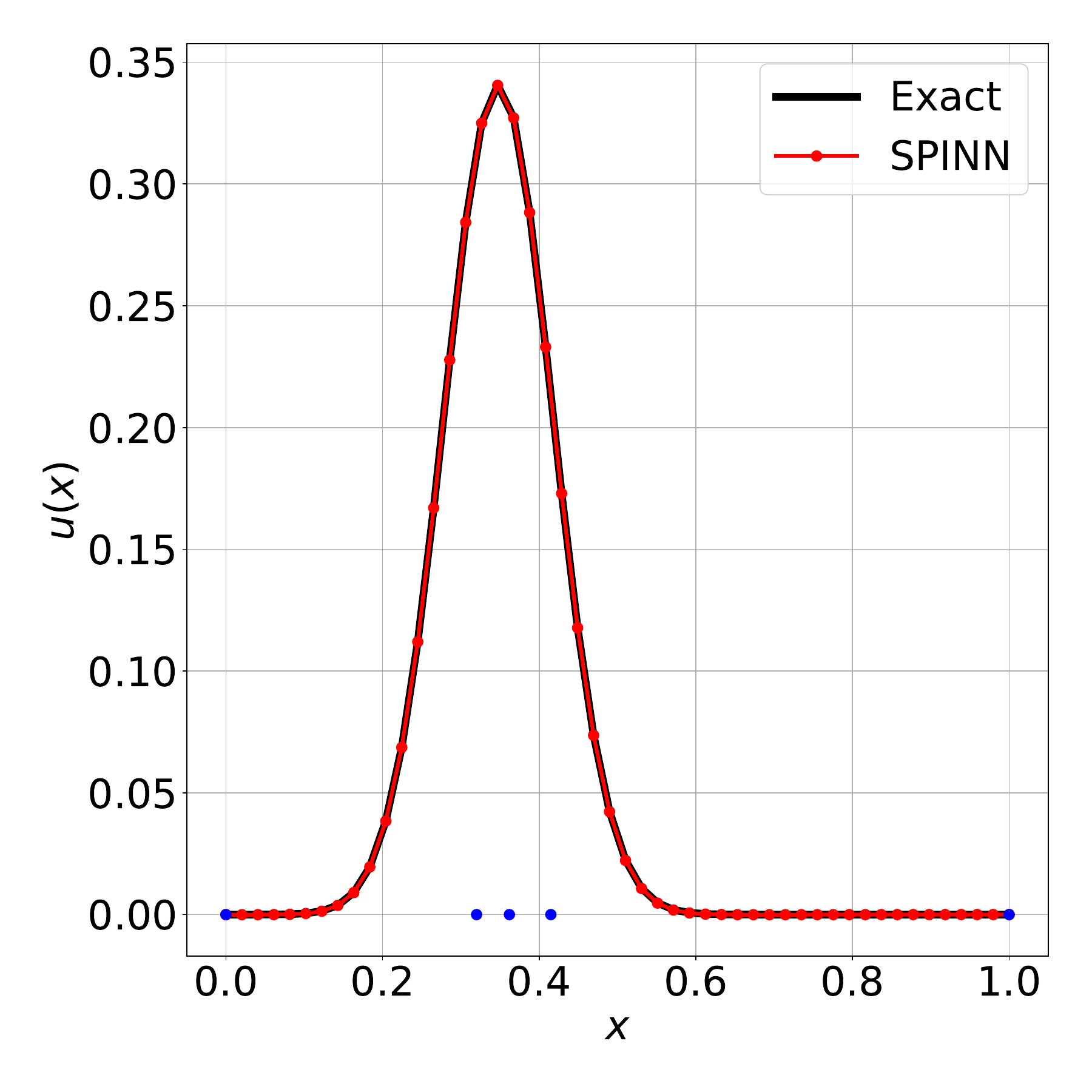}
\caption{Gaussian kernel}
\label{fig:ode3_gaussian_n_3}
\end{subfigure}
~
\begin{subfigure}{0.48\textwidth}
\centering
\includegraphics[width=\textwidth]{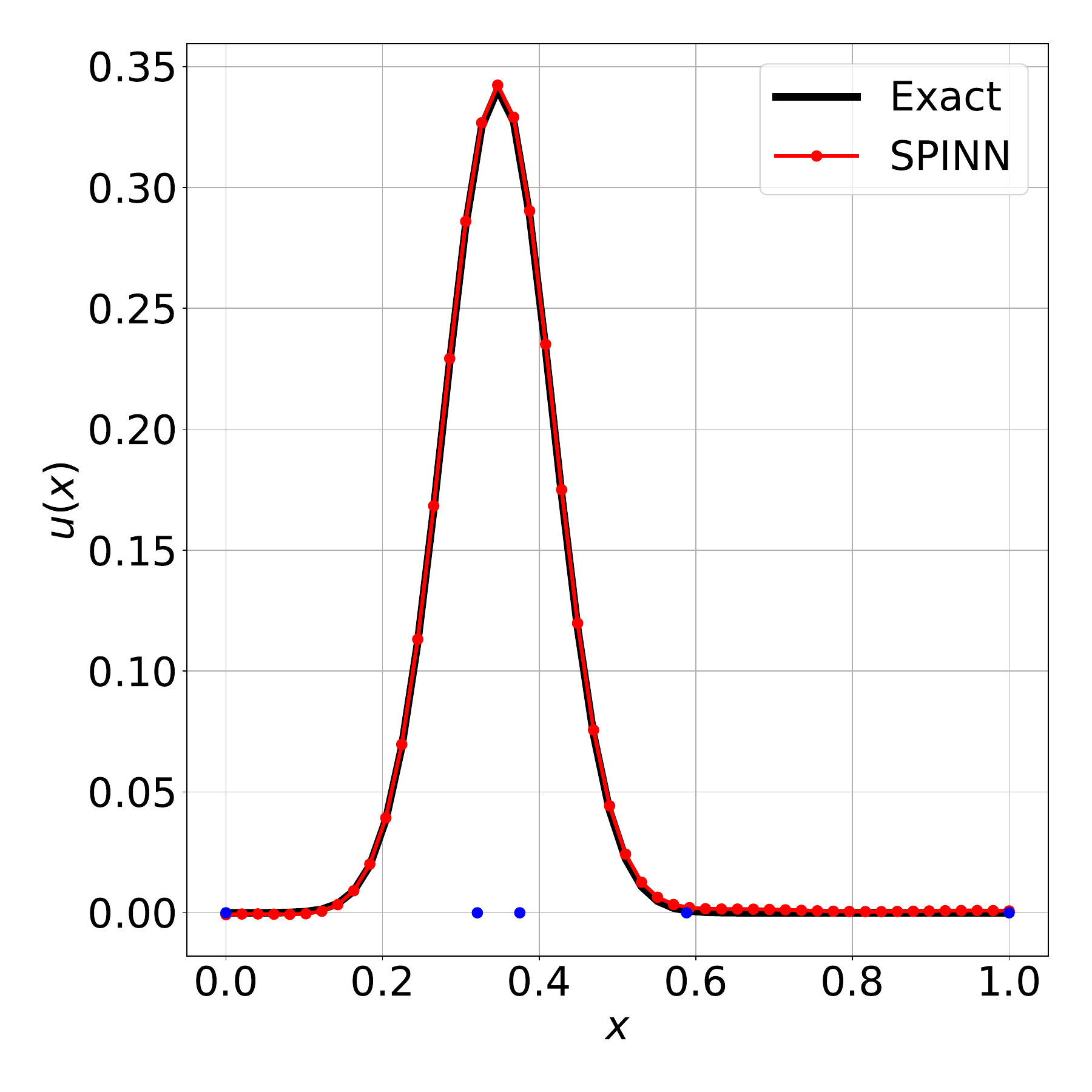}
\caption{Softplus hat kernel}
\label{fig:ode3_softplus_n_3}
\end{subfigure}

\begin{subfigure}{0.48\textwidth}
\centering
\includegraphics[width=\textwidth]{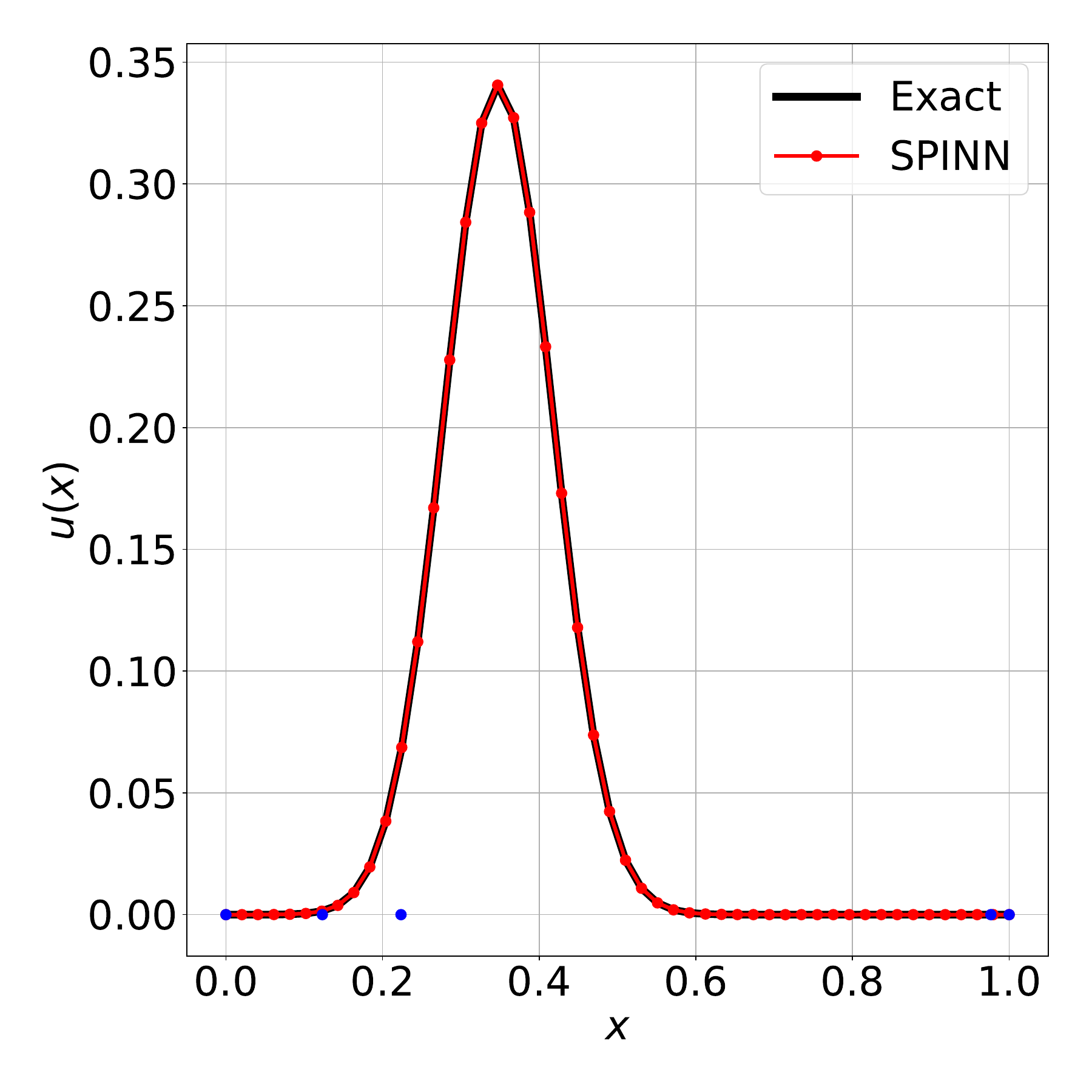}
\caption{Neural network kernel}
\label{fig:ode3_kernel_n_3}
\end{subfigure}
~
\begin{subfigure}{0.48\textwidth}
\centering
\includegraphics[width=\textwidth]{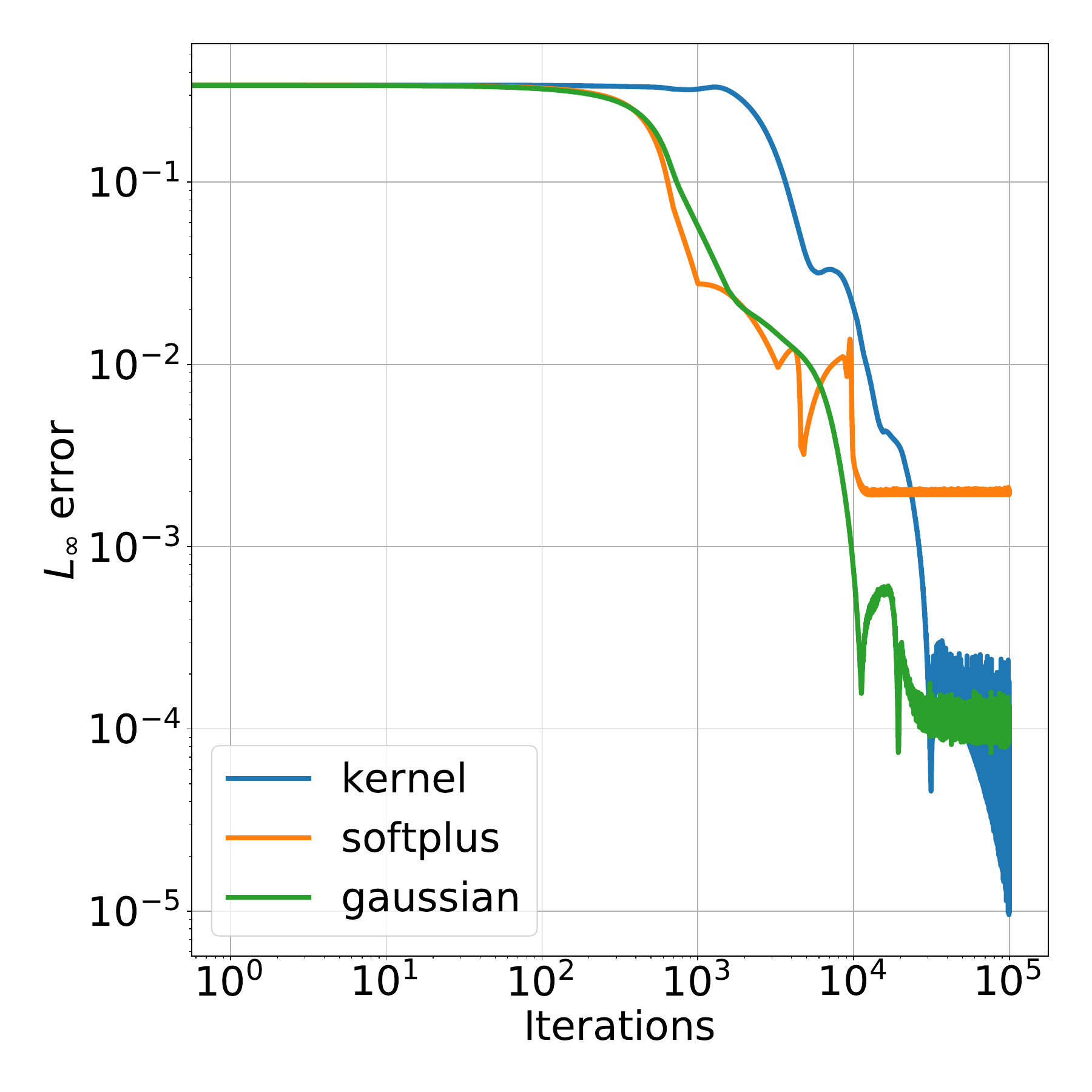}
\caption{$L_{\infty}$ error}
\label{fig:ode3_Linf_n_3_full}
\end{subfigure}
\caption{Solution of ODE \eqref{eq:ode3} with different kernels. $n=3$ internal nodes are used for all these simulations. The position of the nodes learnt by the SPINN algorithm are also shown as filled blue circles on the $x$ axis.  The evolution of the $L_\infty$ error for all three choices of kernels is also shown.}
\label{fig:ode3_kernel_error}
\end{figure}

\new{We consider next the ODE
\begin{equation} \label{eq:ode3}
\begin{split}
u''(x) + x(\exp (-(x - (1/3))^2/K) - \exp (-4/9K)) = 0, &\quad x\in (0,1),\\
u(0) = u(1) &= 0.
\end{split}
\end{equation}
In ODE \eqref{eq:ode3}, we choose $K = 0.01$.  The exact solution for the ODE \eqref{eq:ode3} is
\begin{equation} \label{eq:ode3_exact}
u(x) = -\frac{1}{K^2}\left(K \left(\frac{4}{3} - 6x\right) + 4x\left(x - \frac{1}{3}\right)^2\right)\exp\left(-\frac{1}{K}\left(x - \frac{1}{3}\right)^2\right).
\end{equation}
The exact solution \eqref{eq:ode3_exact} informs us that the solution has sharp gradients, with the gradient being sharper the lower the value of $K$. We chose this example to study the ability of SPINN to capture large gradients in a simplified setting. The solution obtained with SPINN using $n=1,3,7$ nodes is shown in Figure~\ref{fig:spinn_ode_3}.} \rb{For all the simulations, $n_S = 20n$ interior sampling points and two boundary sampling points are used. As in the previous case, both the nodes and sampling points are distributed uniformly initially. A learning rate of $10^{-4}$ is used and the network is trained for $10^5$ iterations.} \rr{The solution corresponding to the choice of different kernels with $n = 3$ interior nodes, along with the evolution of the $L_\infty$ errors is shown in Figure~\ref{fig:ode3_kernel_error}.} \new{As in the case of the previous example, the clustering of the nodes near high gradient regions of the solution and the three phases of the error plots are clearly seen.}

\subsubsection{Treatment of Neumann boundary conditions}
\begin{figure}
\centering
\begin{subfigure}{0.3\textwidth}
\includegraphics[width=\textwidth]{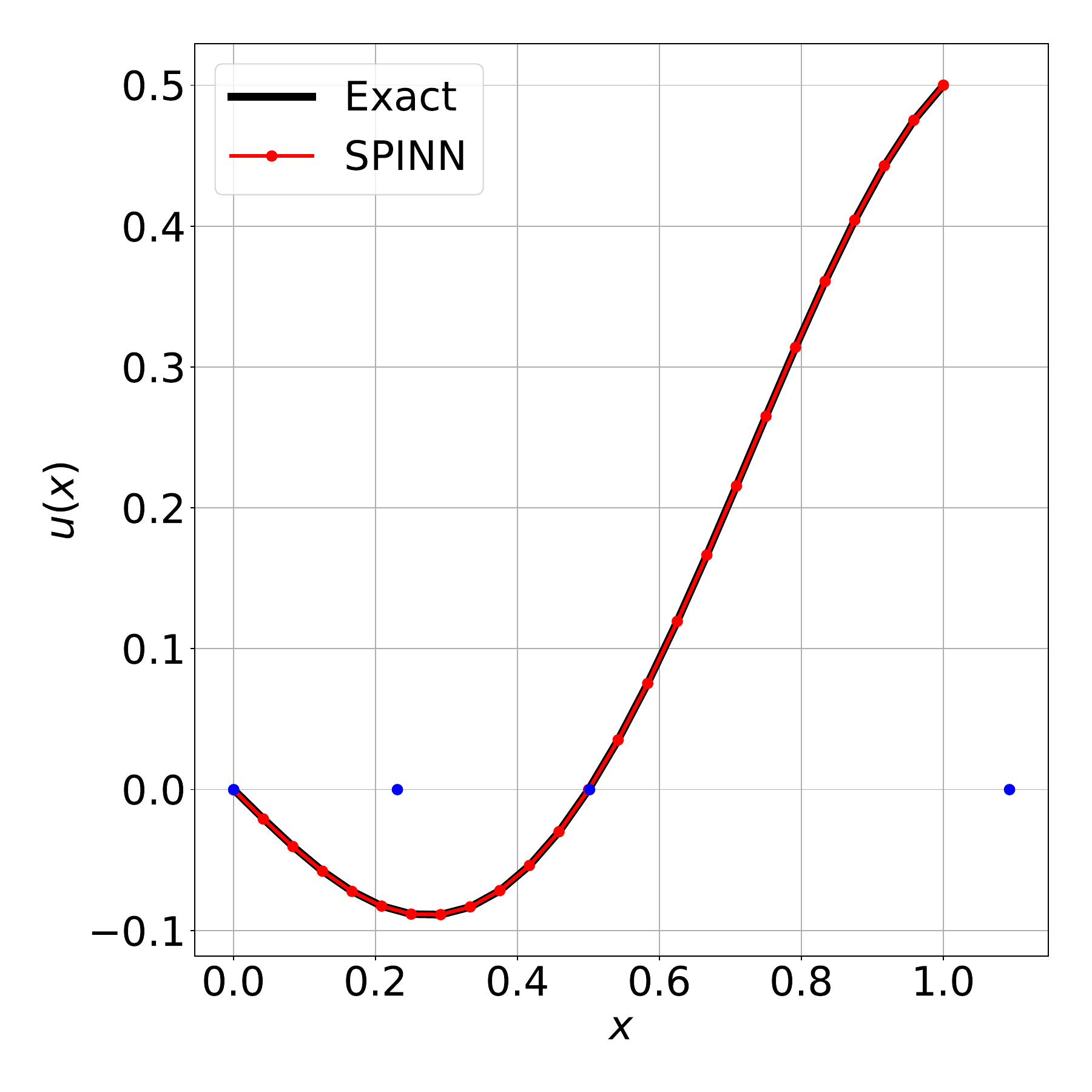}
\caption{$n = 3$}
\label{fig:ode2_n_3}
\end{subfigure}
~
\begin{subfigure}{0.3\textwidth}
\includegraphics[width=\textwidth]{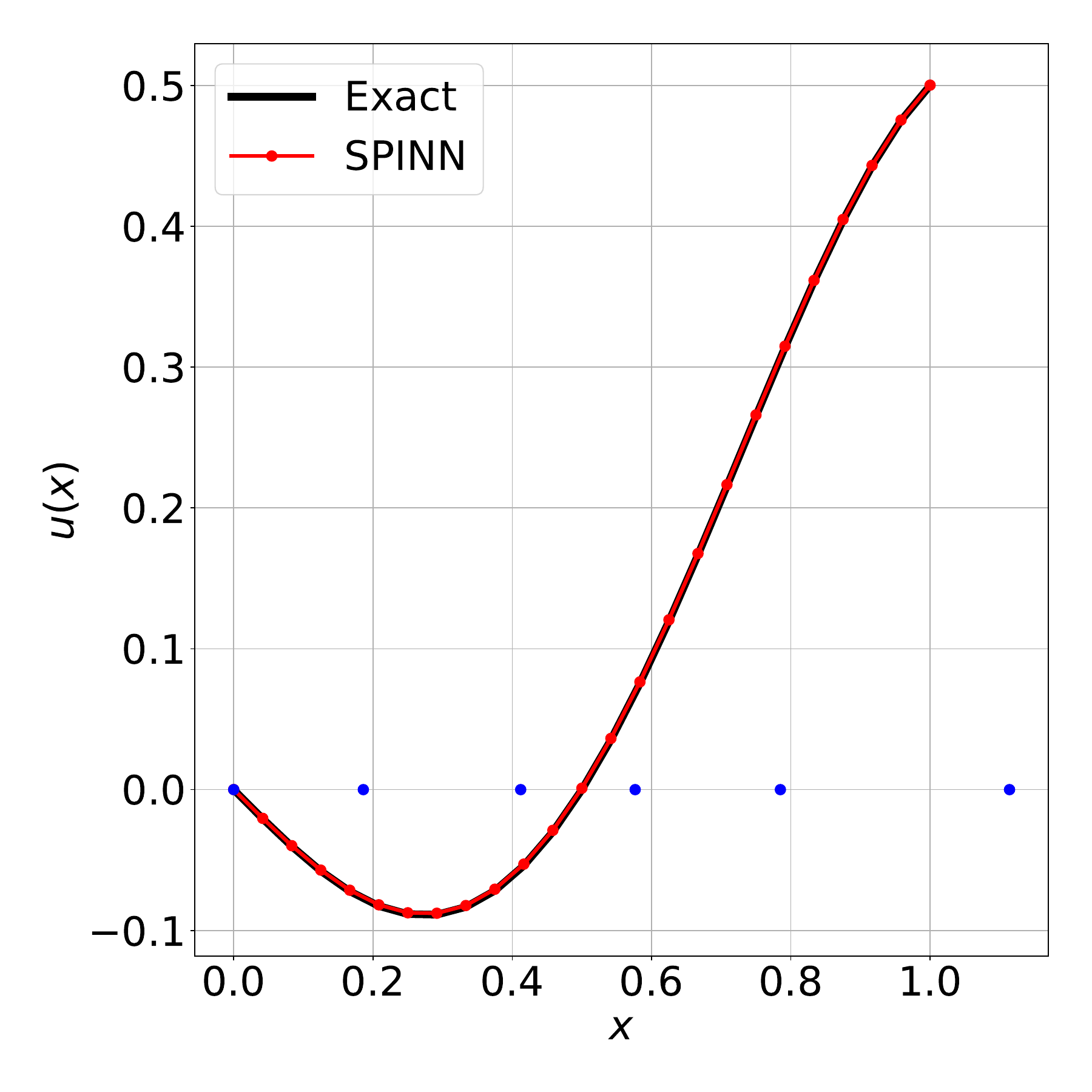}
\caption{$n = 5$}
\label{fig:ode2_n_5}
\end{subfigure}
~
\begin{subfigure}{0.3\textwidth}
\includegraphics[width=\textwidth]{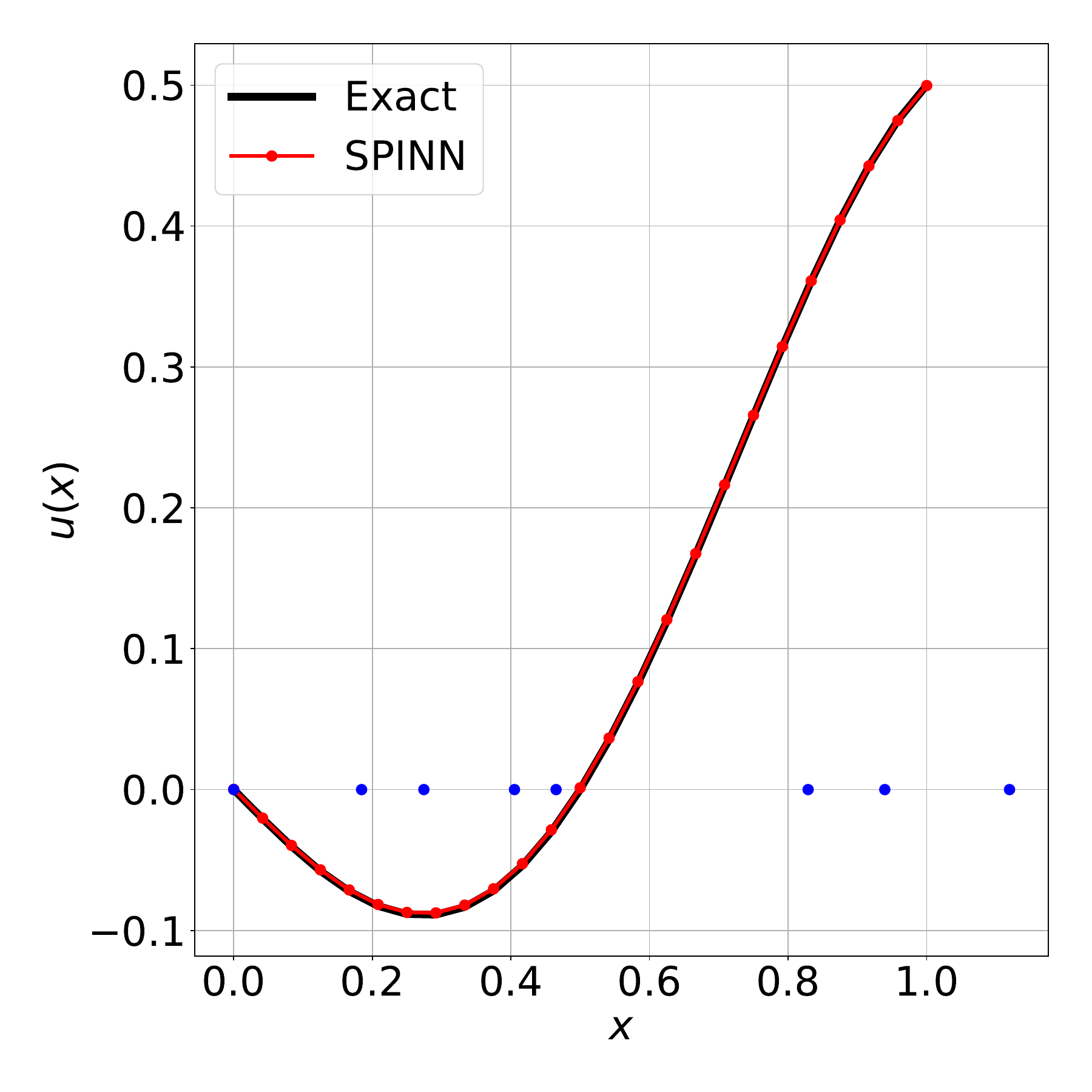}
\caption{$n = 7$}
\label{fig:ode2_n_7}
\end{subfigure}
\caption{Solution of ODE \eqref{eq:ode2} using SPINN with Gaussian kernel and $n = 3, 5, 7$, interior nodes. No fixed node is placed on the Neumann boundary. The nodal positions learnt by SPINN are shown as blue circles along the $x$ axis.}
\label{fig:spinn_ode2}
\end{figure}

We study next the following ODE with both Dirichlet and Neumann boundary conditions:
\begin{equation} \label{eq:ode2}
\begin{split}
u''(x) + \pi^2 u(x) = \pi \sin \pi x, &\quad x \in (0,1),\\
u(0) = 0, &\quad u'(1) = \frac{1}{2}.
\end{split}
\end{equation}
The exact solution to ODE \eqref{eq:ode2} is
\begin{equation} \label{eq:ode2_exact}
u(x) = -\frac{1}{2}x\cos \pi x.
\end{equation}
\new{The solution computed by SPINN using Gaussian kernel and $n=3,5,7$ internal nodes is shown in Figure~\ref{fig:spinn_ode2}.} \rb{In each case, $n_S = 15n$ internal sampling points and two boundary sampling points were used. Both the nodes and samples were distributed uniformly initially.} To handle the Neumann boundary condition at $x=1$, no fixed node is used there. \new{The reason for this is that the kernels employed in this work are symmetric in nature and hence have vanishing gradients at their centers. This in turn leads to an infinite indeterminacy if a fixed node is placed on a Neumann boundary.} The interior nodes are free to move outside the domain to accommodate the Neumann condition. This is indeed seen in the SPINN solution shown in Figure~\ref{fig:spinn_ode2}. \rb{The solutions shown in Figure~\ref{fig:spinn_ode2} were computed using random sampling with sampling ratio $f=0.2$. The learning rate was chosen as $10^{-3}$. Unlike the previous simulations which were run for a fixed number of training steps, the simulations shown in Figure~\ref{fig:spinn_ode2} were terminated once the $L_\infty$ error reached a threshold value of $10^{-3}$, or when the number of iterations exceeded $10^6$.} \new{As can be seen from Figure~\ref{fig:spinn_ode2}, the SPINN algorithm is able to handle Neumann boundary conditions effectively.}

\begin{figure}
\begin{subfigure}{0.32\textwidth}
\centering
\includegraphics[width=\textwidth]{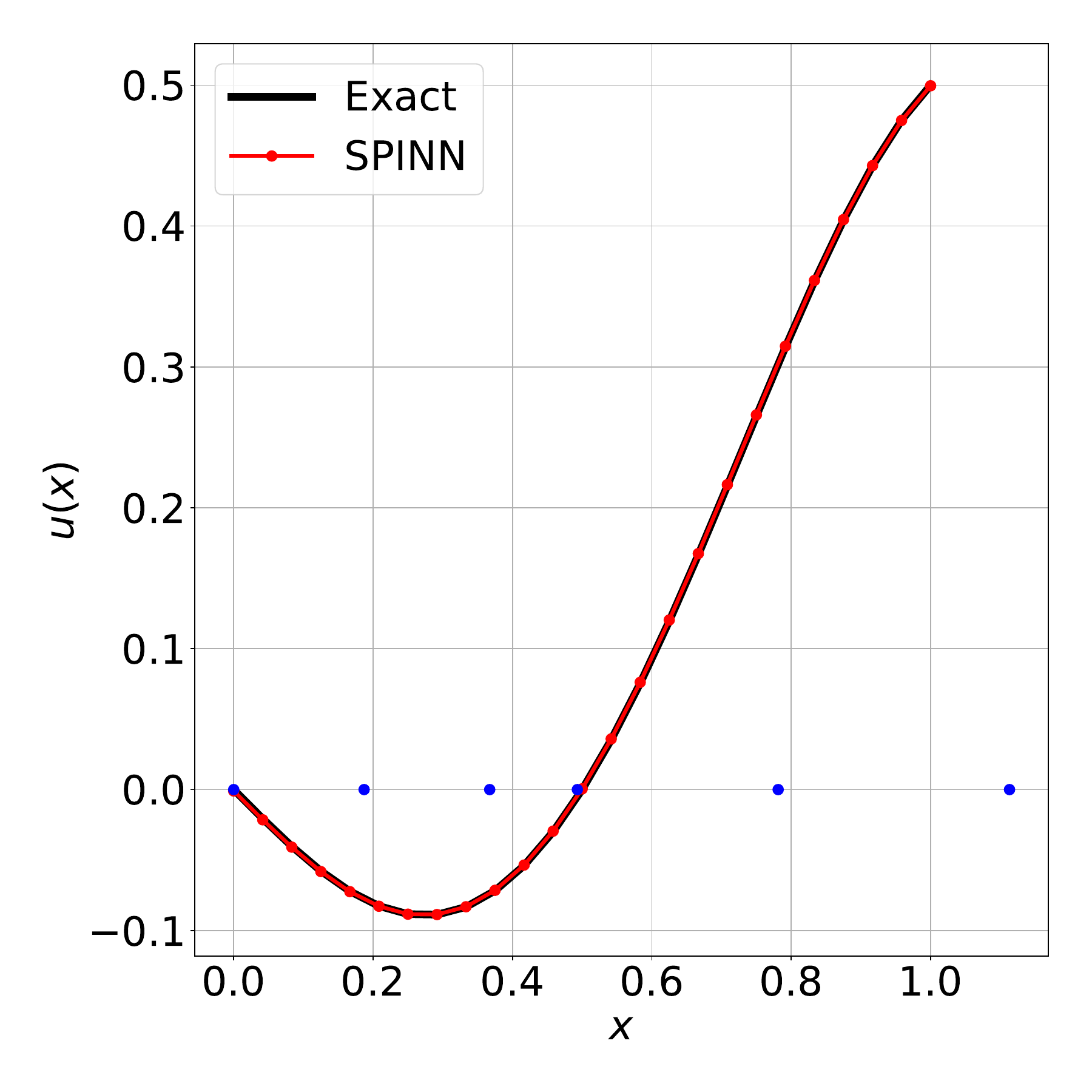}
\caption{$f = 0.1$}
\label{fig:ode2_n_5_f_0p1}
\end{subfigure}
~
\begin{subfigure}{0.32\textwidth}
\centering
\includegraphics[width=\textwidth]{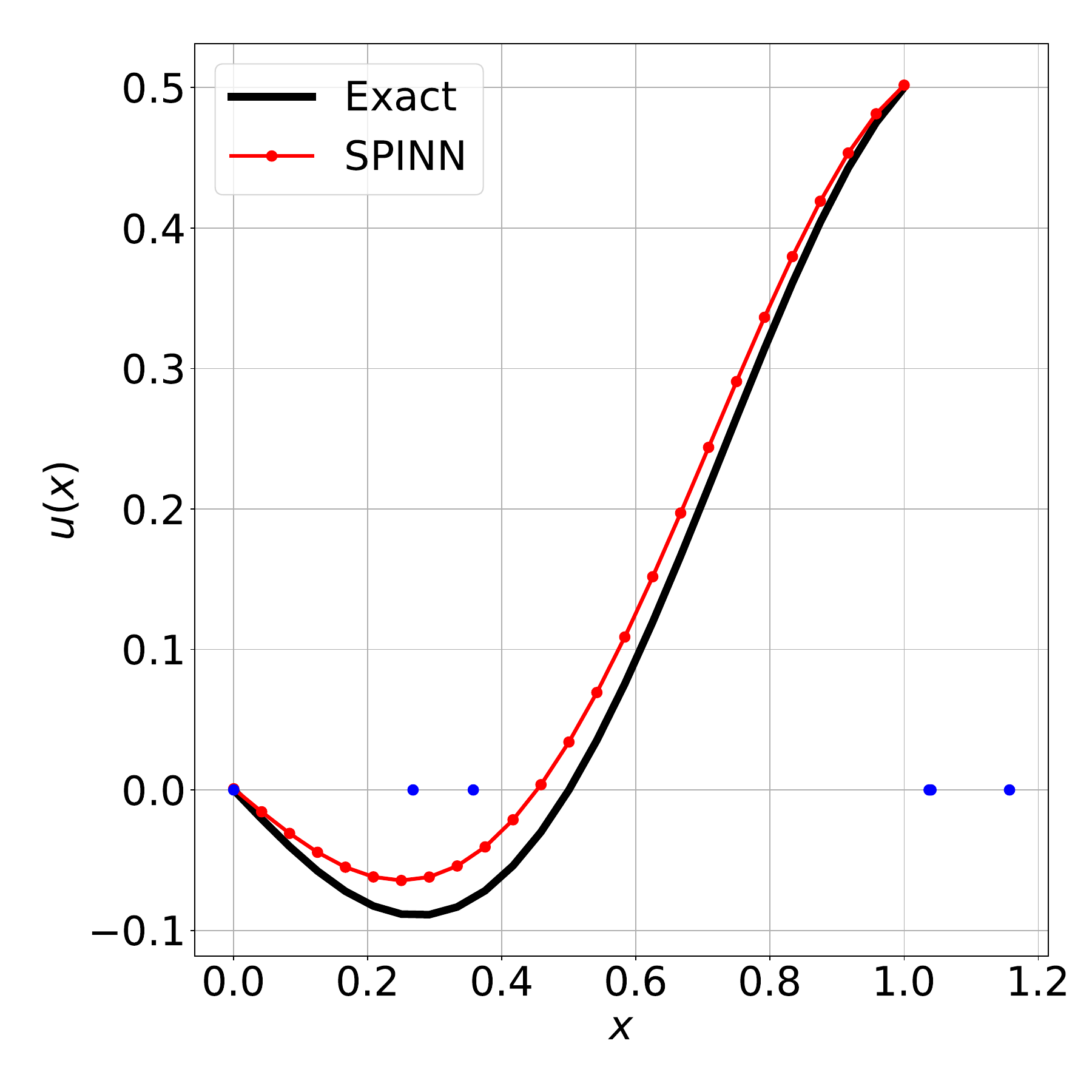}
\caption{$f=0.2$}
\label{fig:ode2_n_5_f_0p2_a}
\end{subfigure}
~
\begin{subfigure}{0.32\textwidth}
\centering
\includegraphics[width=\textwidth]{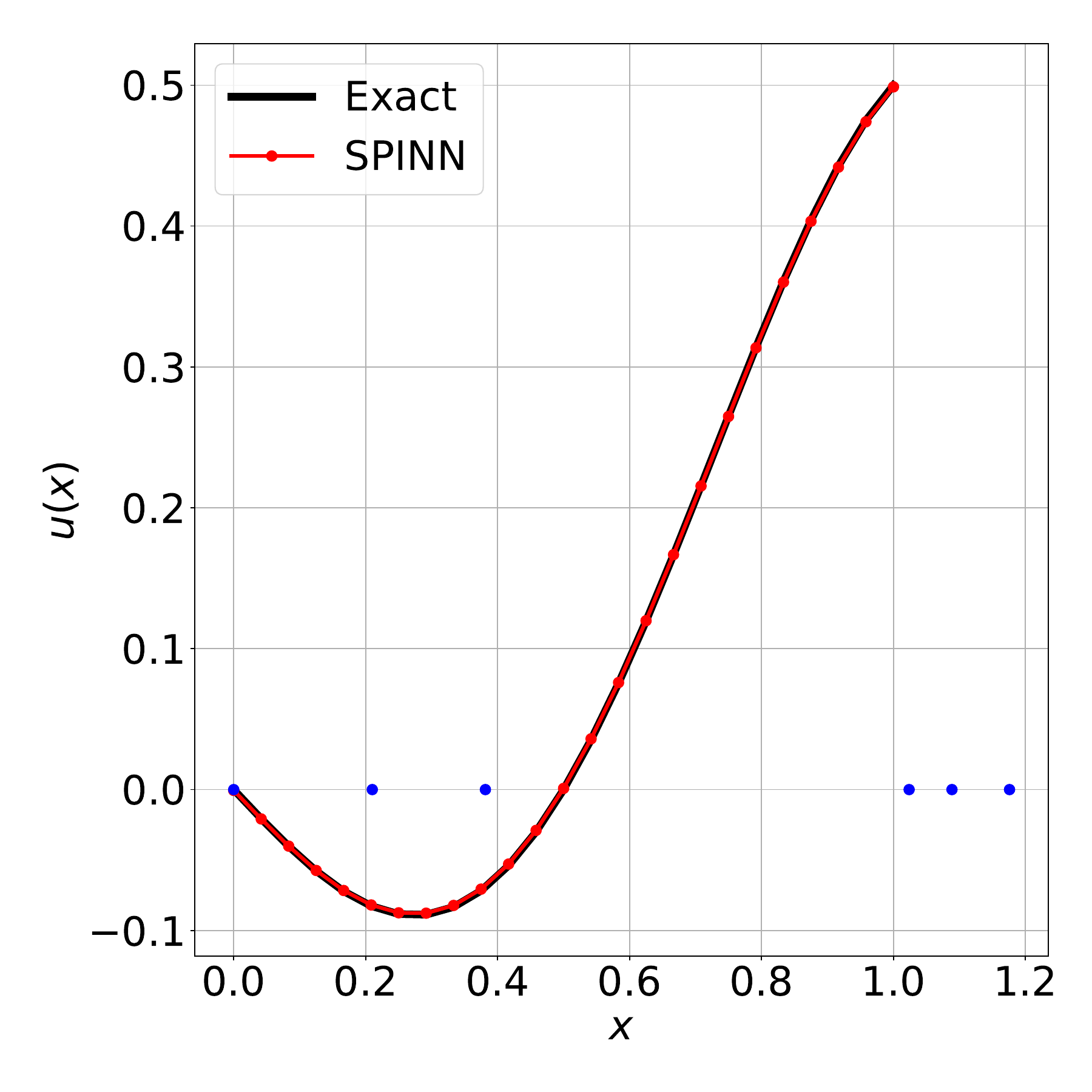}
\caption{$f=0.3$}
\label{fig:ode2_n_5_f_0p3}
\end{subfigure}
\begin{subfigure}{0.32\textwidth}
\centering
\includegraphics[width=\textwidth]{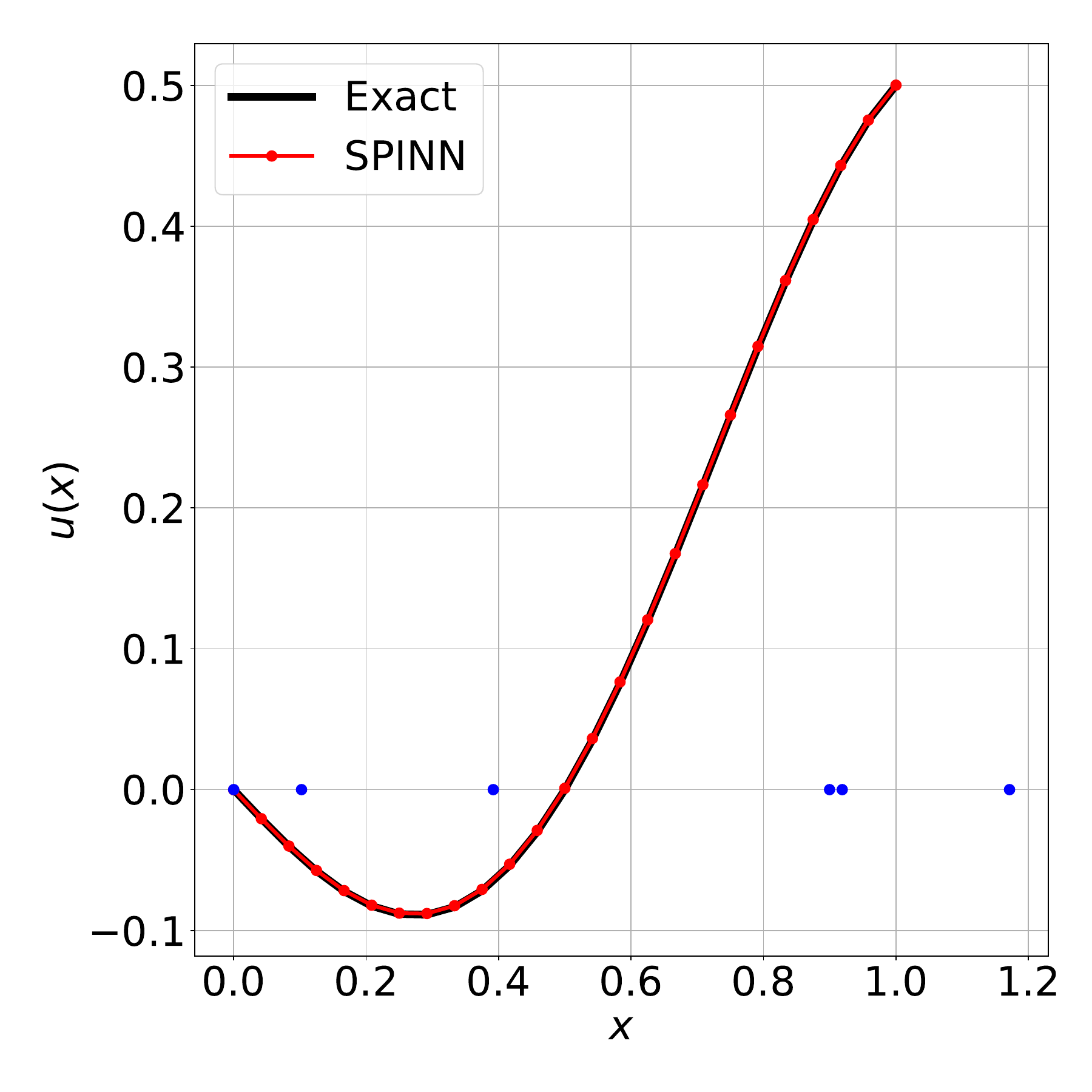}
\caption{$f=0.5$}
\label{fig:ode2_n_5_0p5}
\end{subfigure}
~
\begin{subfigure}{0.32\textwidth}
\centering
\includegraphics[width=\textwidth]{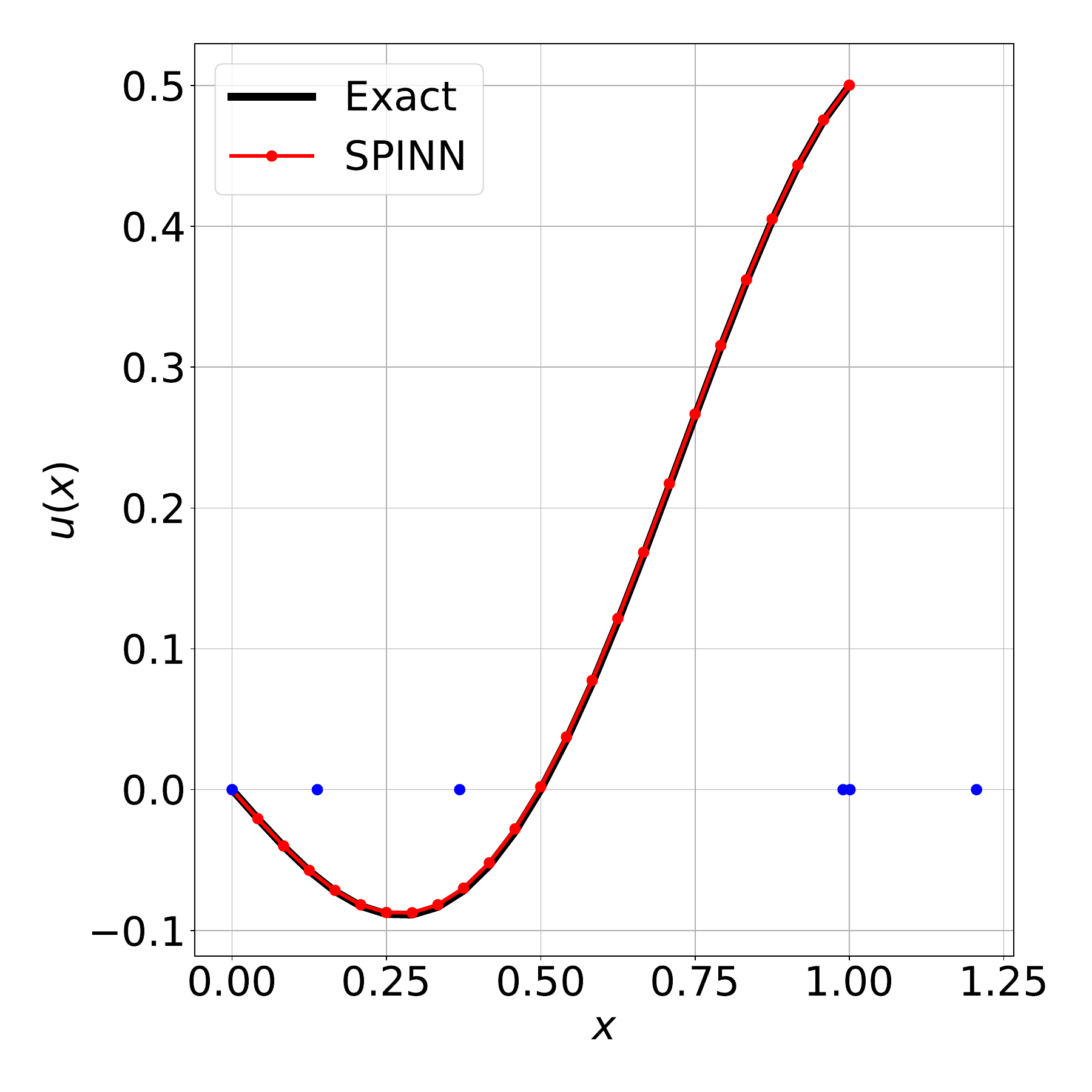}
\caption{$f=0.75$}
\label{fig:ode2_n_5_0p75}
\end{subfigure}
~
\begin{subfigure}{0.32\textwidth}
\centering
\includegraphics[width=\textwidth]{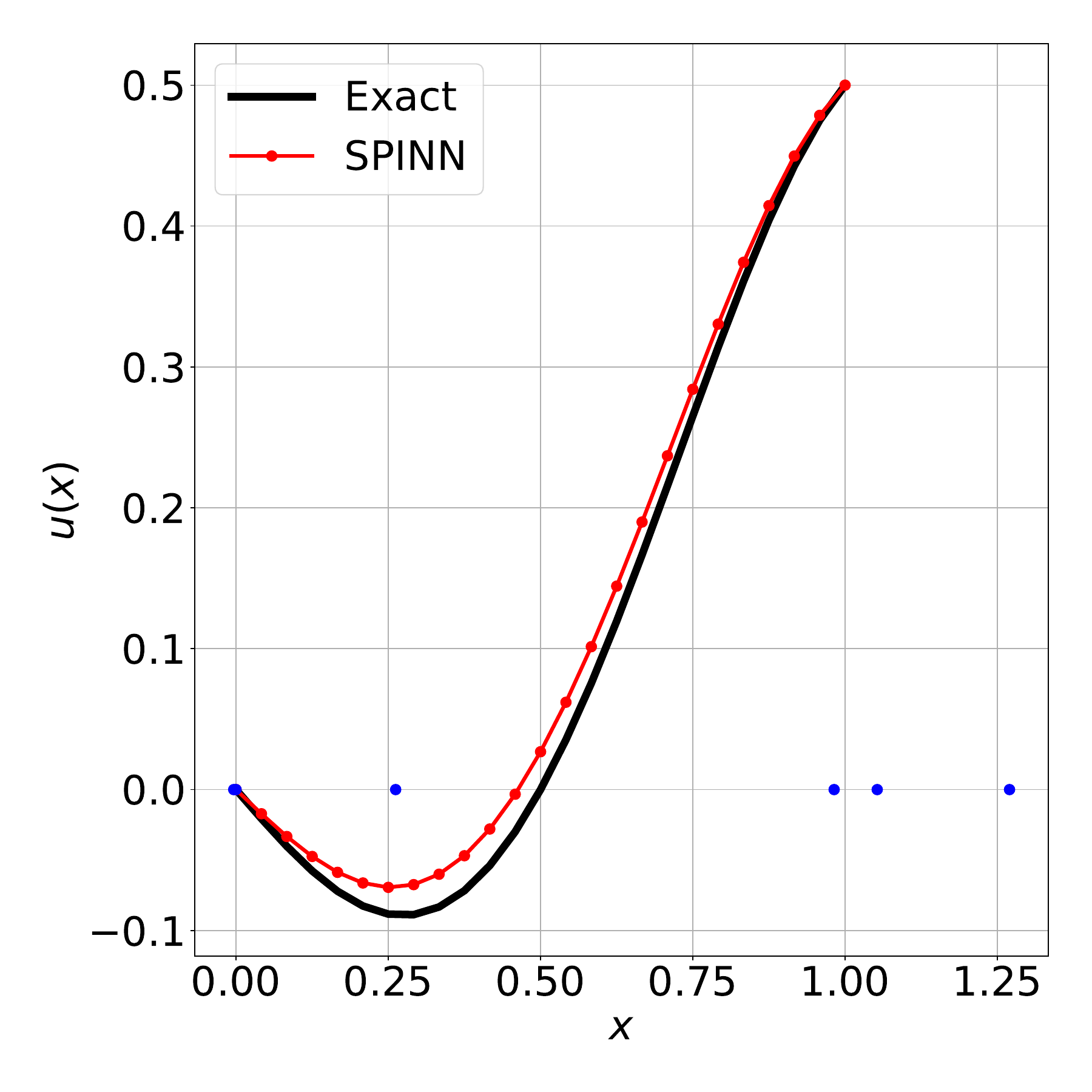}
\caption{$f=1$}
\label{fig:ode2_n_5_1}
\end{subfigure}
\caption{Effect of random sampling on the SPINN solution to ODE \eqref{eq:ode2} using Gaussian activation function. The number of nodes is fixed at $n=5$ and the total number of samples is fixed at $n_s = 15n$. The various graphs shown here correspond to different fractions, $f$, of the samples used to evaluate the loss at each iteration.}
\label{fig:spinn_ode2_rs}
\end{figure}

\begin{figure}
\begin{subfigure}{0.5\textwidth}
\centering
\includegraphics[width=\textwidth]{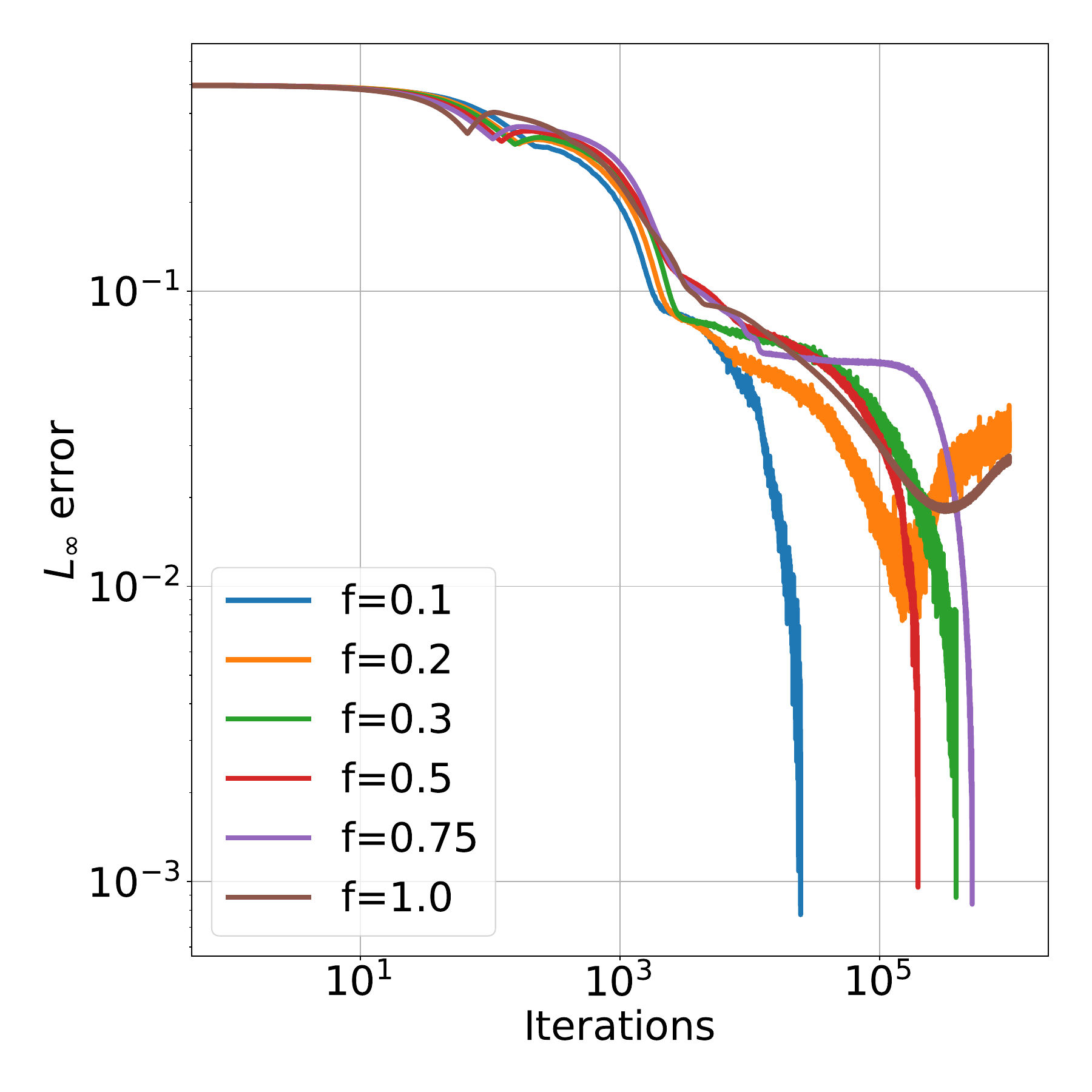}
\caption{$L_{\infty}$ error}
\label{fig:ode2_n_5_Linf}
\end{subfigure}
~
\begin{subfigure}{0.5\textwidth}
\centering
\includegraphics[width=\textwidth]{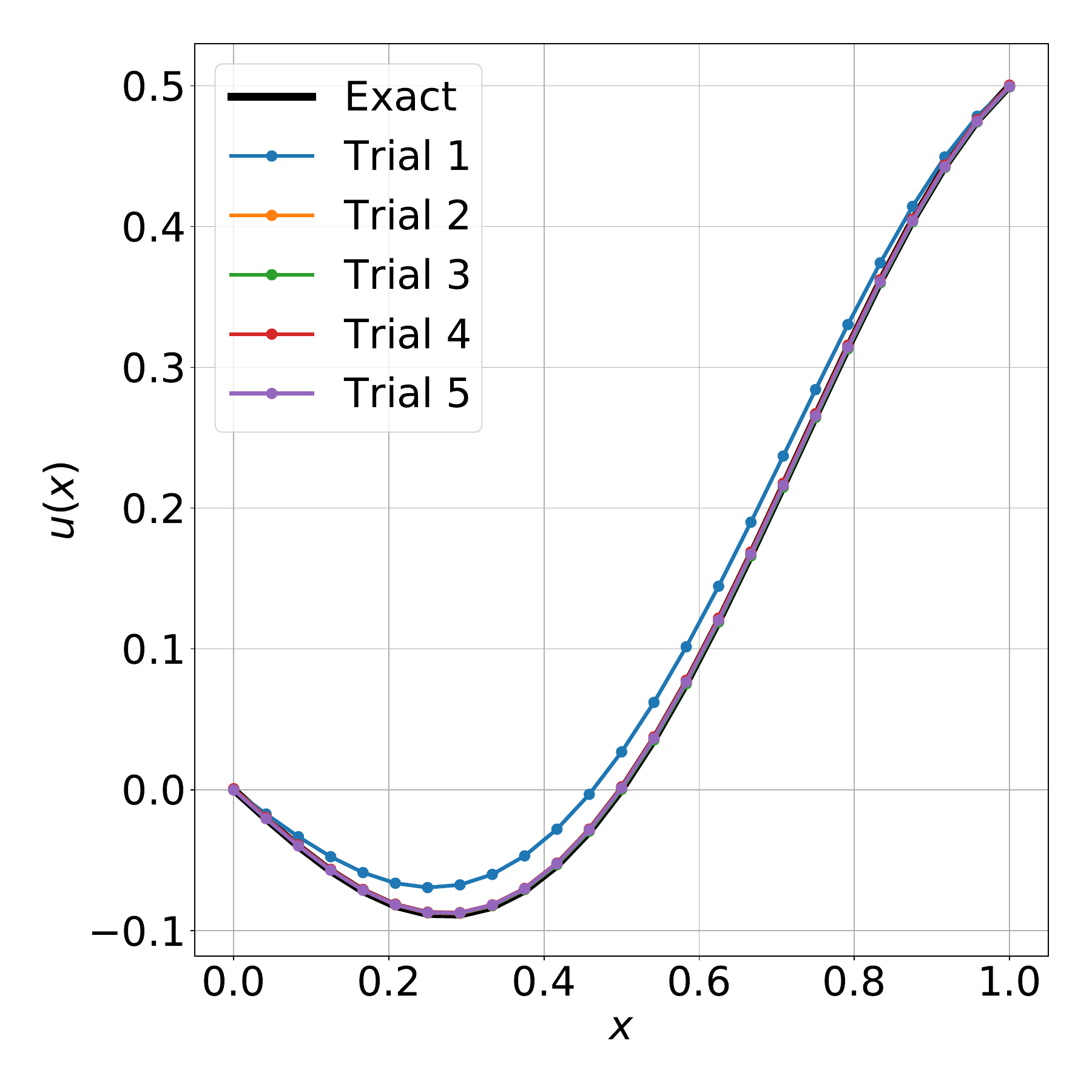}
\caption{Multiple runs for $f=0.2$}
\label{fig:ode2_n_5_rep}
\end{subfigure}
\caption{$L_\infty$ error for SPINN solution of ODE \eqref{eq:ode2} for various sampling ratios. The sampling ratio $f$ is varied, while the number of nodes $n=5$ and number of samples $n_S = 15n$ were fixed. The simulations were terminated when the errors reached a threshold value of $10^{-3}$, or if the number of iterations exceeded $10^6$. The figure on the right shows the results of multiple runs for the same set of parameters, with sampling ratio fixed at $f=0.2$.}
\label{fig:spinn_ode2_errors}
\end{figure}

\new{We present the effect of random sampling on the performance of SPINN in Figure~\ref{fig:spinn_ode2_rs}.} \rr{The corresponding errors for the case of $n=5$ interior nodes, with all other parameters fixed as mentioned above, is shown in Figure~\ref{fig:spinn_ode2_errors}}. We also illustrate the effect of sampling ratio $f$ on the convergence of the SPINN algorithm in Figure~\ref{fig:ode2_n_5_Linf}. \rr{The  $L_\infty$ error plot is shown in Figure~\ref{fig:ode2_n_5_Linf}. We note that since the algorithm is non-deterministic on account of the use of random sampling, there is a finite probability of obtaining a solution that gets stuck in a metastable state. To illustrate this we performed five different runs corresponding to a sampling ratio of $f=0.2$ - the results are shown in Figure~\ref{fig:ode2_n_5_rep}. It can be see that while one of the initial conditions of the five trials gets stuck in a metastable state, the remaining four converge to the true solution. Thus the algorithm is sensitive to a small extent on initial conditions when random sampling is used.}

\subsubsection{Variational SPINN}
\begin{figure}
\begin{subfigure}{0.5\textwidth}
\includegraphics[width=\textwidth]{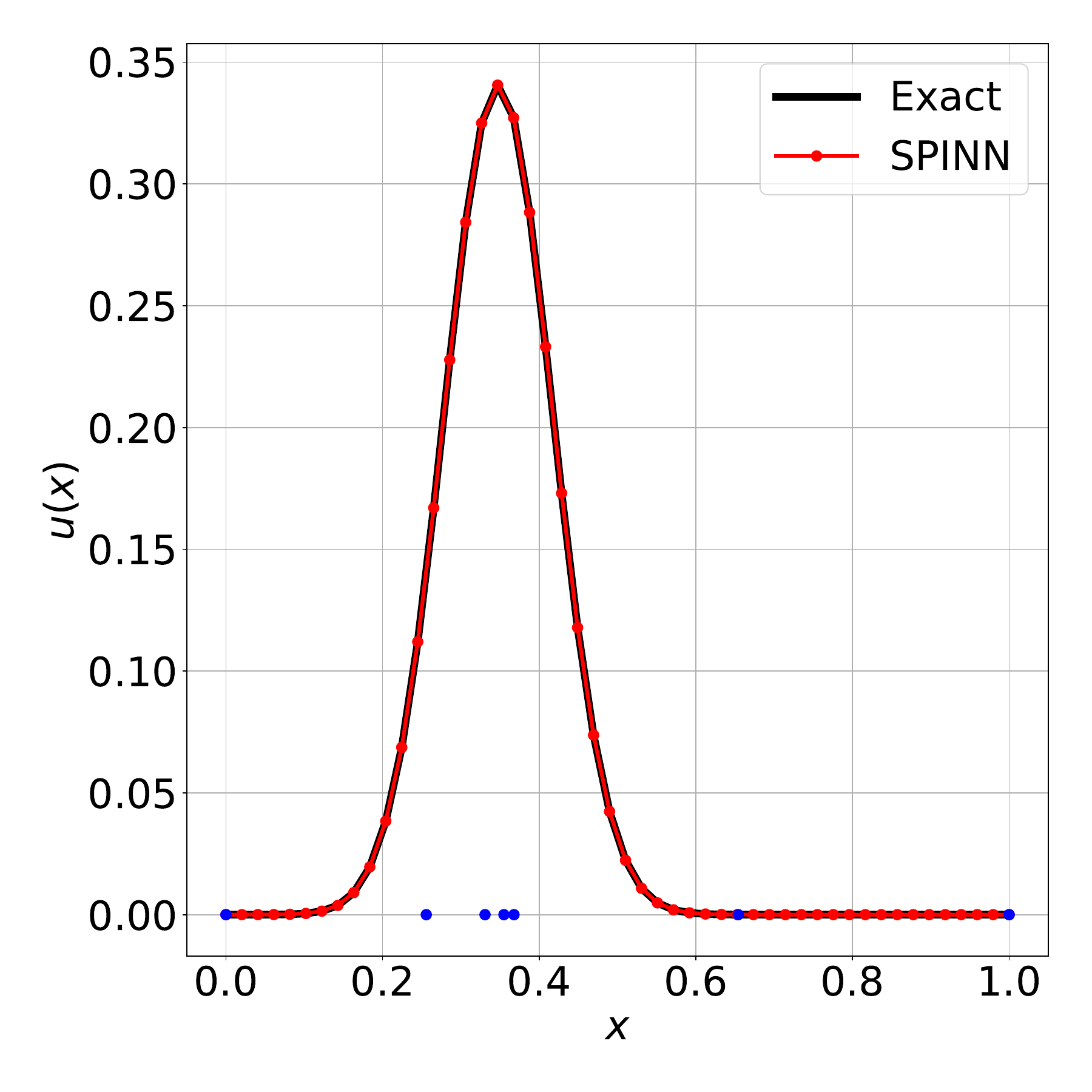}
\caption{$n=10$}
\label{var_spinn_ode3}
\end{subfigure}
~
\begin{subfigure}{0.5\textwidth}
\includegraphics[width=\textwidth]{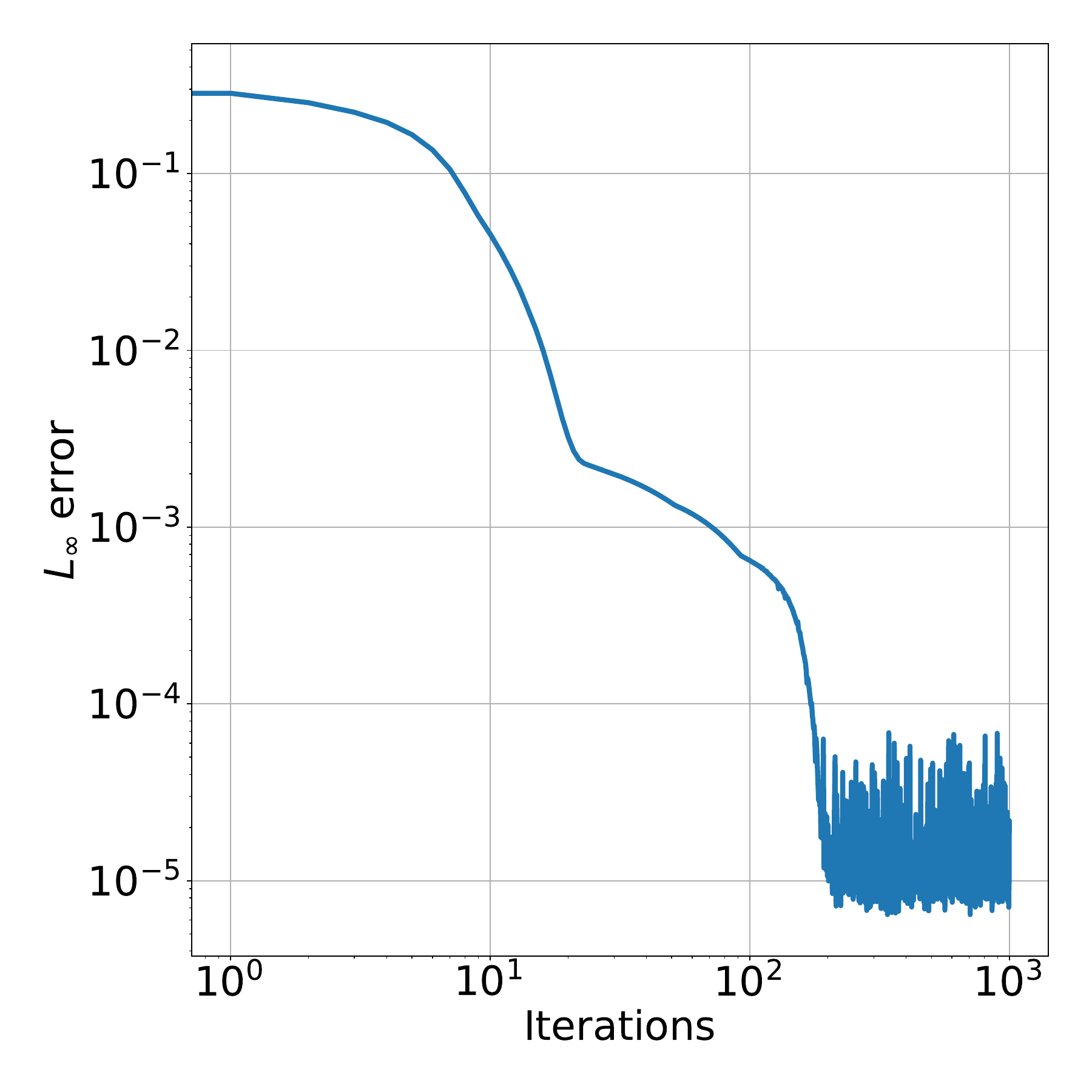}
\caption{$L_{\infty}$ error}
\label{var_spinn_ode3_error}
\end{subfigure}
\caption{Variational SPINN for ODE \eqref{eq:ode3}. The solution obtained by SPINN along with the nodal positions are shown on the left. The evolution of the $L_\infty$ error during training is shown on the right.}
\label{fig:var_spinn_ode3}
\end{figure}

\new{The examples shown so far used collocation on the strong form of the differential equation for the loss computation. SPINNs can also be designed with variational principles directly if they are available. As a proof of concept, we recall from basic variational calculus that the Euler-Lagrange equation of the functional
\begin{displaymath}
I(u) = \int_0^1 \frac{1}{2} \left(\frac{du(x)}{dx}\right)^2 \, dx - \int_0^1 f(x) u(x) \, dx,
\end{displaymath}
where $u:[0,1]\to\mathbb{R}$ with suitable regularity and such that $u(0) = u(1) = 0$, is precisely the differential equation
\begin{displaymath}
\frac{d^2u (x)}{dx^2} + f(x) = 0,
\end{displaymath}
defined over the domain $(0,1)$ with zero Dirichlet boundary conditions. The functional associated with the differential equation is chosen as the loss function. Since the functional is defined via integrals, an appropriate choice of quadrature is required. For the examples shown here a simple Riemann sum over a uniform partition of the interval $[0,1]$ is used to evaluate the integral. Basic examples are shown in Figure~\ref{fig:var_spinn_ode3}. We note that the variational formulation of SPINN is sensitive to the order of quadrature used.  A coarse discretization results in a loss of stability for the true solution.  The correct solution is however captured when the discretization is fine enough.}

\subsubsection{Fourier-SPINN}
\begin{figure}
\begin{subfigure}{0.5\textwidth}
\includegraphics[width=\textwidth]{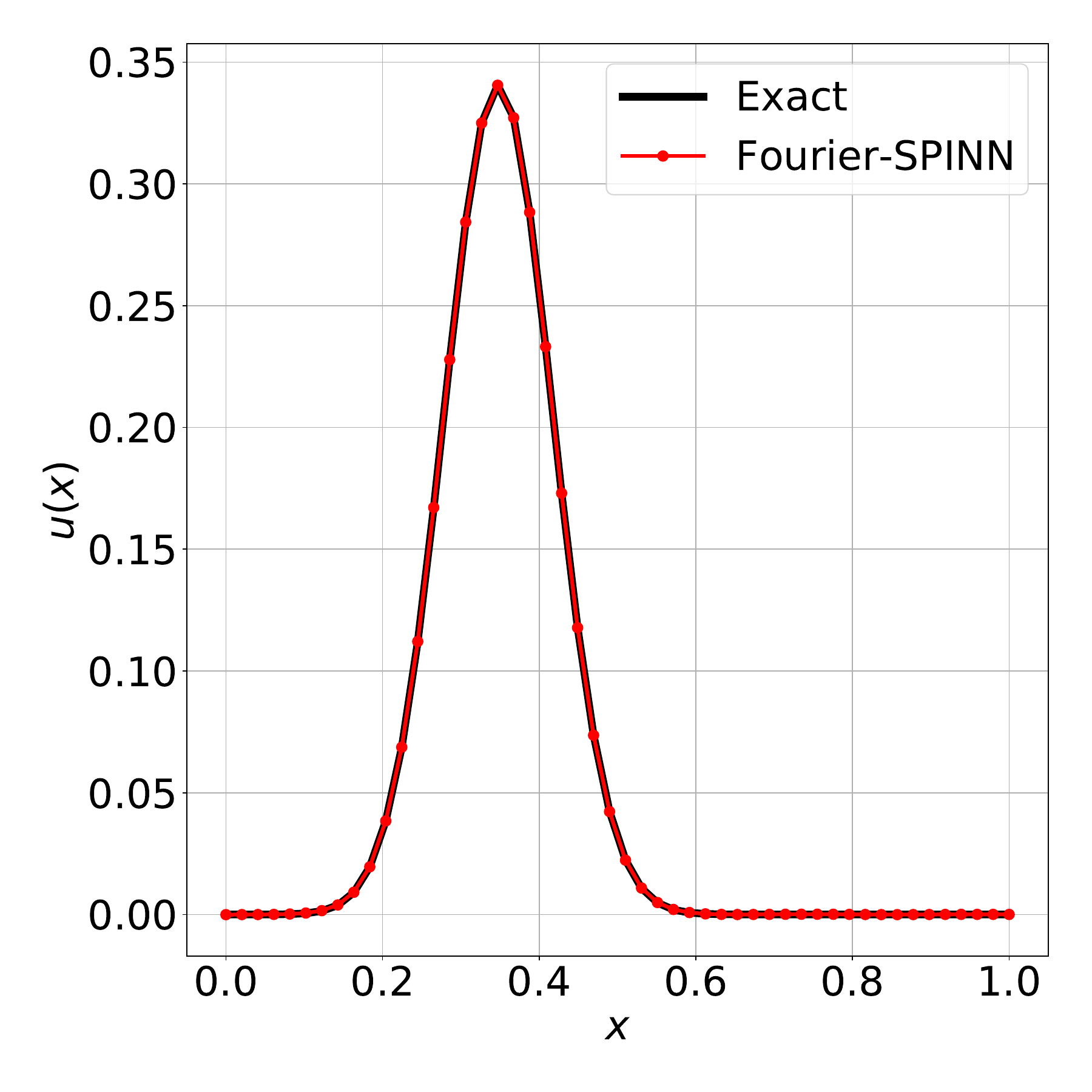}
\caption{$50$ Fourier modes}
\label{fig:fourier_spinn_ode3_m50}
\end{subfigure}
~
\begin{subfigure}{0.5\textwidth}
\includegraphics[width=\textwidth]{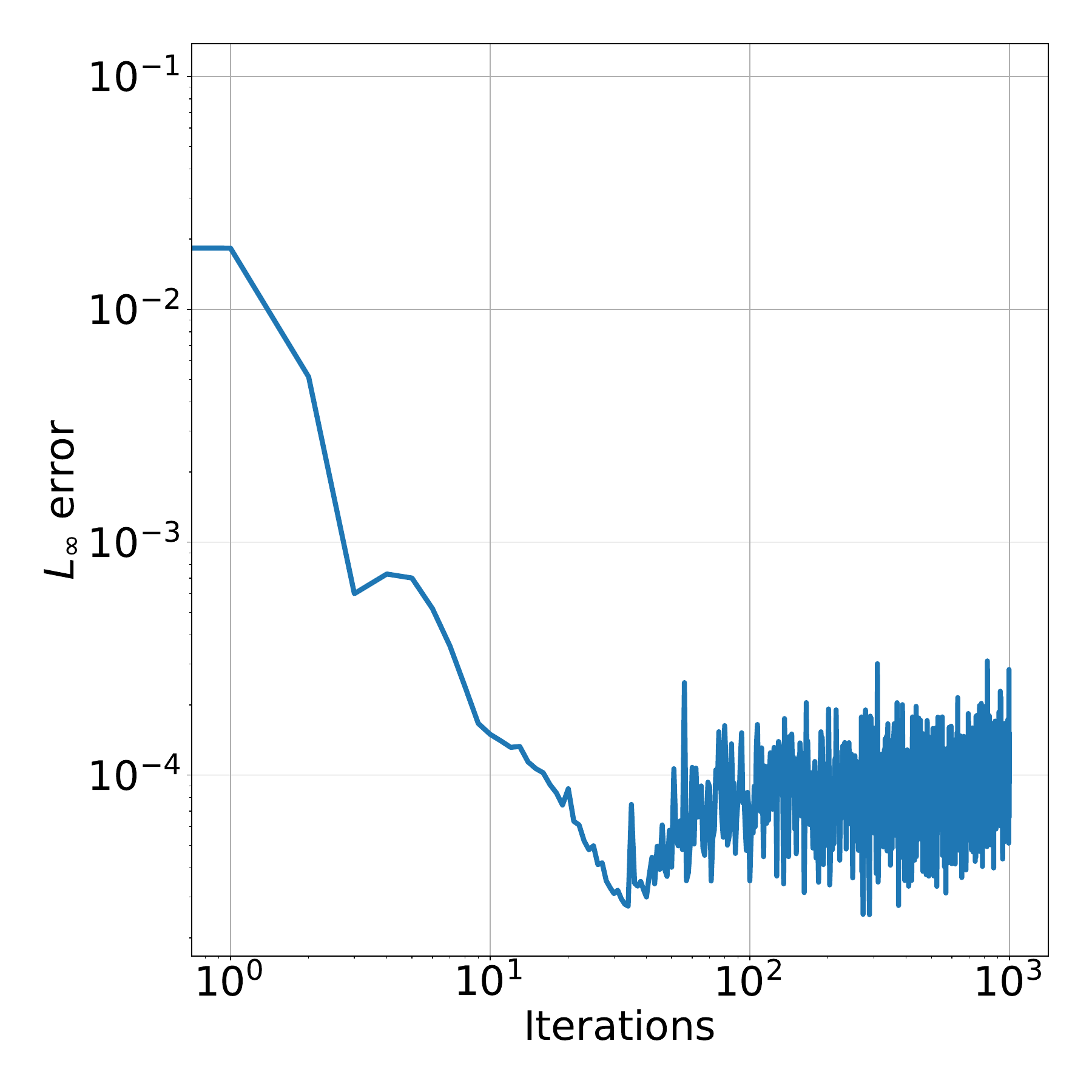}
\caption{$L_\infty$ error}
\label{fig:fourier_spinn_ode3_error}
\end{subfigure}
\caption{Fourier-SPINN solution of ODE \eqref{eq:ode3} with $50$ modes, along with the $L_\infty$ error during training.}
\label{fig:fourier_spinn_ode3}
\end{figure}

\new{We also use this opportunity to illustrate Fourier-SPINN. Preliminary results showing the solution of ODE \eqref{eq:ode3} using Fourier SPINN with strong form collocation and Gaussian kernel is shown in Figure~\ref{fig:fourier_spinn_ode3}.} \rb{The solution obtained required the use of $m=50$ modes. The learning rate was set to $10^{-4}$.} \rr{The corresponding $L_\infty$ error plot is also shown in Figure~\ref{fig:fourier_spinn_ode3}.} \new{We note that using fewer modes results in oscillatory solutions, as expected from a Fourier fit. We remark that the Fourier version of SPINN is useful for problems with periodic boundary conditions.}

\subsubsection{ODE with highly oscillatory solution}
\begin{figure}
\begin{subfigure}{0.5\textwidth}
\includegraphics[width=\textwidth]{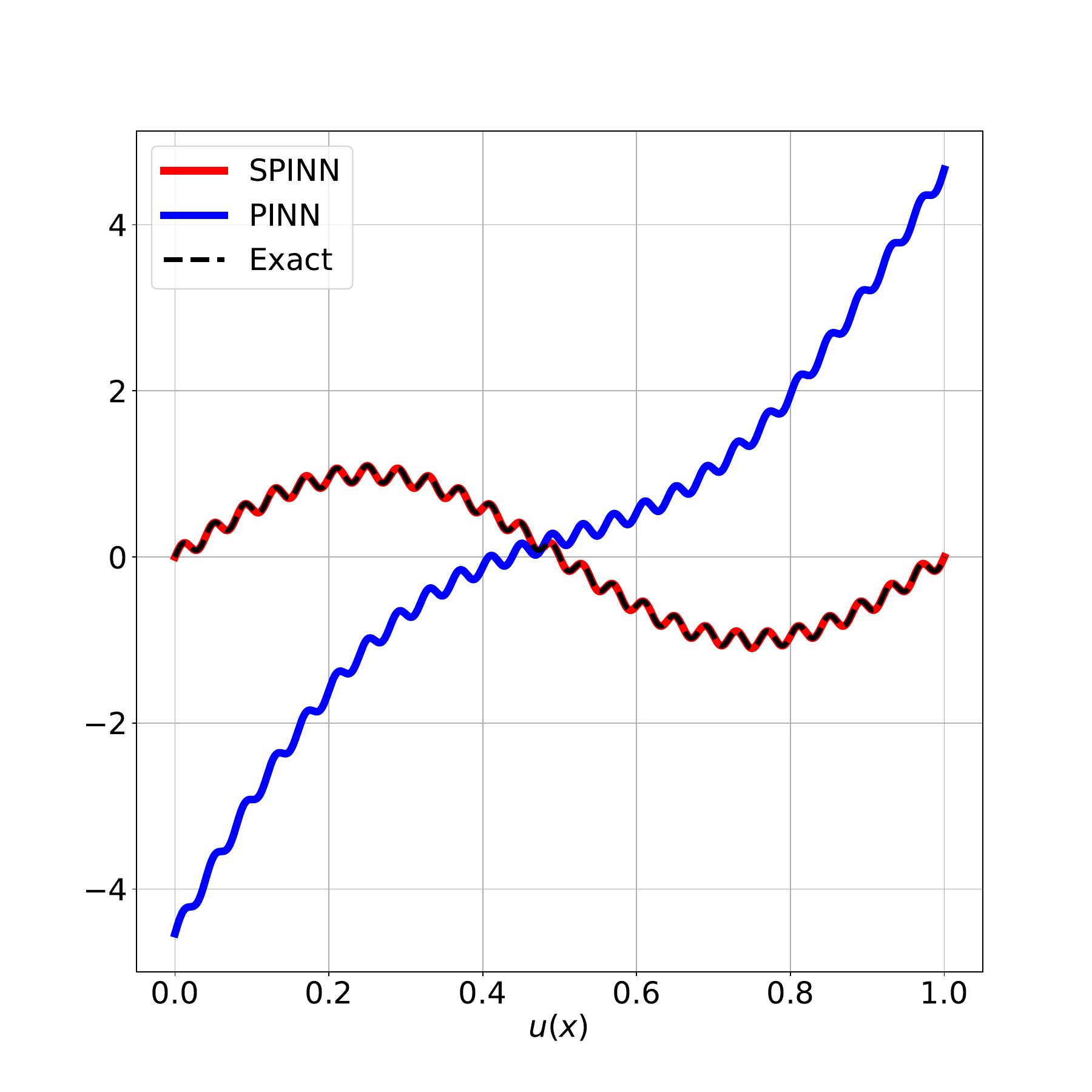}
\caption{SPINN vs.\ PINN}
\label{ode4_pinn}
\end{subfigure}
~
\begin{subfigure}{0.5\textwidth}
\includegraphics[width=\textwidth]{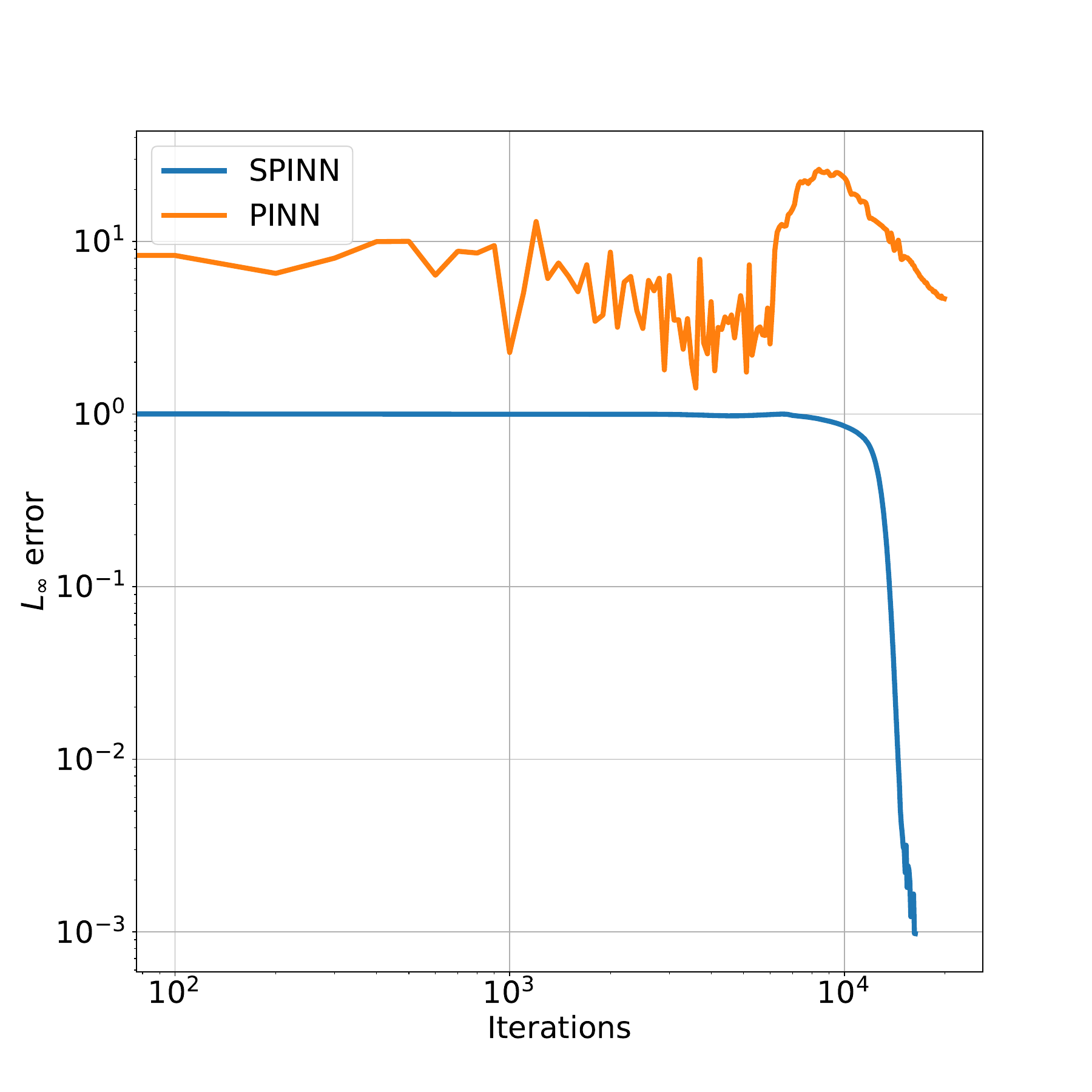}
\caption{$L_\infty$ error}
\label{ode4_pinn_Linf}
\end{subfigure}
\caption{Solution of ODE \eqref{eq:ode4} using SPINN compared with that obtained using PINN. The corresponding $L_\infty$ plot is also shown.}
\label{fig:spinn_ode4_pinn}
\end{figure}

\rr{As a final example involving ODEs, we now consider the following ODE:
\begin{equation} \label{eq:ode4}
\begin{split}
\frac{d^2u(x)}{dx^2} + 4\pi^2\sin(2\pi x) + 250\pi^2\sin(50\pi x), &\quad x \in (0,1),\\
u(0) = u(1) &= 0.
\end{split}
\end{equation}
The exact solution to ODE \eqref{eq:ode4} is easily computed as
\begin{equation} \label{eq:ode4_exact}
u(x) = \sin(2\pi x) + 0.1\sin(50\pi x).
\end{equation}
This ODE was studied using PINN in \cite{WANG2021113938} as an example of an ODE with multiscale features. This is an example of a problem where PINN performs poorly. In \cite{WANG2021113938}, the authors provide a solution of this problem using a PINN with a 5 layer fully connected network with 200 neurons in each layer and report poor convergence and an inability of PINN to learn the correct solution even after $10^7$ iterations. In sharp contrast, we illustrate below the superior performance of SPINN. We present SPINN solutions of \eqref{eq:ode4} using Gaussian kernel, a small neural network kernel and using Fourier-SPINN. In all three cases, we find that the SPINN algorithm is not only accurate but also converges to the correct solution with much smaller number of iterations.

The SPINN solution ODE \eqref{eq:ode4} using Gaussian, softplus hat and neural network kernels is shown in Figure~\ref{fig:spinn_ode4}. The corresponding $L_\infty$ error is also shown in Figure~\ref{fig:spinn_ode4}. For the Gaussian and softplus hat kernels, $n = 100$ internal nodes, $n_S = 800$ sampling points and a learning rate of $5\times 10^{-4}$ was used. For the neural network kernel, a small fully connected two layer network with $5$ neurons each was used. For all these simulations, the iterations were terminated when the $L_\infty$ error dipped below $10^{-3}$. It is worthwhile noting that the SPINN algorithm with different choices of kernels is able to satisfactorily capture the exact solution in $O(10^4)$ iterations while the standard PINN algorithm as reported in \cite{WANG2021113938} fails to converge even after $10^7$ iterations. For completeness, we also show the solution of \eqref{eq:ode4} computed using Fourier-SPINN with $200$ Fourier modes, and remaining parameters as before, in Figure~\ref{fig:spinn_ode4_fourier}. It is clearly seen that the Fourier-SPINN model too captures the exact solution within an error of approximately $10^{-3}$ in $O(10^2)$ iterations. This example thus highlights the strength of SPINN in comparison to PINN.

We plot the solution obtained using PINN with 5 hidden layers and 200 neurons in each layer as mentioned in \cite{WANG2021113938} and compare the solution obtained with SPINN. The same parameters for SPINN as used above are used here.  We iterate for a total of 20000 iterations for both methods.  For the PINN implementation, we use 2000 sampling points with sampling ratio 0.25, and a learning rate of $5\times 10^{-4}$. The results are shown in Fig.~\ref{fig:spinn_ode4_pinn}.  The $L_{\infty}$ errors as a function of iterations are also shown.  As can be clearly seen in the figure, and as reported in \cite{WANG2021113938}, the PINN solution fails to converge in sharp contrast to the SPINN solution which not only converges but is also more efficient.  The SPINN solution takes around 216 seconds to execute on a CPU whereas the PINN solution takes 1047 seconds.  This shows that for this problem SPINN is around 5 times as fast and much more accurate.

\begin{figure}
\begin{subfigure}{0.5\textwidth}
\includegraphics[width=\textwidth]{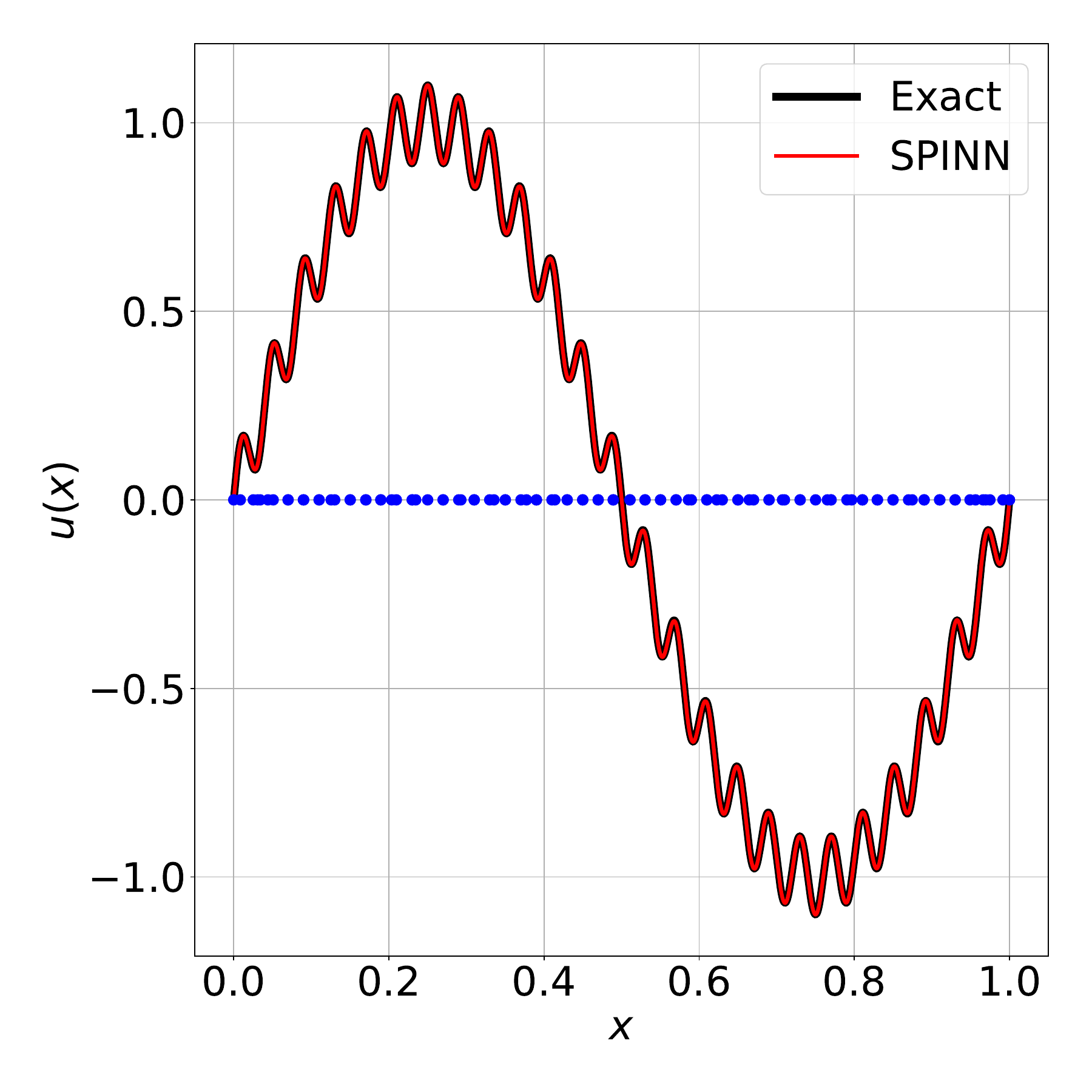}
\caption{Gaussian kernel}
\label{ode4_gaussian}
\end{subfigure}
~
\begin{subfigure}{0.5\textwidth}
\includegraphics[width=\textwidth]{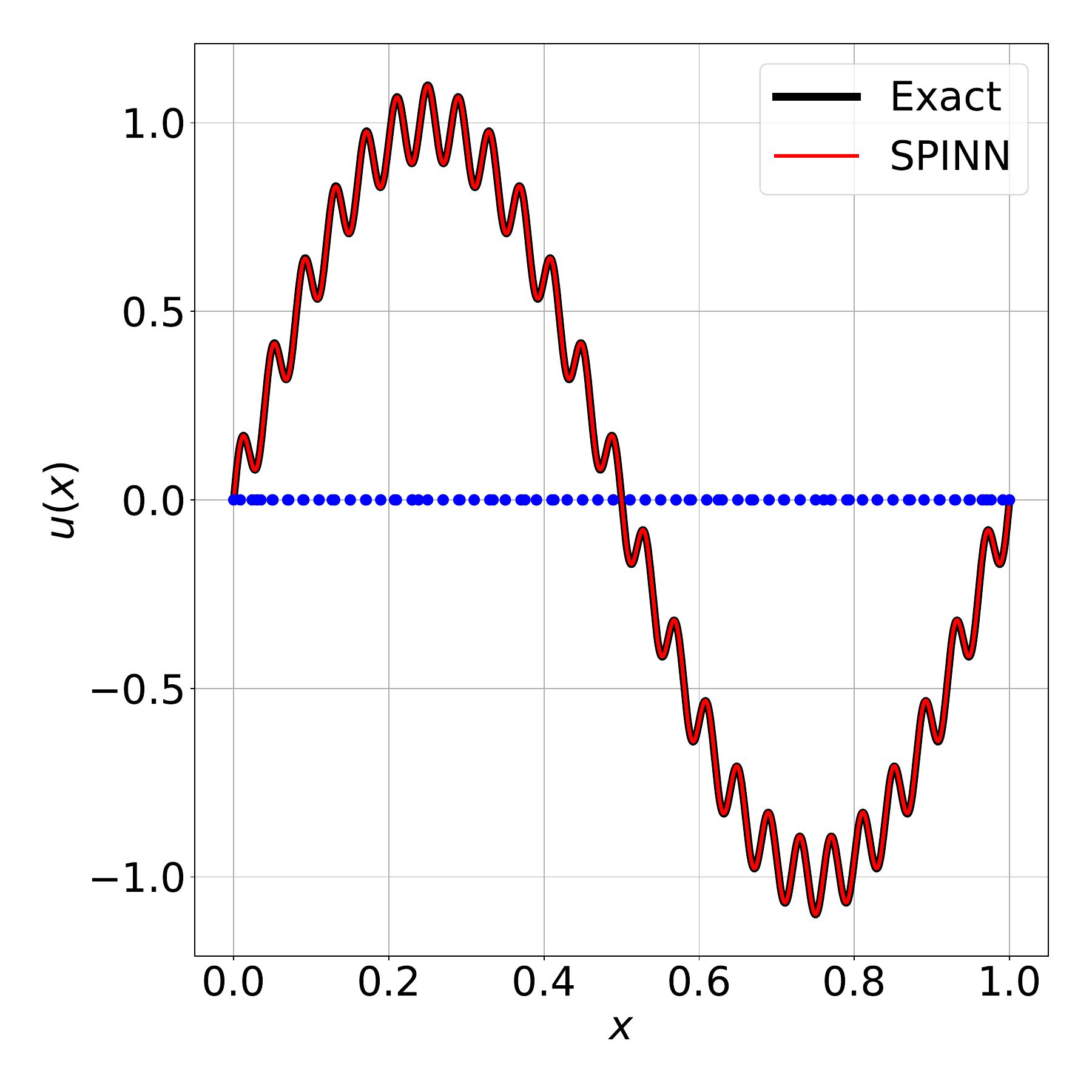}
\caption{Softplus hat kernel}
\label{ode4_softplus}
\end{subfigure}
\begin{subfigure}{0.5\textwidth}
\includegraphics[width=\textwidth]{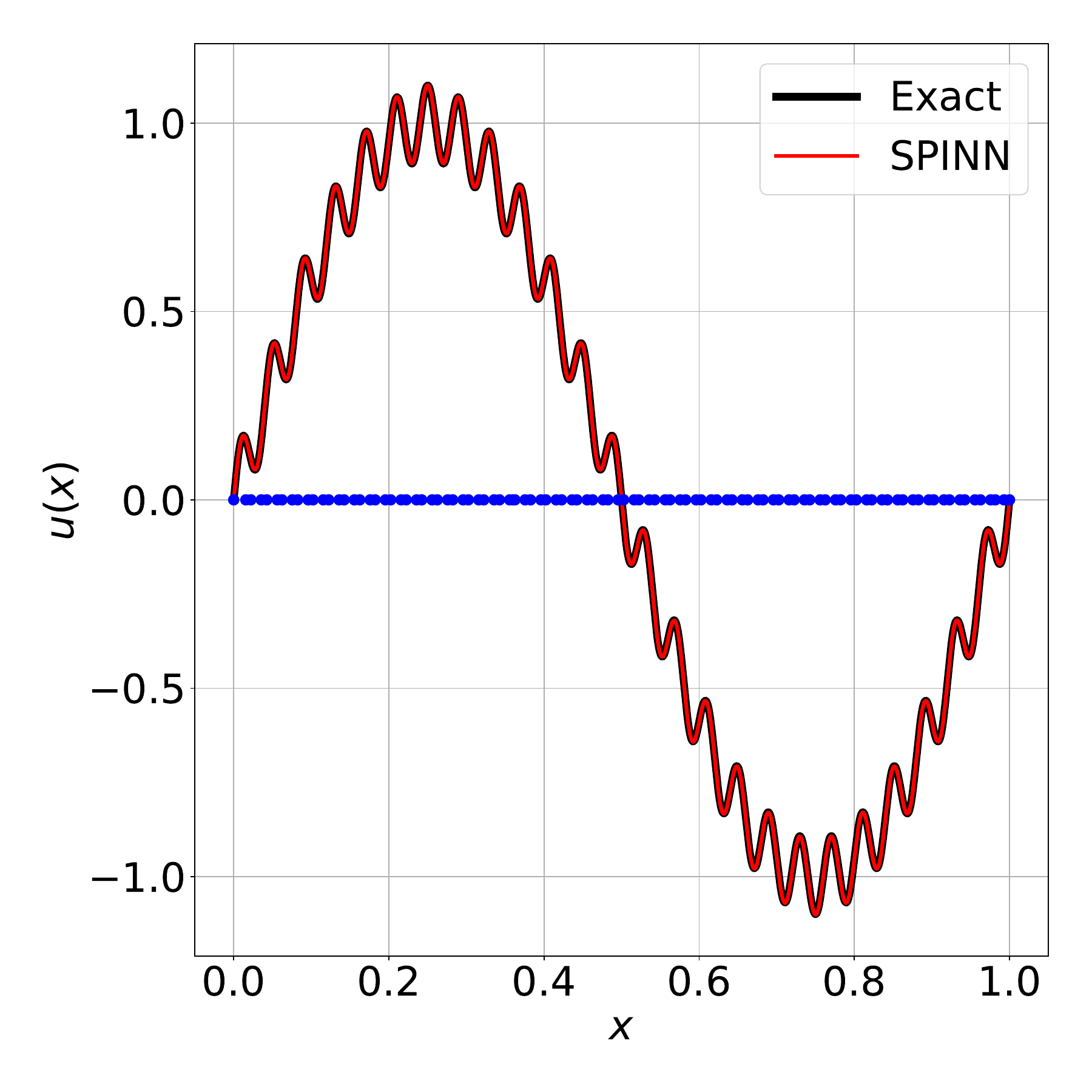}
\caption{Neural network kernel}
\label{ode4_nn_kernel}
\end{subfigure}
~
\begin{subfigure}{0.5\textwidth}
\includegraphics[width=\textwidth]{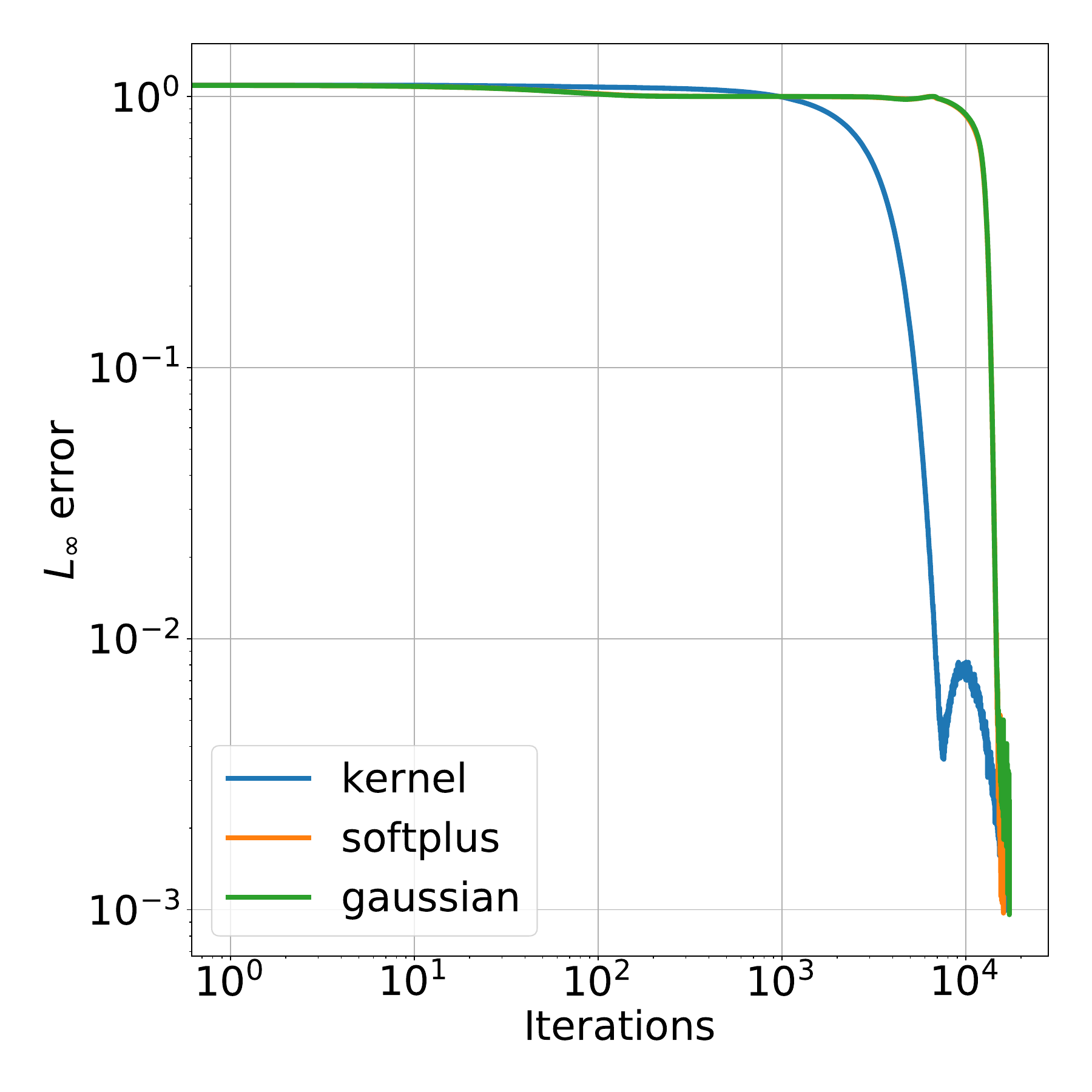}
\caption{$L_\infty$ error}
\label{ode4_Linf}
\end{subfigure}
\caption{Solution of ODE \eqref{eq:ode4} using Gaussian, softplus hat and neural network kernels. The nodal positions learnt by the corresponding SPINN models are also shown. The figure on the bottom right shows the $L_\infty$ error for these simulations. The simulations were terminated when the $L_\infty$ error dipped below $10^{-3}$.}
\label{fig:spinn_ode4}
\end{figure}

\begin{figure}
\begin{subfigure}{0.5\textwidth}
\includegraphics[width=\textwidth]{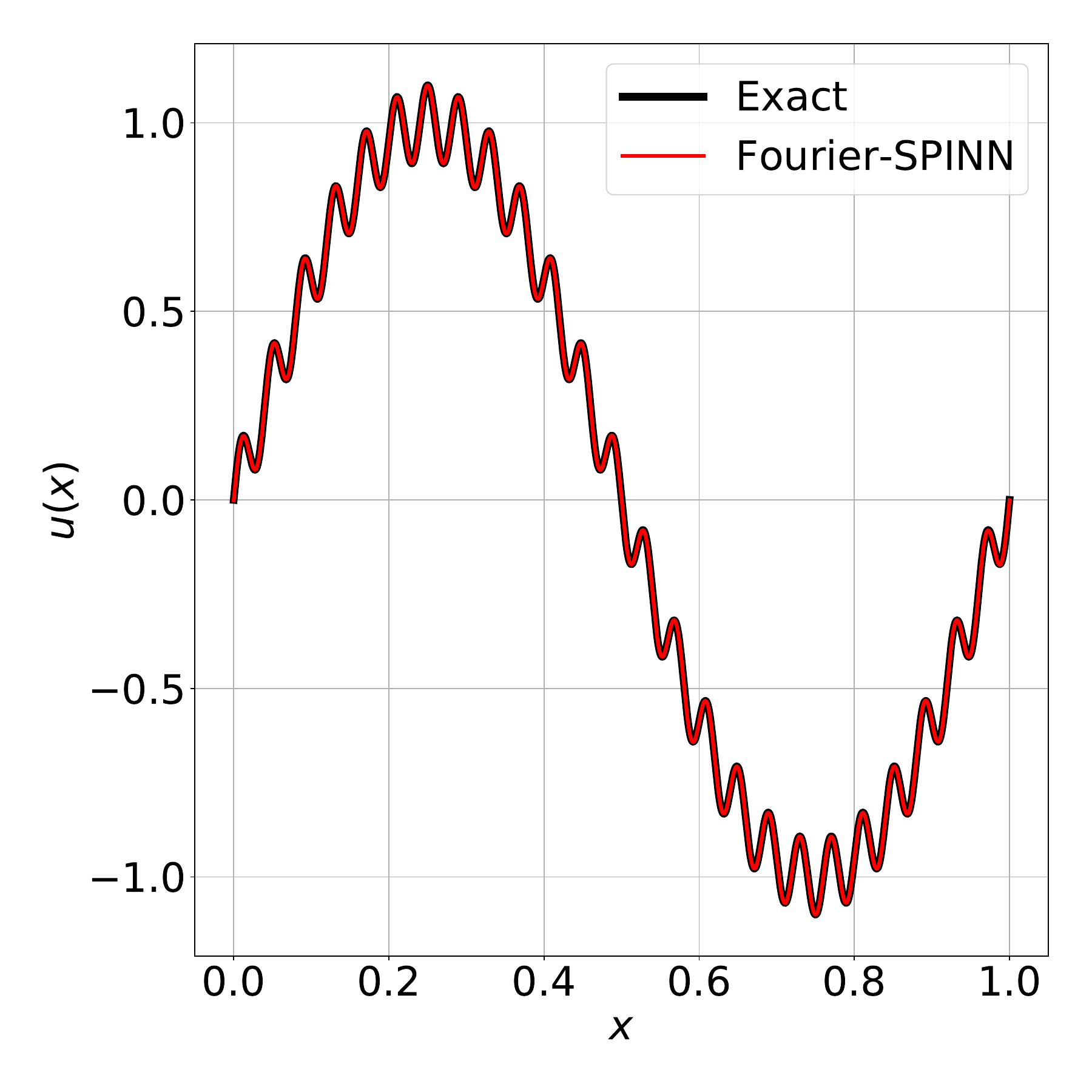}
\caption{$200$ Fourier modes}
\label{ode4_fourier_200}
\end{subfigure}
~
\begin{subfigure}{0.5\textwidth}
\includegraphics[width=\textwidth]{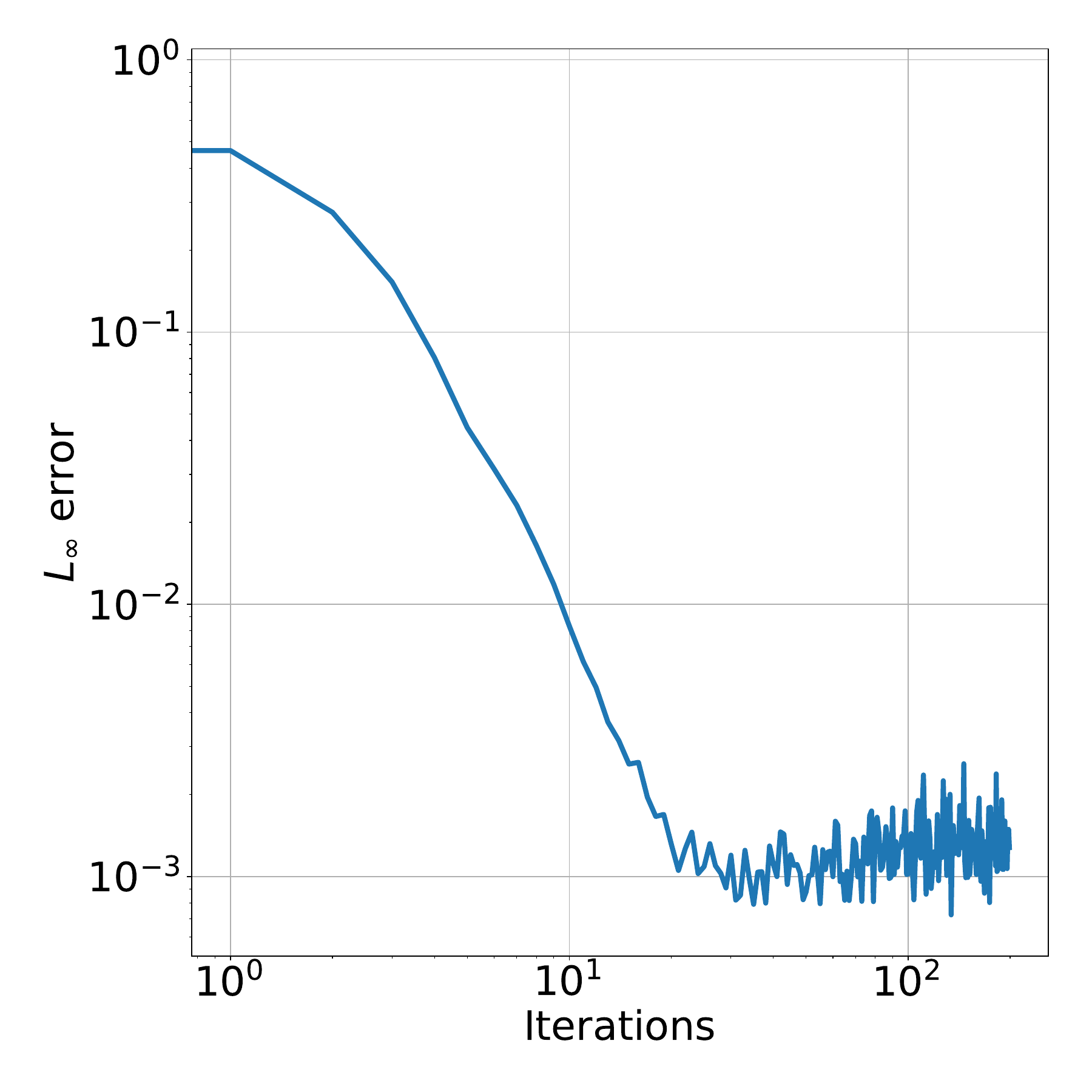}
\caption{$L_\infty$ error}
\label{ode4_fourier_Linf}
\end{subfigure}
\caption{Solution of ODE \eqref{eq:ode4} using Fourier-SPINN with $200$ Fourier modes. The corresponding $L_\infty$ plot is also shown.}
\label{fig:spinn_ode4_fourier}
\end{figure}
}

\subsection{PDEs in two dimensions}
We now present a few examples solving PDEs \new{in two dimensions. We focus in particular on linear elliptic PDEs, and discuss the solution of the Poisson equation in both regular and irregular domains.}

\subsubsection{Poisson equation}
\begin{figure}
\begin{subfigure}{0.32\textwidth}
\centering
\includegraphics[width=\textwidth]{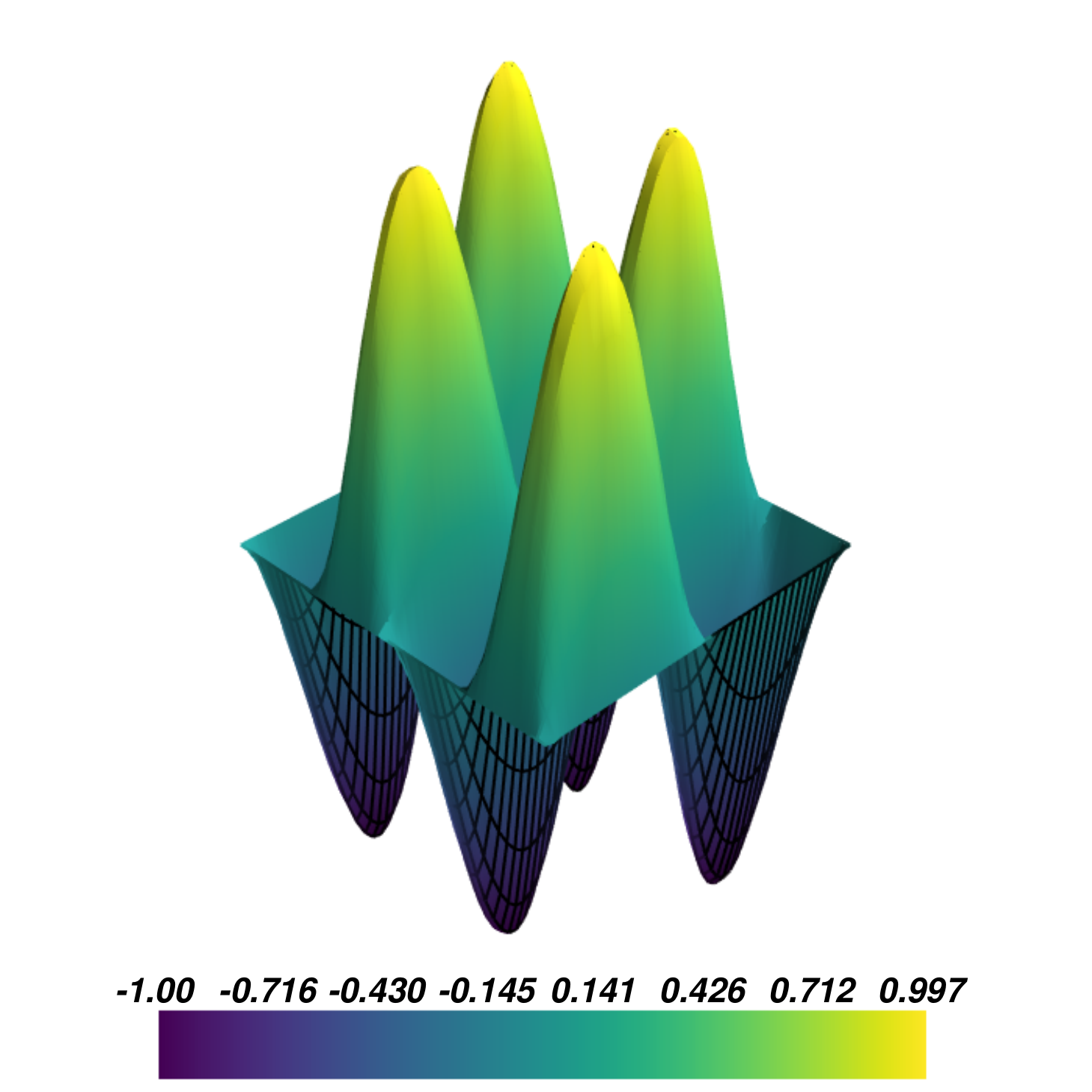}
\caption{Gaussian kernel}
\label{fig:2d_A_gaussian_n_100}
\end{subfigure}
~
\begin{subfigure}{0.32\textwidth}
\centering
\includegraphics[width=\textwidth]{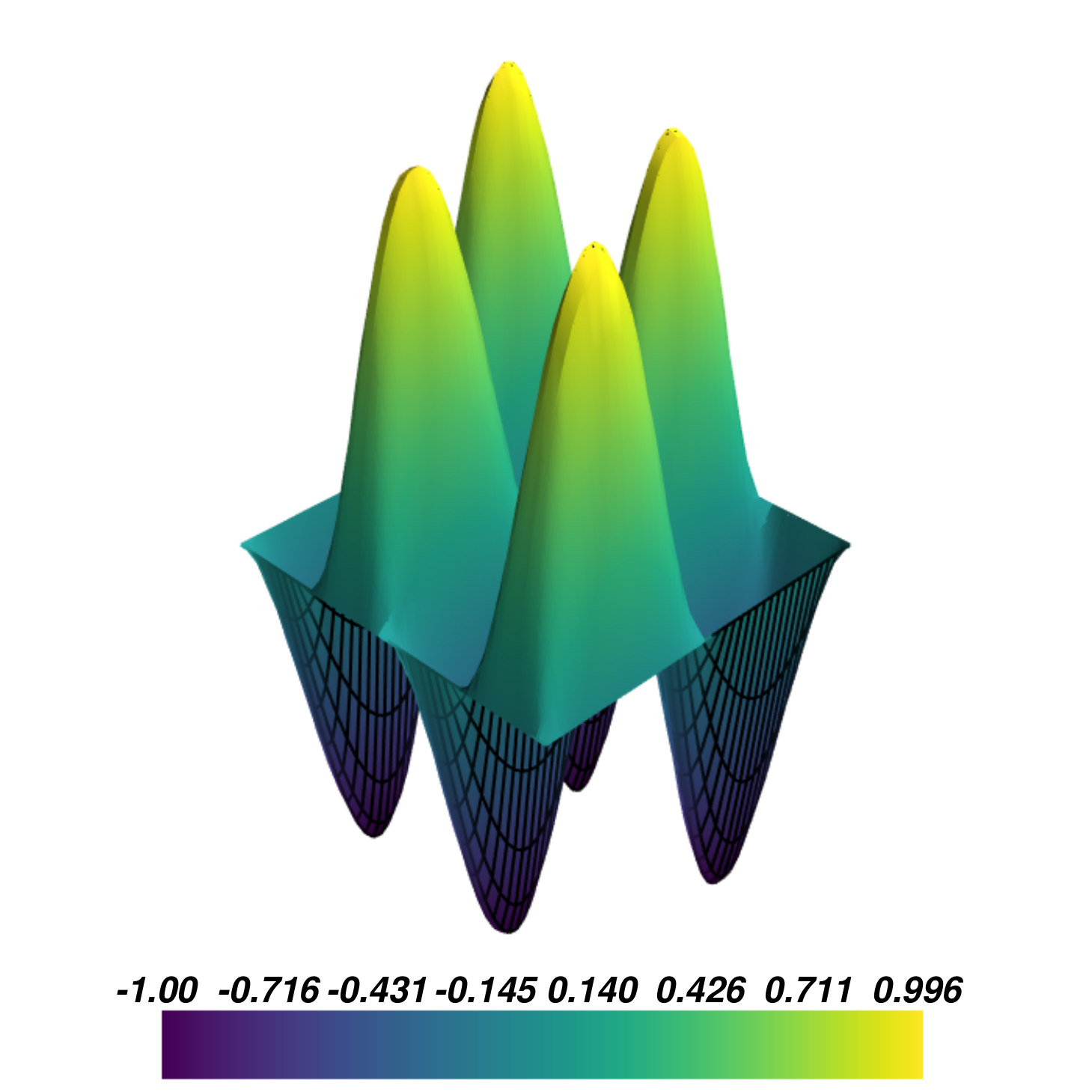}
\caption{Softplus hat kernel}
\label{fig:2d_A_softplus_n_100_a}
\end{subfigure}
~
\begin{subfigure}{0.32\textwidth}
\centering
\includegraphics[width=\textwidth]{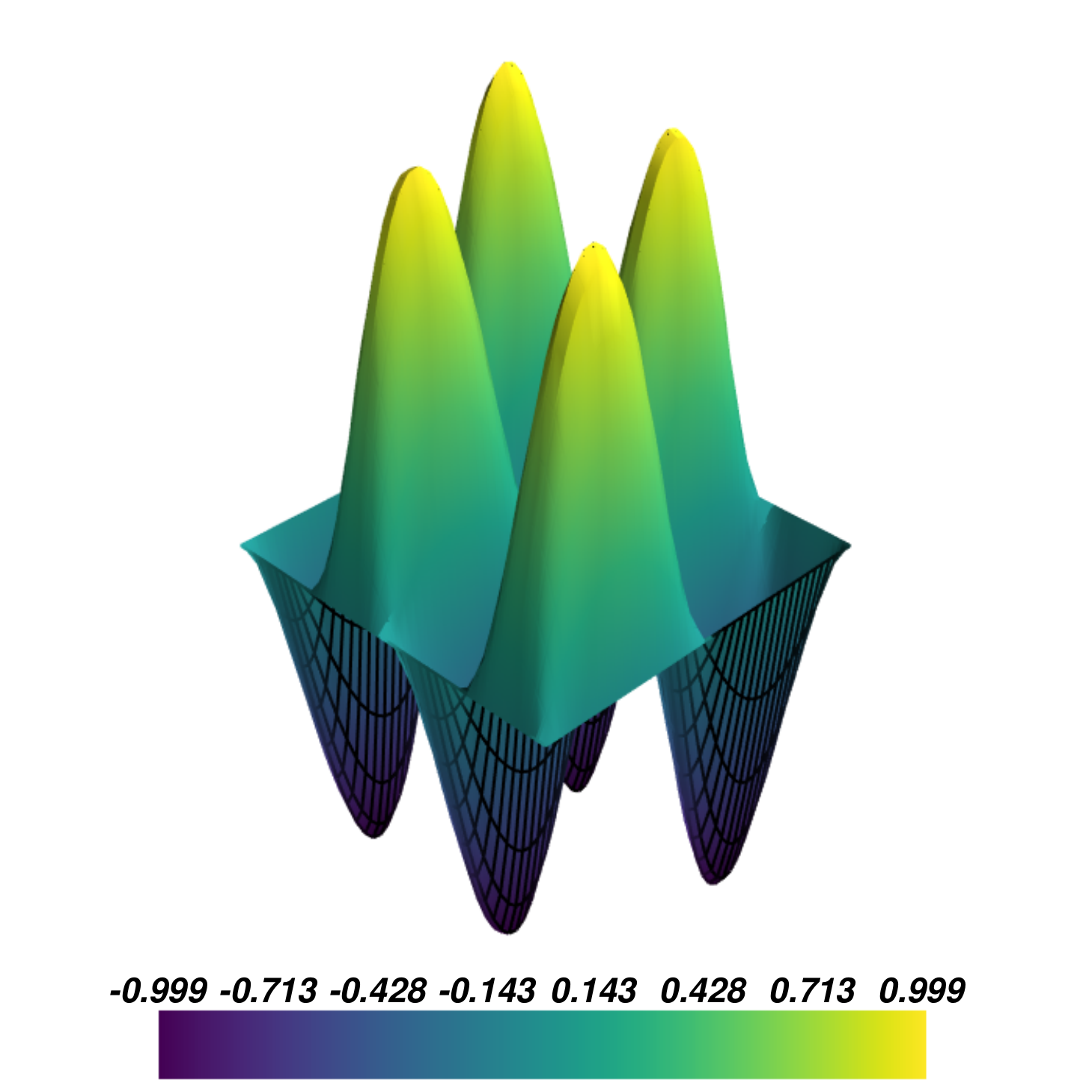}
\caption{Neural network kernel}
\label{fig:2d_A_kernel_n_100}
\end{subfigure}
\caption{Solution of PDE \eqref{eq:poisson_2d} using SPINN with different kernels. All simulations have approximately $100$ interior nodes, and about $400$ interior sampling points. The exact solution is shown in each case as a wiremesh, and the solution computed using SPINN is shown using continuous surface maps.}
\label{fig:spinn_poisson_2d}
\end{figure}

\begin{figure}
\begin{subfigure}{0.48\textwidth}
\centering
\includegraphics[width=\textwidth]{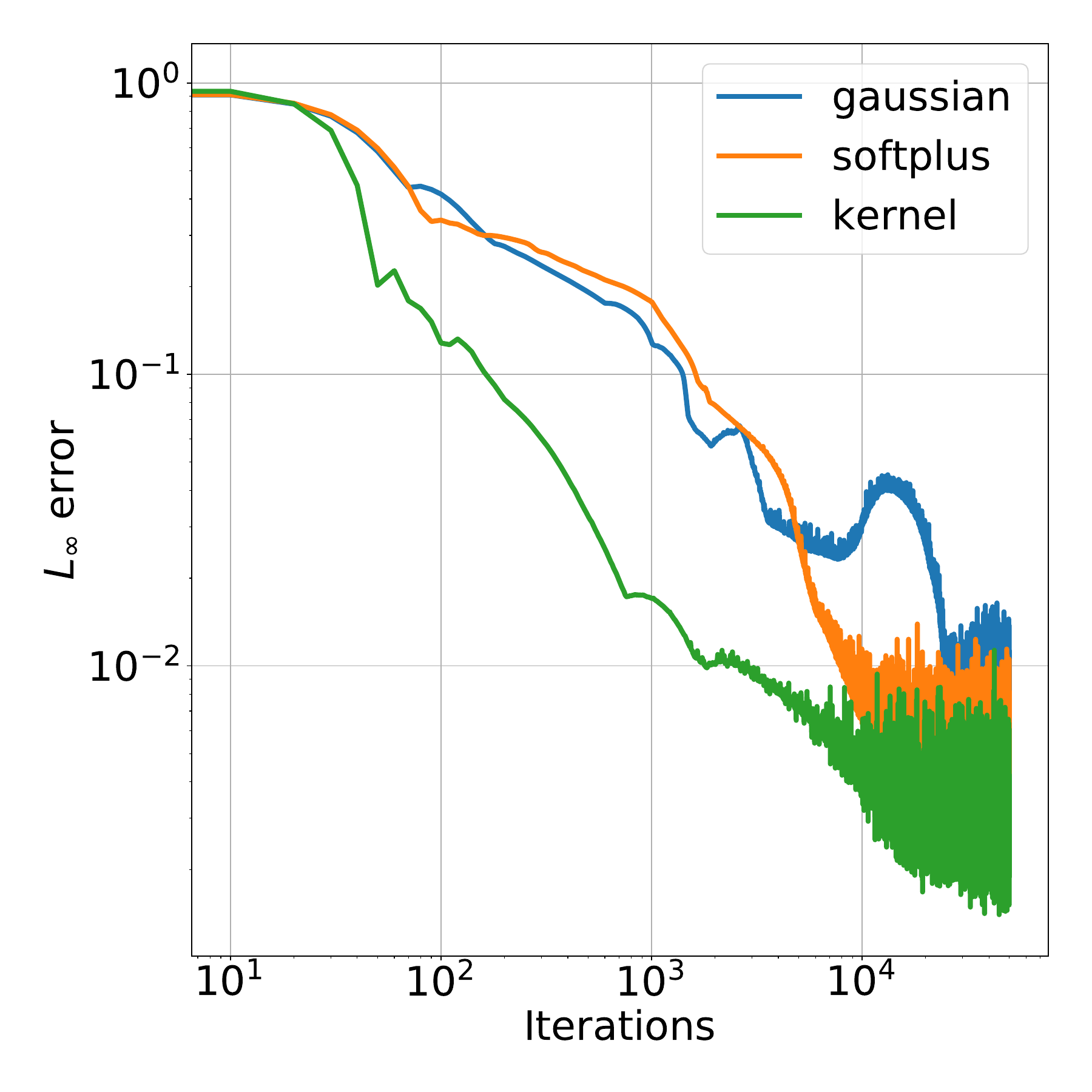}
\caption{$L_{\infty}$ error}
\label{fig:poisson_2d_Linf_n_100}
\end{subfigure}
~
\begin{subfigure}{0.48\textwidth}
\centering
\includegraphics[width=\textwidth]{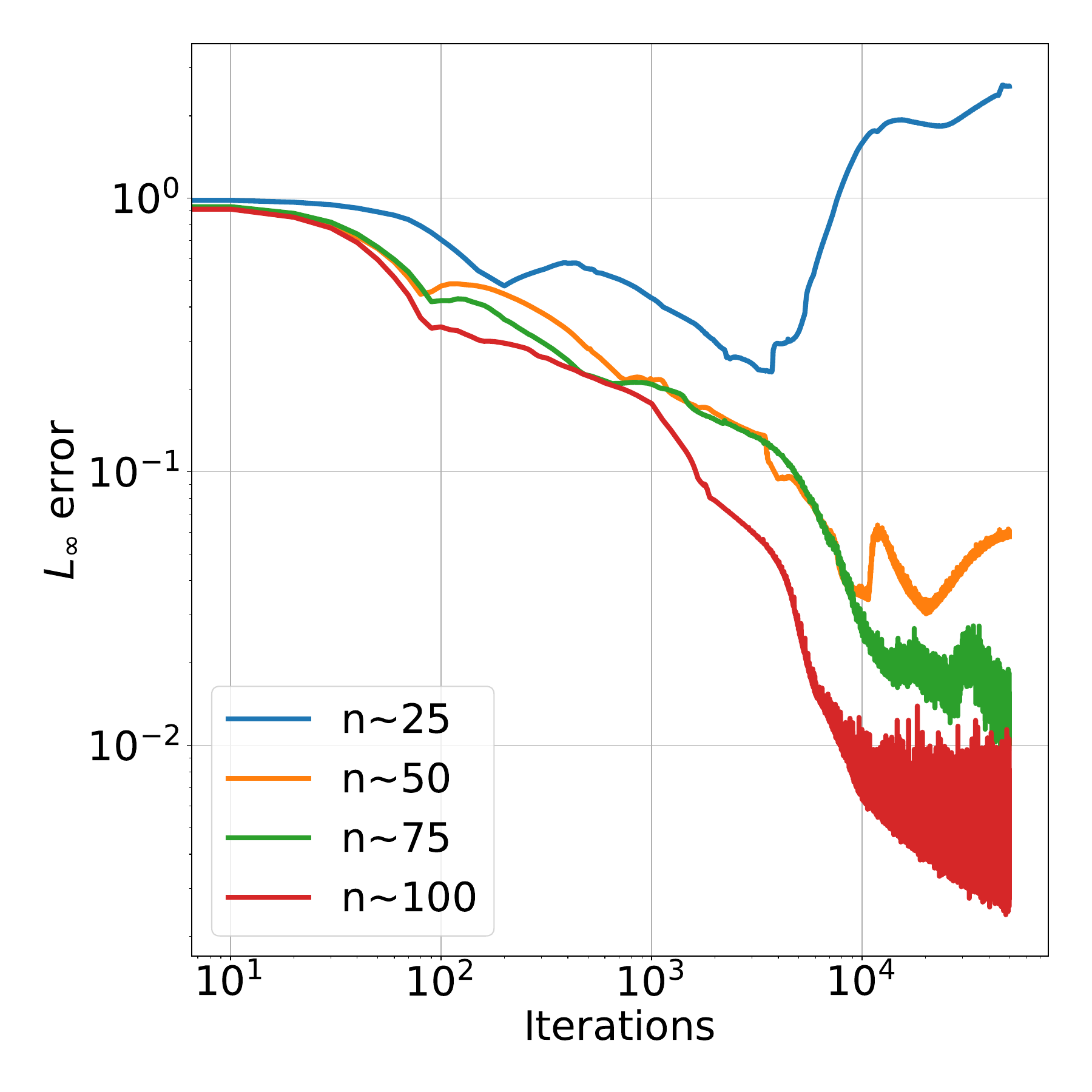}
\caption{$L_{\infty}$ error}
\label{fig:poisson_2d_Linf_nodes}
\end{subfigure}
\caption{Evolution of $L_\infty$ errors for the SPINN solution of PDE \eqref{eq:poisson_2d} for different kernels is shown in the figure on the left. The parameters used are identical to that used to generate Figure~\ref{fig:spinn_poisson_2d}. The effect of the number of interior nodes on the $L_\infty$ error is shown in the figure on the right. As expected, the errors decrease with larger number of internal nodes.}
\label{fig:poisson_2d_error}
\end{figure}

As a basic illustration of SPINN's capability to solve PDEs, we consider the Poisson equation
\begin{equation} \label{eq:poisson_2d}
\begin{split}
\nabla^2 u(x, y) = 20\pi^2 \sin 2\pi x \, \sin 4 \pi y, &\quad x,y \in D = (0,1)\times(0,1),\\
u(x,y) = 0, &\quad x, y \in \partial D.
\end{split}
\end{equation}
The exact solution to the PDE \eqref{eq:poisson_2d} is given by
\begin{equation} \label{eq:poisson_2d_exact}
u(x,y) = \sin 2\pi x \sin 4 \pi y.
\end{equation}
We first present the convergence of SPINN with different activation functions \rb{with around $100$ internal points, $40$ fixed boundary nodes, $400$ internal sampling points and $80$ boundary sampling points, in Figure~\ref{fig:spinn_poisson_2d}. The learning rate was chosen as $10^{-3}$ and a total of $5 \times 10^4$ iterations.} The $L_{\infty}$ error as a function of the iteration number is plotted for different kernels in Figure~\ref{fig:poisson_2d_Linf_n_100}. As before, we observe two distinct regimes in the error graph corresponding to learning the interior and learning the boundary conditions. While all three kernels provide a good solution, the softplus hat kernel performs better than the other two for this PDE. The convergence of the SPINN solution as a function of the number of interior nodes used is shown in Figure~\ref{fig:poisson_2d_Linf_nodes}; we observe that the error decreases with increase in the number of nodes, as expected.

\subsubsection{Square slit problem}
\begin{figure}
\begin{subfigure}{0.32\textwidth}
\includegraphics[width=\textwidth]{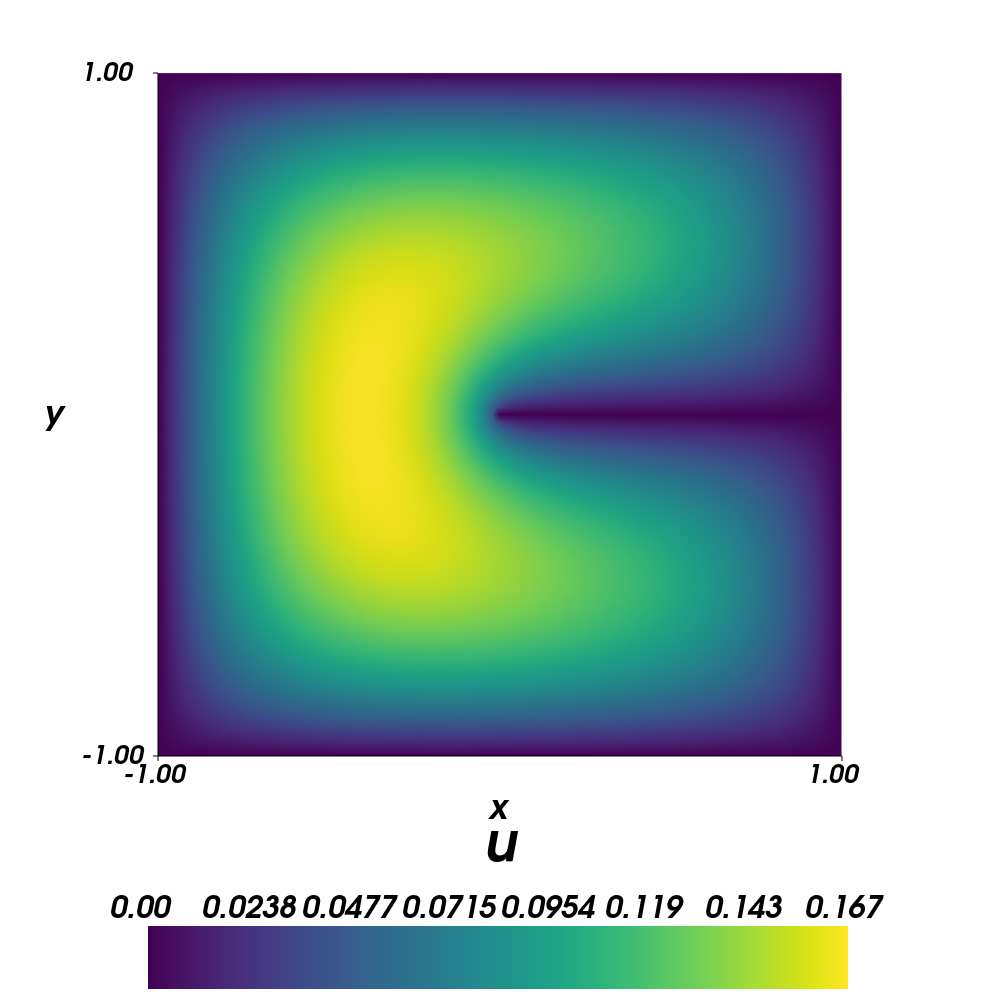}
\caption{SPINN solution of the square slit problem.}
\label{fig:poisson_2d_square_slit_sol}
\end{subfigure}
~
\begin{subfigure}{0.32\textwidth}
\includegraphics[width=\textwidth]{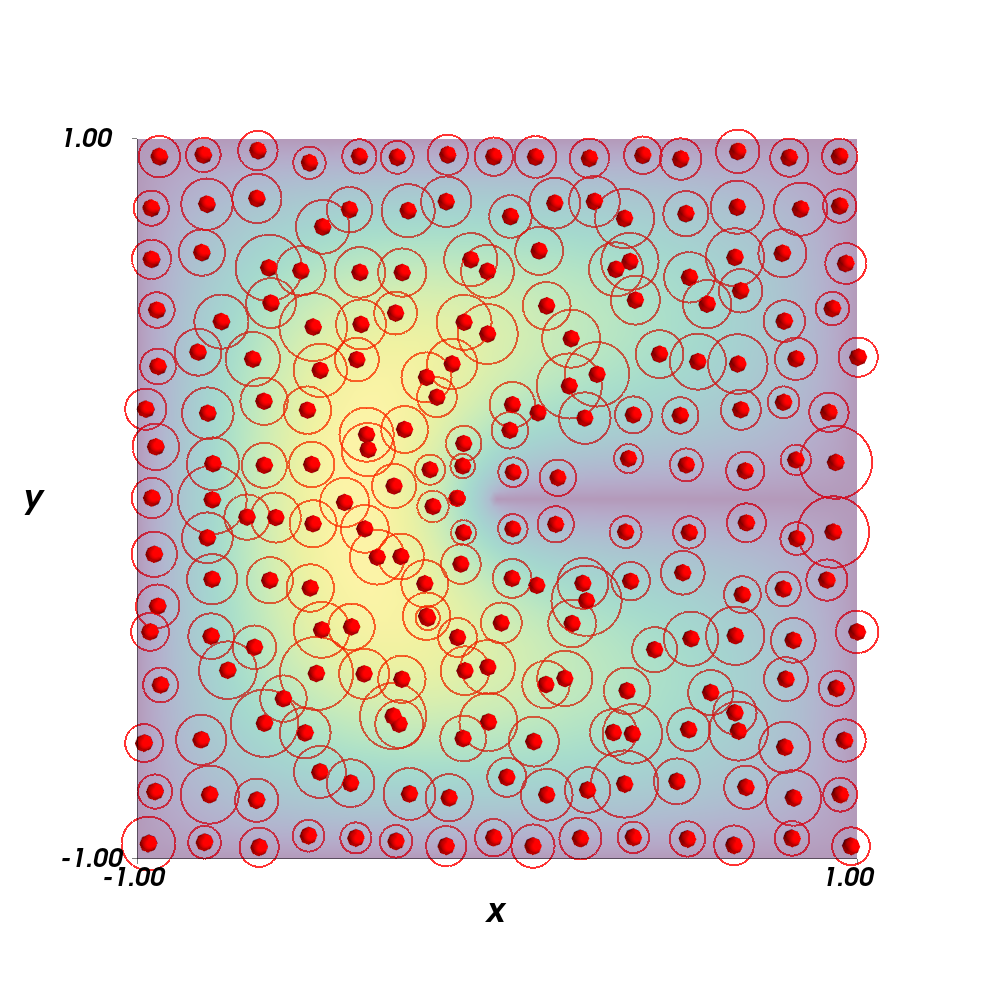}
\caption{Node and kernel width distribution.}
\label{fig:poisson_2d_square_slit_nodes}
\end{subfigure}
~
\begin{subfigure}{0.32\textwidth}
\includegraphics[width=\textwidth]{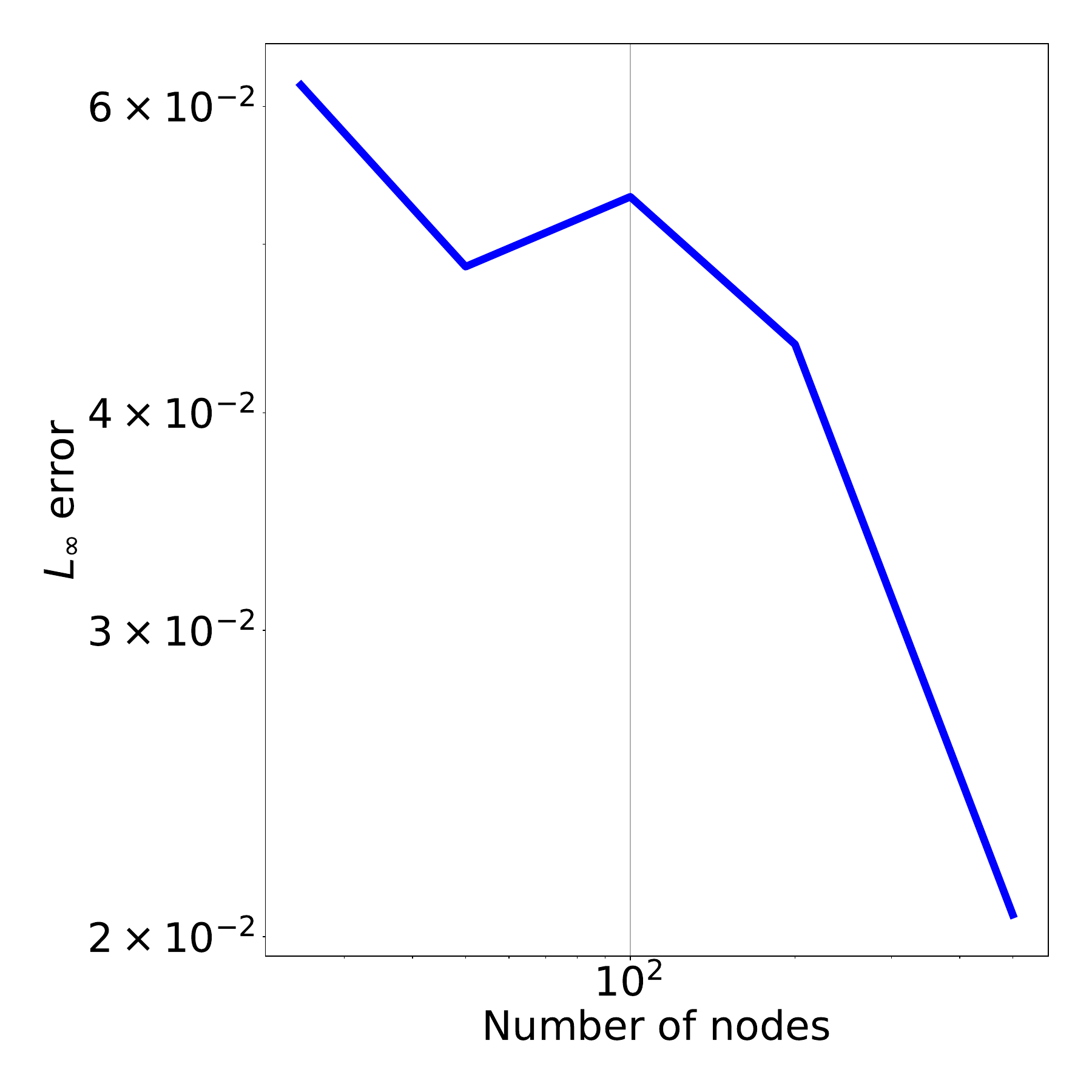}
\caption{$L_{\infty}$ error as function of number of internal nodes.}
\label{fig:poisson_2d_square_slit_convergence}
\end{subfigure}
\caption{Solution of the square slit problem. The node and kernel width distributions for the solutions learnt by SPINN are shown, along with a plot of the $L_\infty$ error as a function of the number of nodes.}
\label{fig:spinn_pde2d_square_slit}
\end{figure}

As a second example we consider the Poisson equation on a square domain with a slit:
\begin{equation} \label{eq:square_slit}
\begin{split}
\nabla^2 u(x, y) + 1 = 0, &\quad D = (-1,1)\times(-1,1) \setminus [0,1),\\
u(x,y) = 0, &\quad x, y \in \partial D.
\end{split}
\end{equation}
This PDE does not have an exact solution, but it has known asymptotic properties around the origin \rb{\cite{EYu2018}}. The solution obtained using SPINN is shown in Figure~\ref{fig:poisson_2d_square_slit_sol}. \rb{For this simulation, around $200$ interior nodes, $70$ fixed boundary nodes, $600$ interior sampling points and $110$ boundary sampling points were used. The learning rate was chosen as $10^{-3}$. The computations were performed for $10^4$ iterations.}  The corresponding nodal positions are shown in Figure~\ref{fig:poisson_2d_square_slit_nodes}. It is seen that the SPINN algorithm learns the optimal position of the nodes and the size of the kernels at the nodal positions appropriately. \rb{In more detail, we expect the initially uniform nodes to rearrange themselves so as to better reflect the local geometry of the domain, which in this case is most  notable near the slit. Since the solution varies rapidly over a small spatial distance close to the slit, we expect the nodes to have smaller widths. This indeed is what we observe in our simulations and can be seen clearly in Figure~\ref{fig:poisson_2d_square_slit_nodes}. Notice how the radius of the circles around the nodes is smaller closer to the slit, thereby indicating the adaptivity of the SPINN solution to local gradients in the solution.} A comparison of the error of the SPINN solution with respect to a reference finite element solution using a very fine mesh is shown in Figure~\ref{fig:poisson_2d_square_slit_convergence}.

The mesh for the finite element solution was created using Gmsh \cite{gmsh}. The finite element solution was computed using Fenics \cite{AlnaesBlechta2015a, LoggMardalEtAl2012a}.

\subsubsection{Irregular domains}
\begin{figure}
\begin{subfigure}{0.32\textwidth}
\centering
\includegraphics[width=\textwidth]{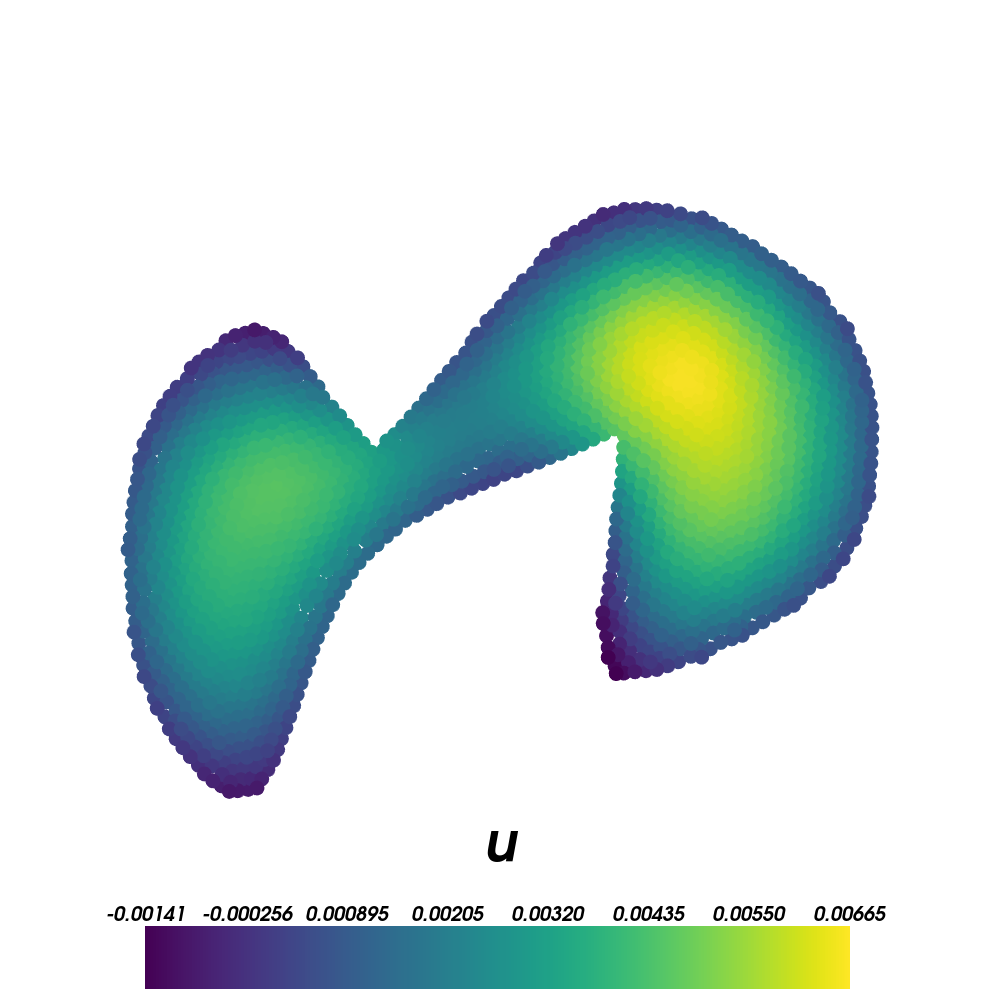}
\caption{Poisson equation on irregular domain.}
\label{fig:poisson2d_irregular}
\end{subfigure}
~
\begin{subfigure}{0.32\textwidth}
\includegraphics[width=\textwidth]{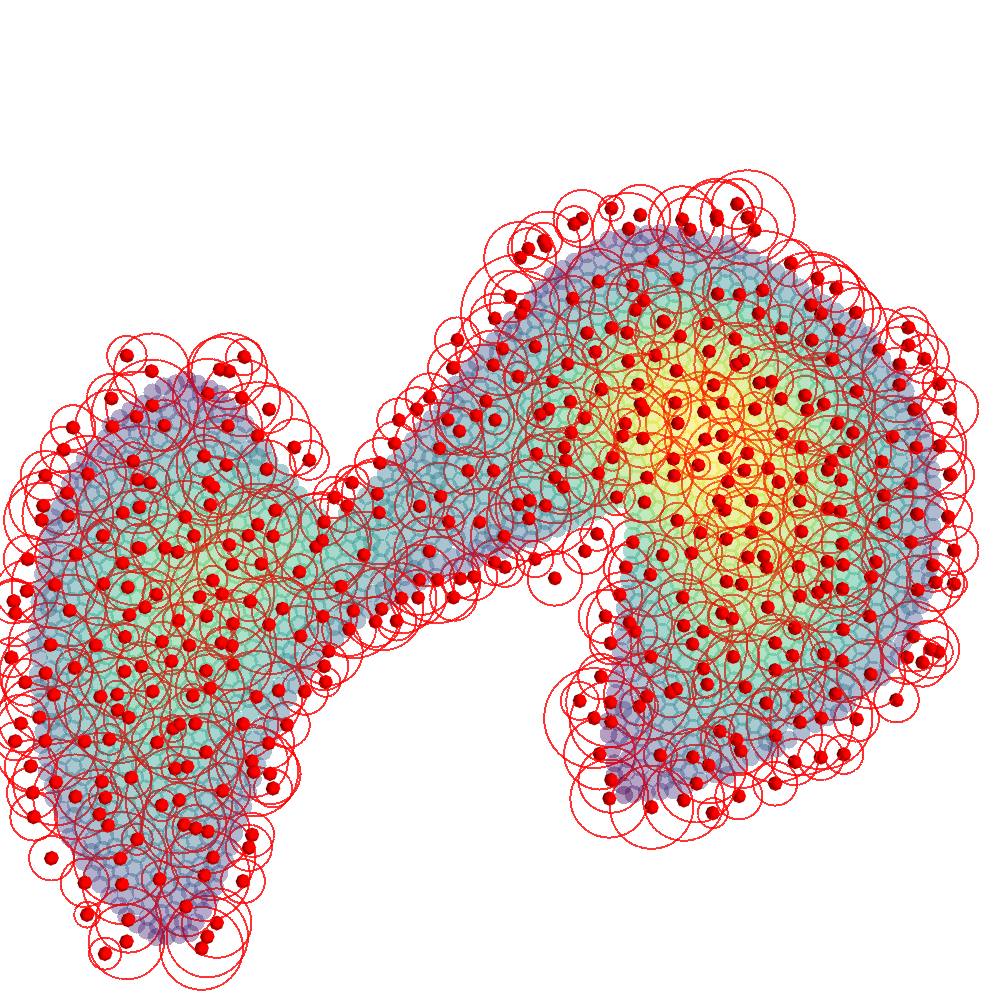}
\caption{Node and kernel width distribution.}
\label{fig:poisson2d_irregular_nodes}
\end{subfigure}
~
\begin{subfigure}{0.32\textwidth}
\includegraphics[width=\textwidth]{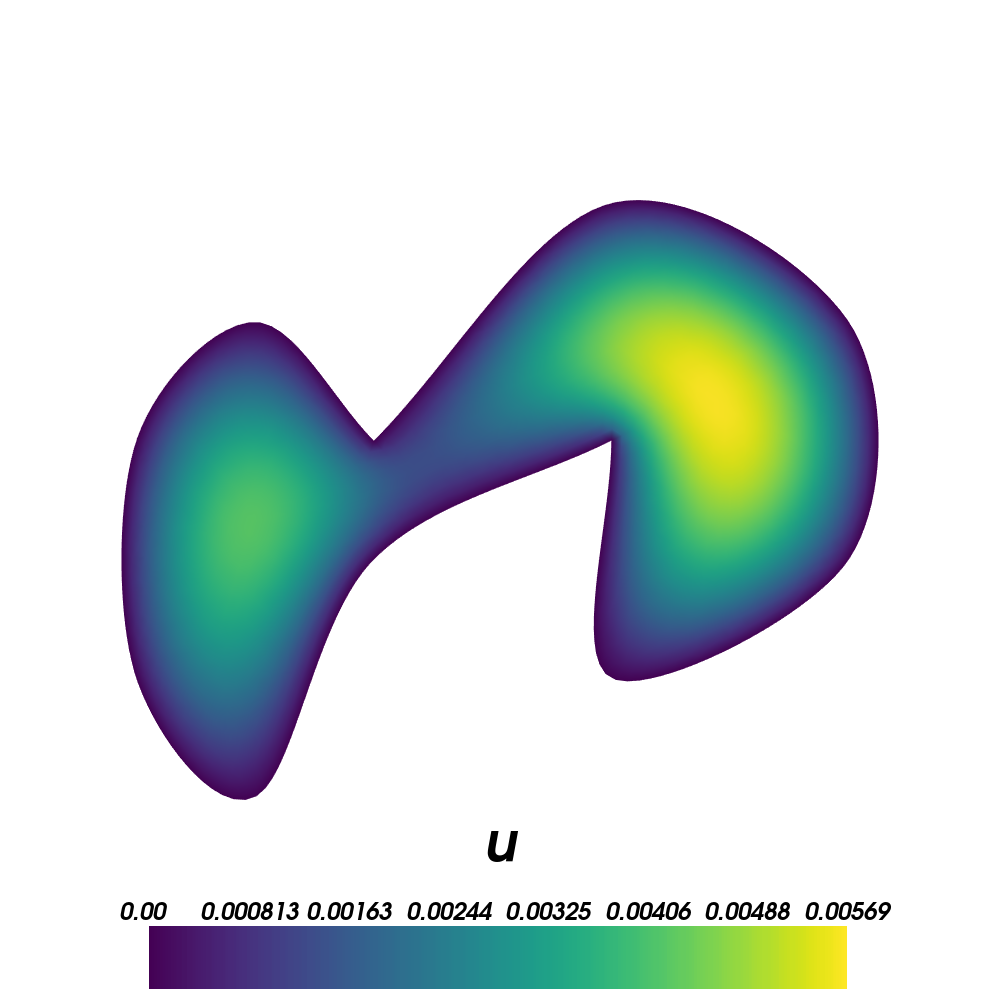}
\caption{Reference FEM solution}
\label{fig:poisson2d_irregular_fem}
\end{subfigure}
\caption{Illustration of the applicability of the SPINN algorithm to solve PDEs defined on complex geometries. Both the nodal positions and reference finite element solutions are shown for comparison.}
\label{fig:spinn_pde2d_irred_dom}
\end{figure}

The examples shown so far feature domains with regular geometric shapes. The SPINN algorithm, however, works well on arbitrarily shaped domains too. The solution of the Poisson equation $\nabla^2 u(x,y) + 1 = 0$ on an irregularly shaped domain is shown in Figure~\ref{fig:poisson2d_irregular}, and a reference finite element solution computed using a fine mesh is shown in Figure~\ref{fig:poisson2d_irregular_fem}. The distribution of the nodes in this case is shown in Figure~\ref{fig:poisson2d_irregular_nodes}. \rb{The simulation used around 450 internal nodes, 110 fixed boundary nodes, 1800 internal sampling points and 220 boundary sampling points. The learning rate was chosen as $10^{-3}$ and the iterations were carried for $5\times10^{3}$ training steps.} The $L_{\infty}$ error of the SPINN solution was found to be around $4.9\times 10^{-3}$.

\subsection{Time dependent PDEs}
\new{We present a few applications of SPINN to solve time dependent PDEs, namely the heat equation (parabolic PDE), linear advection equation (hyperbolic PDE), Burgers equation (shock formation) and Allen-Cahn equation (periodic boundary conditions and sharp gradients). We chose the various examples to highlight specific features of SPINN as highlighted in the brackets.}

\subsubsection{Heat equation}
We now present examples involving time dependent PDEs. To start with, we consider the one-dimensional heat equation
\begin{displaymath}
\begin{split}
\frac{\partial u(x,t)}{\partial t} = c^2\frac{\partial^2 u(x,t)}{\partial x^2},& \quad x \in (0,L), \; t \in [0, T],\\
u(x, 0) = f(x),& \quad x \in (0,L),\\
u(0,t) = u(L,t) = 0,& \quad t \in [0,T].
\end{split}
\end{displaymath}
We record for reference the exact solution to the heat equation displayed above. The coefficients $(b_k)_{k=1}^{\infty}$ are first defined as the Fourier coefficients of $f$: $f(x) = \sum_{k=1}^{\infty} b_k \sin k\pi x$. The coefficients $(b_k)$ are easily computed as $b_k = \frac{2}{L}\int_{0}^{L} f(x) \sin \frac{n\pi x}{L}\, dx$. The exact solution of the heat equation is then computed as
\begin{displaymath}
u(x,t) = \sum_{k=1}^{\infty} b_k \exp (-\alpha_k^2 t) \sin k\pi x, \quad \alpha_k = \frac{k\pi c}{L}.
\end{displaymath}
The example considered in the paper uses the following inputs: $f(x) = 2\sin \pi x$, $c = 1$, $L = 1$ and \rb{$T = 0.2$}. The coefficients $(b_k)$ are easily computed as $b_1 = 2$, $b_k = 0, k = 2, 3, \ldots$.

We consider two different methods to solve the heat equation. First, we solve it using the implicit FD-SPINN algorithm
\begin{displaymath}
u^{n+1} = u^n + c^2 \Delta t \, u^{n+1}_{xx}.
\end{displaymath}
It is emphasized that the spatial derivative on the right hand side of the FD-SPINN algorithm displayed above is computed exactly using automatic differentiation, in contrast to typical finite difference schemes which employ a numerical discretization of the derivative operator.  \rb{We use a learning rate of $10^{-3}$, a time step of 0.0025, 20 interior nodes, 200 sample points that are fully sampled. We iterate either until a loss of $2\times 10^{-5}$ is attained, or up to 5000 iterations.} \rr{The $L_{\infty}$ error at $T=0.2$ is $O(10^{-2})$.}

\begin{figure}
\begin{subfigure}{0.5\textwidth}
\includegraphics[width=\textwidth]{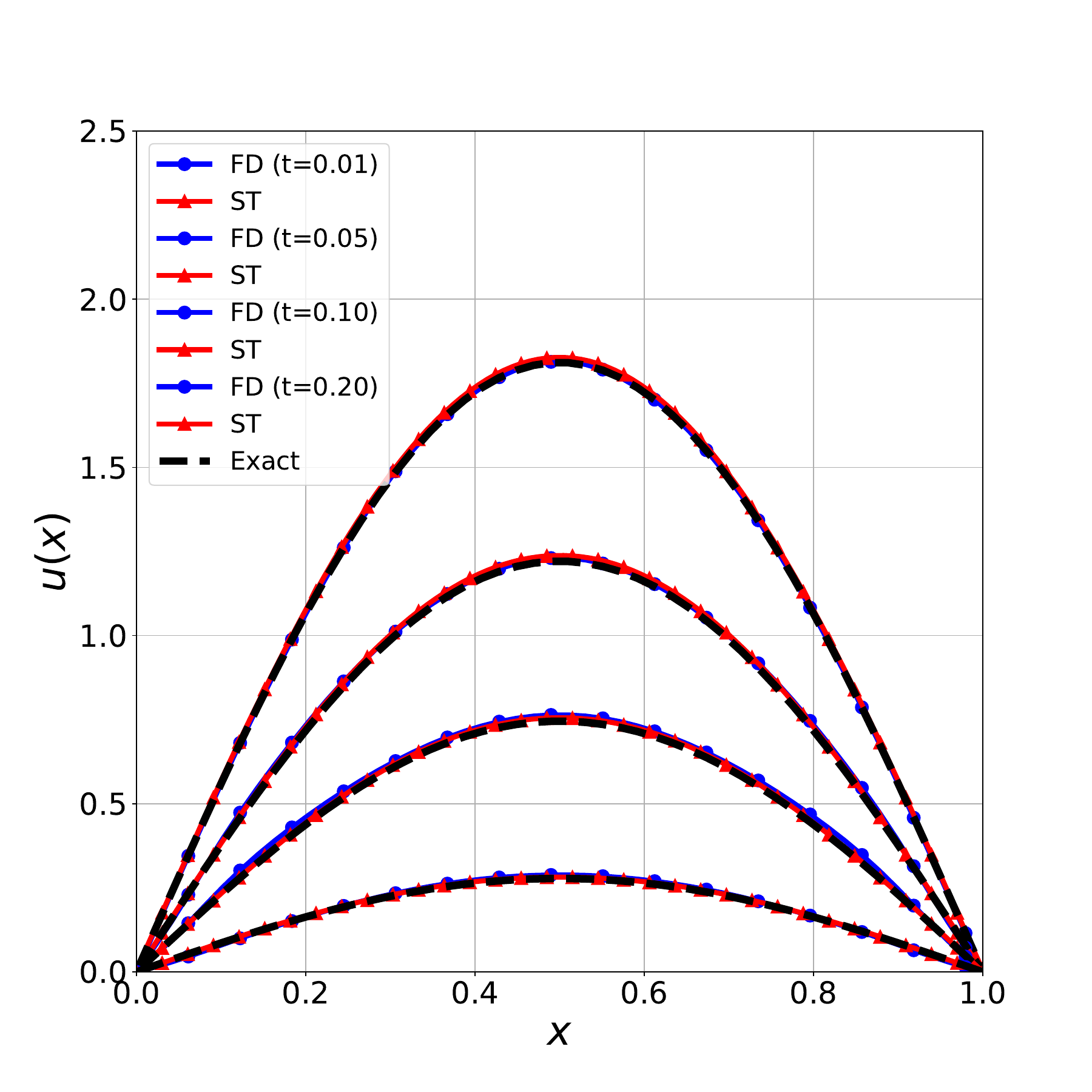}
\caption{Comparison of exact and SPINN solutions of the heat equation \rb{up to a time of $T=0.2$}.}
\label{fig:heat_eqn_compare}
\end{subfigure}
~
\begin{subfigure}{0.5\textwidth}
\includegraphics[width=\textwidth]{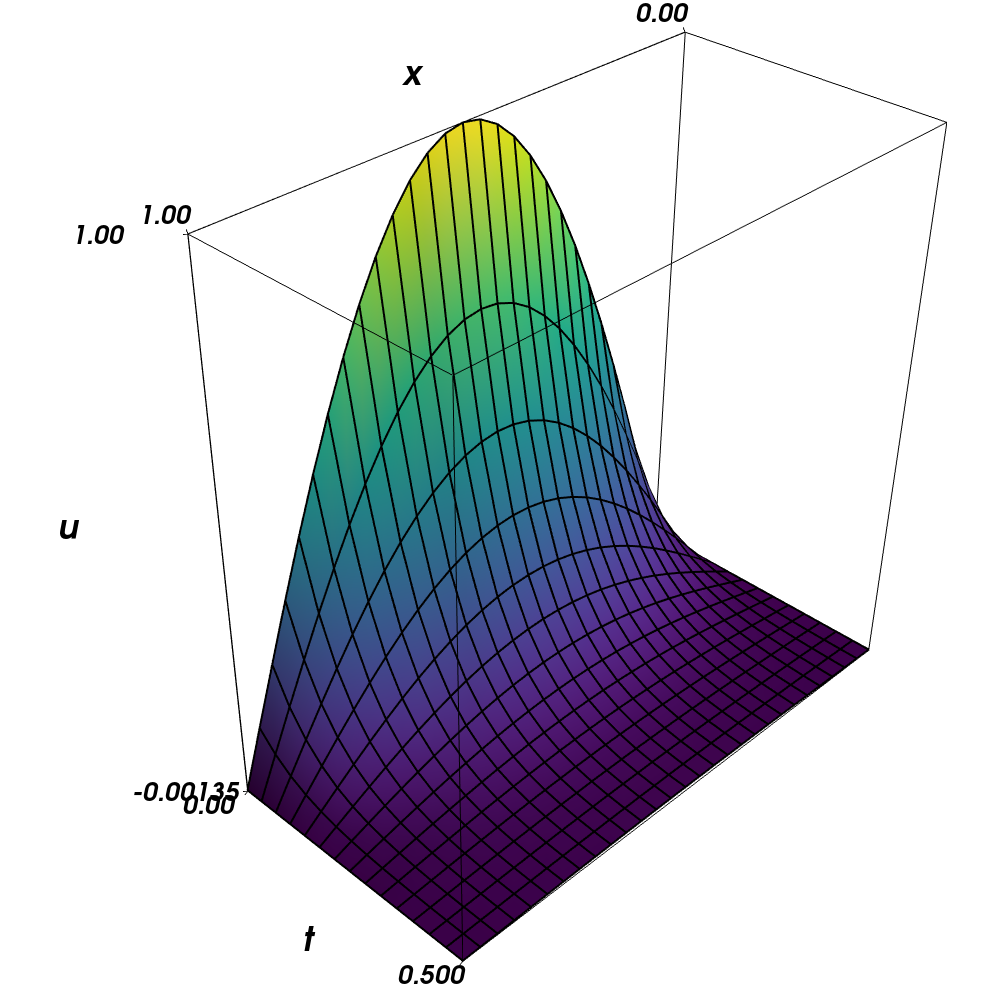}
\caption{Space-time solution of the heat equation \rb{up to a time of $T=0.5$}.}
\label{fig:heat_eqn_st_sol}
\end{subfigure}
\caption{Solution of the one dimensional heat equation using both the space-time version of SPINN and a first order implicit FD-SPINN algorithm.}
\label{fig:heat_eqn}
\end{figure}

\begin{figure}
\begin{subfigure}{0.45\textwidth}
\centering
\includegraphics[width=\textwidth]{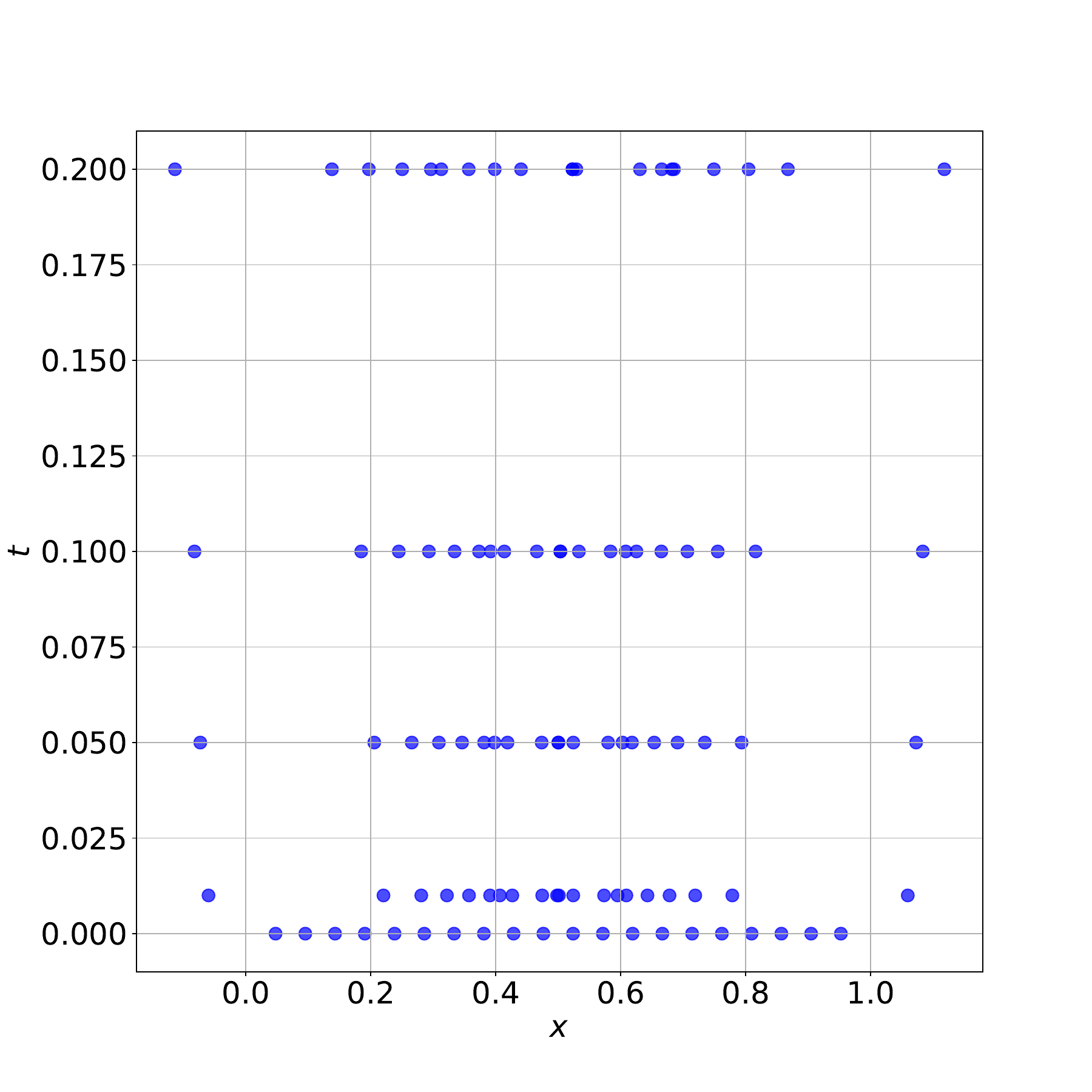}
\caption{First order Implicit FD-SPINN}
\label{fig:heat_fdnodes}
\end{subfigure}
~
\begin{subfigure}{0.45\textwidth}
\centering
\includegraphics[width=\textwidth]{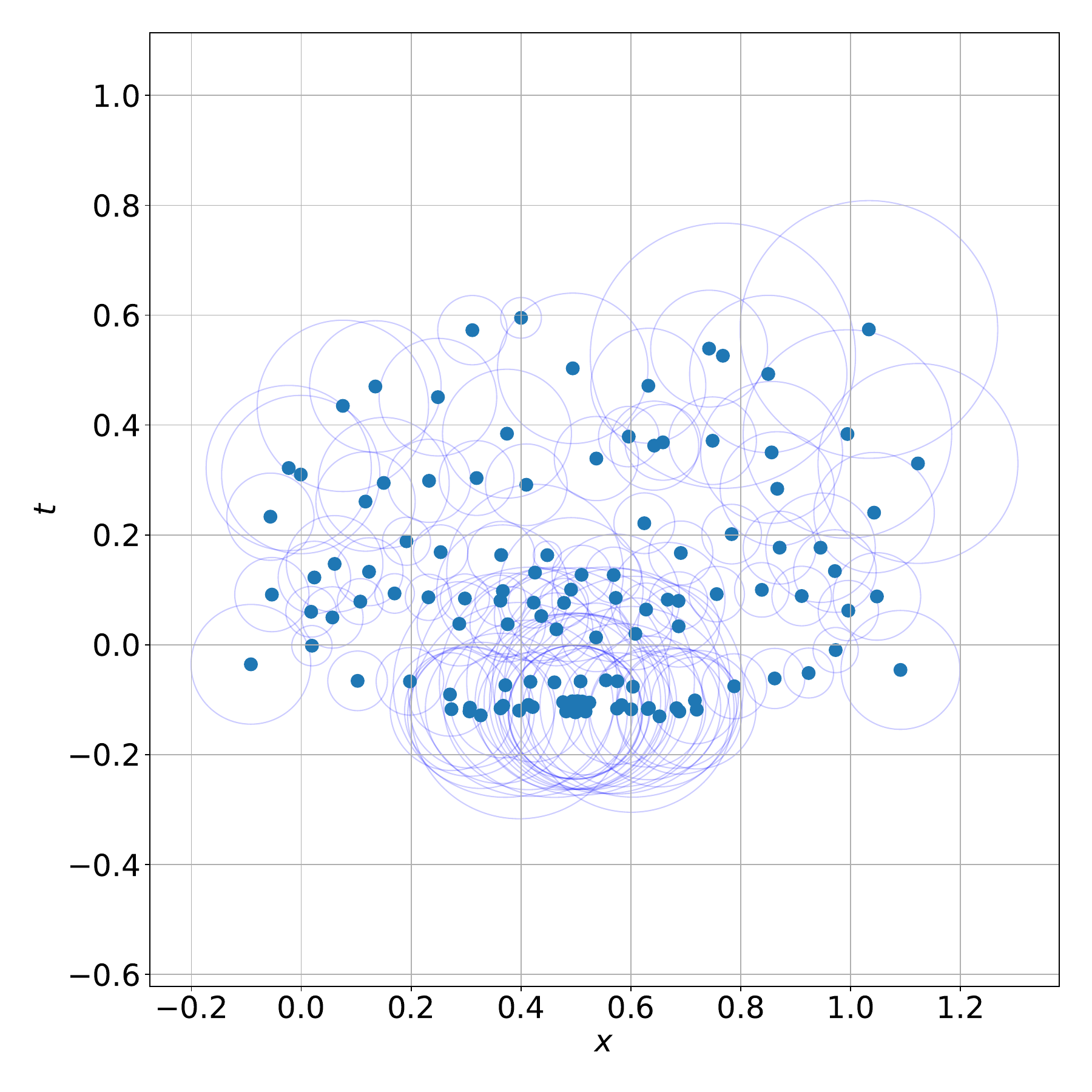}
\caption{Space-time SPINN}
\label{fig:heat_st_nodes}
\end{subfigure}
\caption{Location of the nodes for the heat equation for space-time SPINN and FD-SPINN.}
\label{fig:heat_nodes}
\end{figure}

We show time snapshots of the solution in Figure~\ref{fig:heat_eqn_compare}. We also solve this problem as a space-time PDE by employing SPINN to simultaneously approximate the solution in space and time.  \rb{In this case, we solve the problem until $T=0.5$.  We use 100 internal nodes, and 400 sample points which are fully sampled. The learning rate is $10^{-3}$. We train the network for 10000 iterations.}  \rr{The $L_{\infty}$ error is around $2 \times 10^{-2}$.  Both approaches have similar computational time.} The solution is compared with the exact solution in Figure~\ref{fig:heat_eqn_compare} and the space-time solution is shown in Figure~\ref{fig:heat_eqn_st_sol}; the exact solution is also shown as a wireframe for comparison. In Figure~\ref{fig:heat_nodes}, the final positions of the internal nodes learnt by the space-time SPINN and FD-SPINN methods are shown. \rb{It is seen that the nodes move to capture the correct solution and that this may result in the nodes even moving outside of the domain of the solution.}

\subsubsection{Linear advection equation}
We study next the linear advection equation. The linear advection equation is a classic hyperbolic PDE commonly solved when testing new finite volume algorithms.  We solve the following equation
\begin{displaymath}
\begin{split}
\frac{\partial u}{\partial t} + a \frac{\partial u}{\partial x} &= 0, \quad \quad x \in \mathbb{R}, t \in [0, T],\\
u(x, 0) &= u_0(x), \quad x \in \mathbb{R}\\
u(x, t) &= 0 , \quad |x| \rightarrow \infty.
\end{split}
\end{displaymath}
The exact solution is $u(x, t) = u_0(x - at)$.  We consider a simple Gaussian pulse, $u(x, 0) = e^{-((x + \mu)/2 \sigma)^2}$, where $\mu = 0.3$ and $\sigma=0.15$ and consider the evolution of this, with $a=0.5$ and $T=1$.  Given the almost compact nature of the initial condition we solve the PDE in a finite spatial domain $[-1, 1]$ and over the time interval $[0, 1]$.  Since the initial condition almost vanishes on the boundaries, we set the boundary conditions at $x= -1$ and $x=1$ uniformly to zero.  While this is strictly not correct, we expect the error associated with this to be negligible.  We note however that this restriction can be removed by using the Fourier SPINN model to implement the corresponding spatially periodic model.

As done earlier for the heat equation, we solve the problem using both a first order implicit FD-SPINN as well as a space-time SPINN. \rb{For the FD-SPINN case we use 15 internal nodes, 100 sample points, and a timestep of 0.0025. We iterate either until a loss of $10^{-6}$ is attained, or up to 5000 iterations, using a learning rate of $2.5\times 10^{-3}$.  For the space-time SPINN we use 40 nodes, with 800 sample points, a learning rate of $10^{-3}$ and train the network for 5000 iterations.}

\begin{figure}
\begin{subfigure}{0.32\textwidth}
\includegraphics[width=\textwidth]{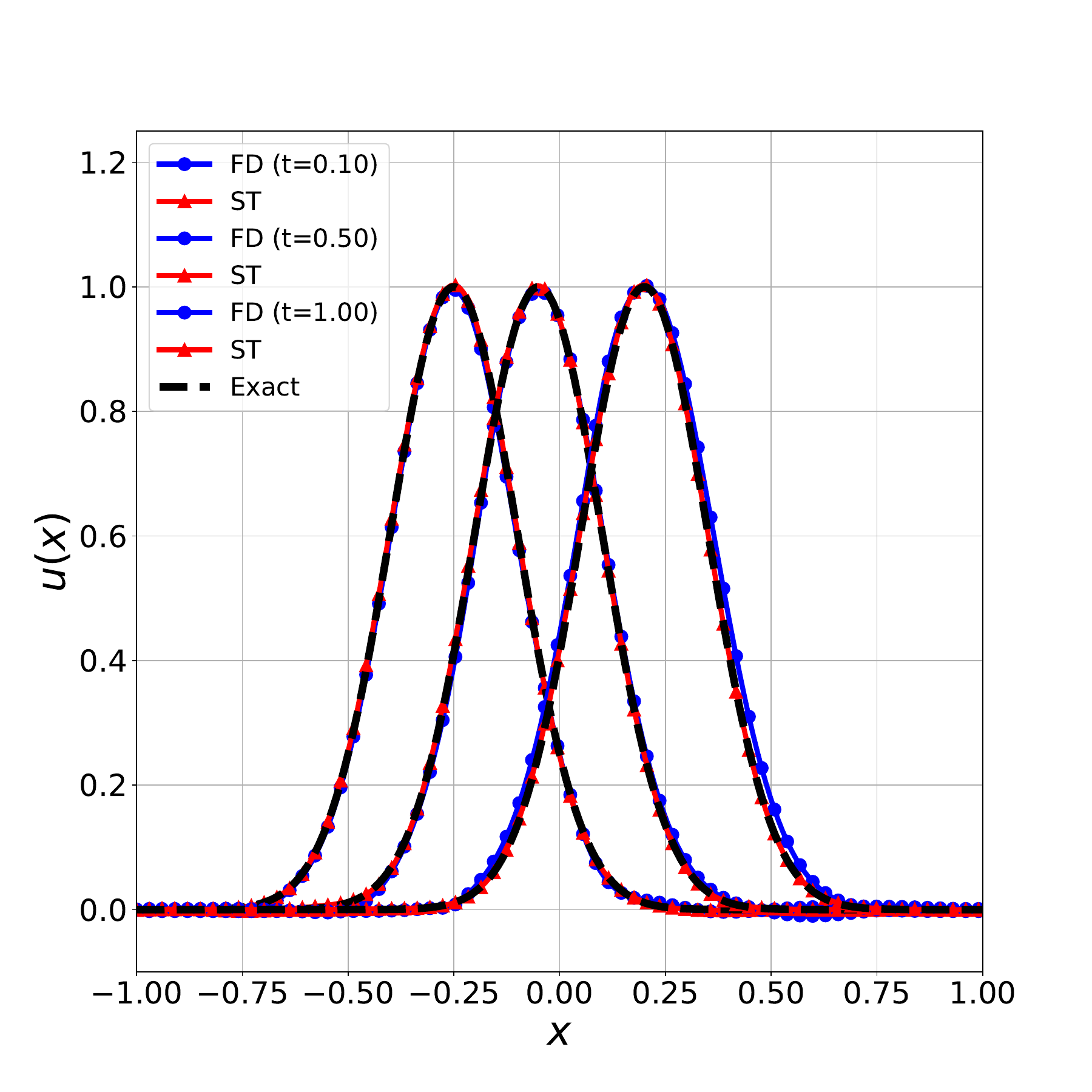}
\caption{Comparison of exact and SPINN solutions}
\label{fig:advection_comp}
\end{subfigure}
~
\begin{subfigure}{0.32\textwidth}
\includegraphics[width=\textwidth]{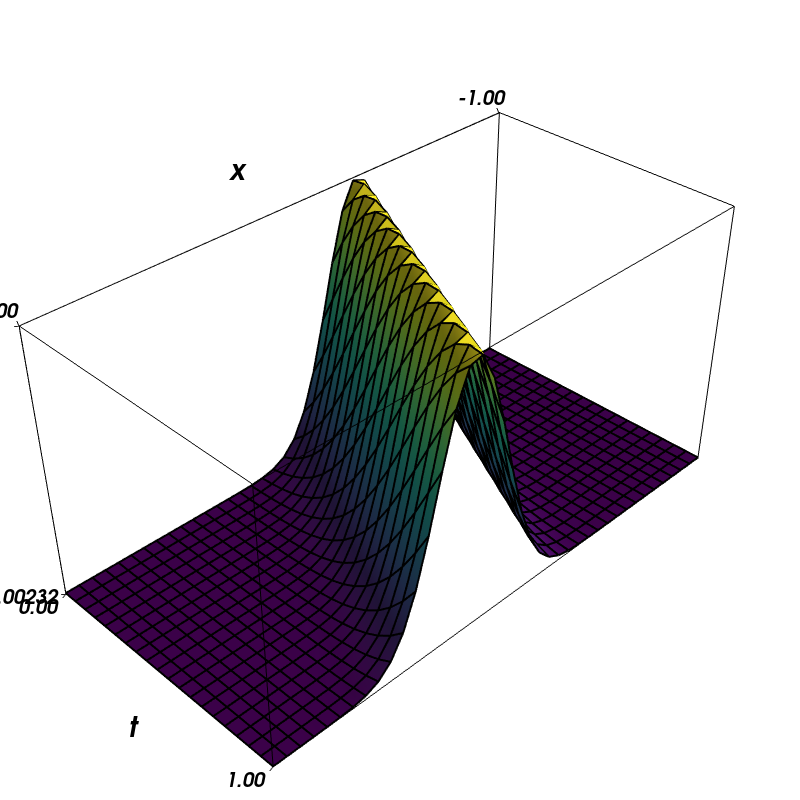}
\caption{Space-time solution}
\label{fig:advection_st}
\end{subfigure}
~
\begin{subfigure}{0.32\textwidth}
\includegraphics[scale=0.2]{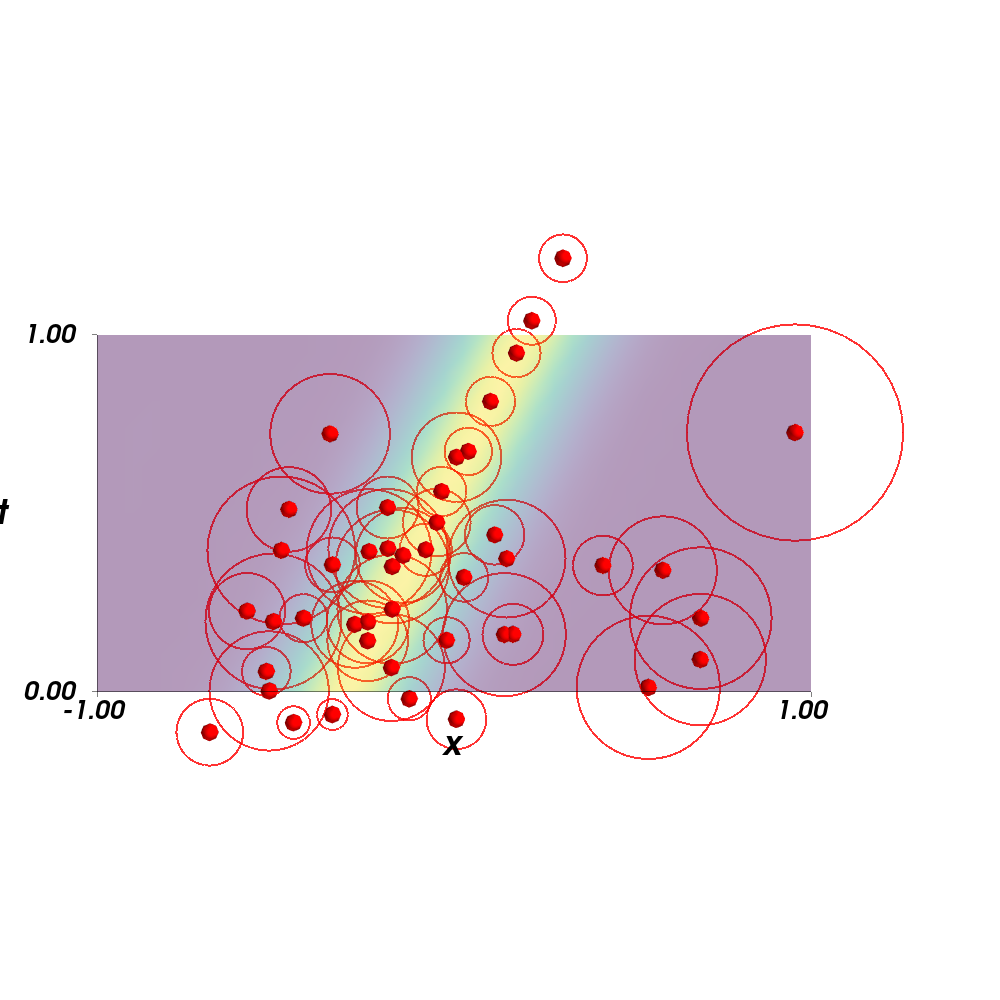}
\caption{Node and kernel width distributions}
\label{fig:advection_nodes}
\end{subfigure}
\caption{Solution of the linear advection equation using both FD-SPINN and space-time SPINN. The alignment of the nodes and the kernel widths as shown in Figure~\ref{fig:advection_nodes} illustrates both the implicit mesh adaptivity and the interpretability of SPINN.}
\label{fig:linear_advection}
\end{figure}

Time snapshots of the solution with a Gaussian pulse as initial condition are shown in Figure~\ref{fig:advection_comp}.  \rr{The $L_\infty$ error at the end of the simulation is $O(10^{-2})$.} In Figure~\ref{fig:advection_st} the space-time solution is compared with the exact solution, shown in wireframe. \rr{The $L_\infty$ error at the end of the simulation is $O(10^{-2})$ for the space-time SPINN as well.}  The location of the interior nodes along with the kernel widths is shown in Figure~\ref{fig:advection_nodes}.  The nodes are initially placed uniformly. It is worth emphasizing that the nodes adapt in time and space to capture the features of the solution, which in this case is a travelling wave. We also point out that widths of the kernel are narrow around the peak of the wave while they are broad away from the peak thereby demonstrating mesh-adaptivity. This also illustrates the interpretability of the SPINN algorithm; the position of the nodes and their corresponding widths, as shown in Figure~\ref{fig:advection_nodes} gives an immediately visual representation of the weights and biases of the SPINN model. We would like to point out that such an interpretation is seldom achievable using conventional methods like PINN.

\rb{We note that both the methods require roughly the same order of computational time.  For problems with smooth solutions, either of the methods is effective.}

\subsubsection{Burgers' equation}
\begin{figure}
\begin{subfigure}{0.5\textwidth}
\includegraphics[width=\textwidth]{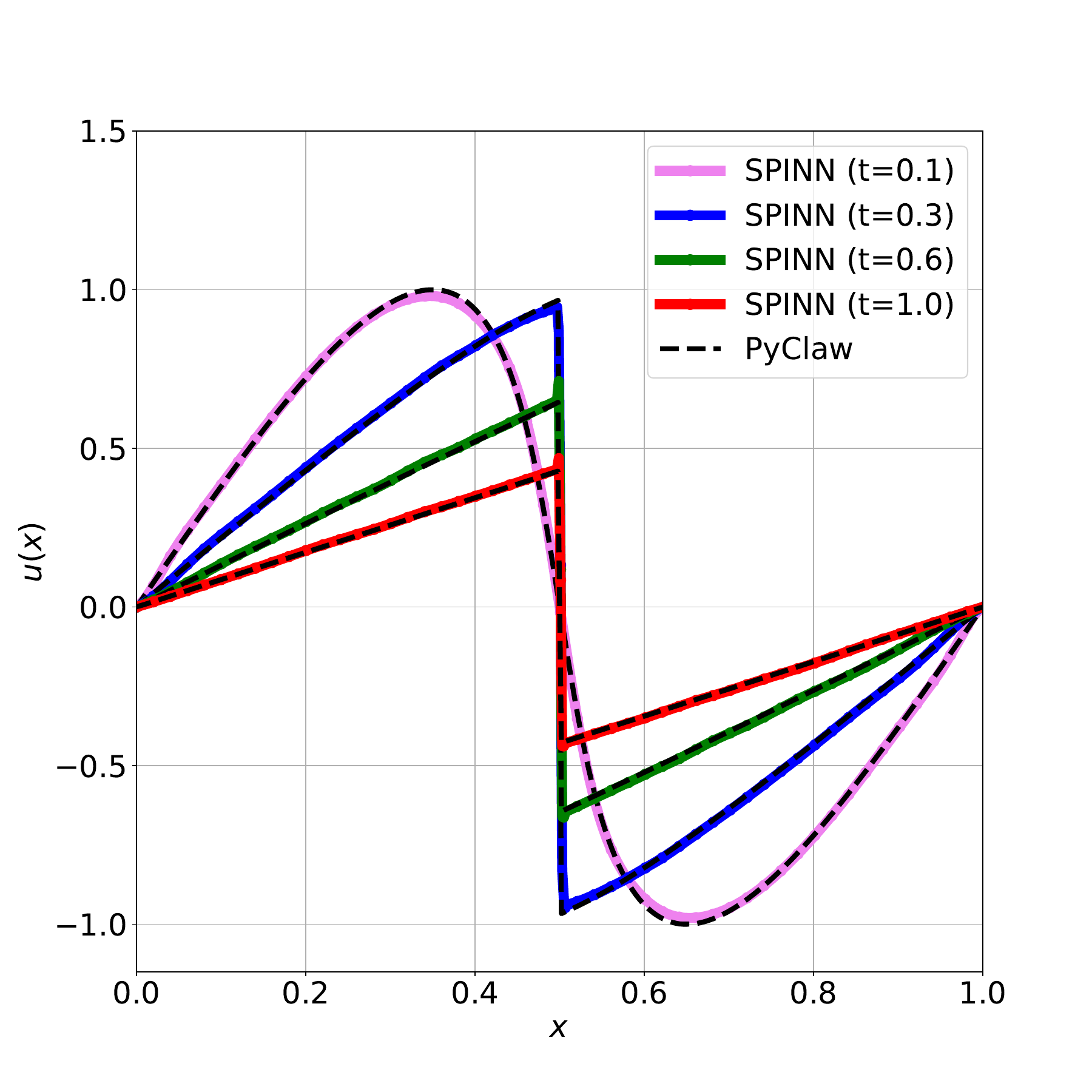}
\caption{FD-SPINN solution compared to reference PyClaw solution}
\label{fig:burgers_comp}
\end{subfigure}
~
\begin{subfigure}{0.5\textwidth}
\includegraphics[width=\textwidth]{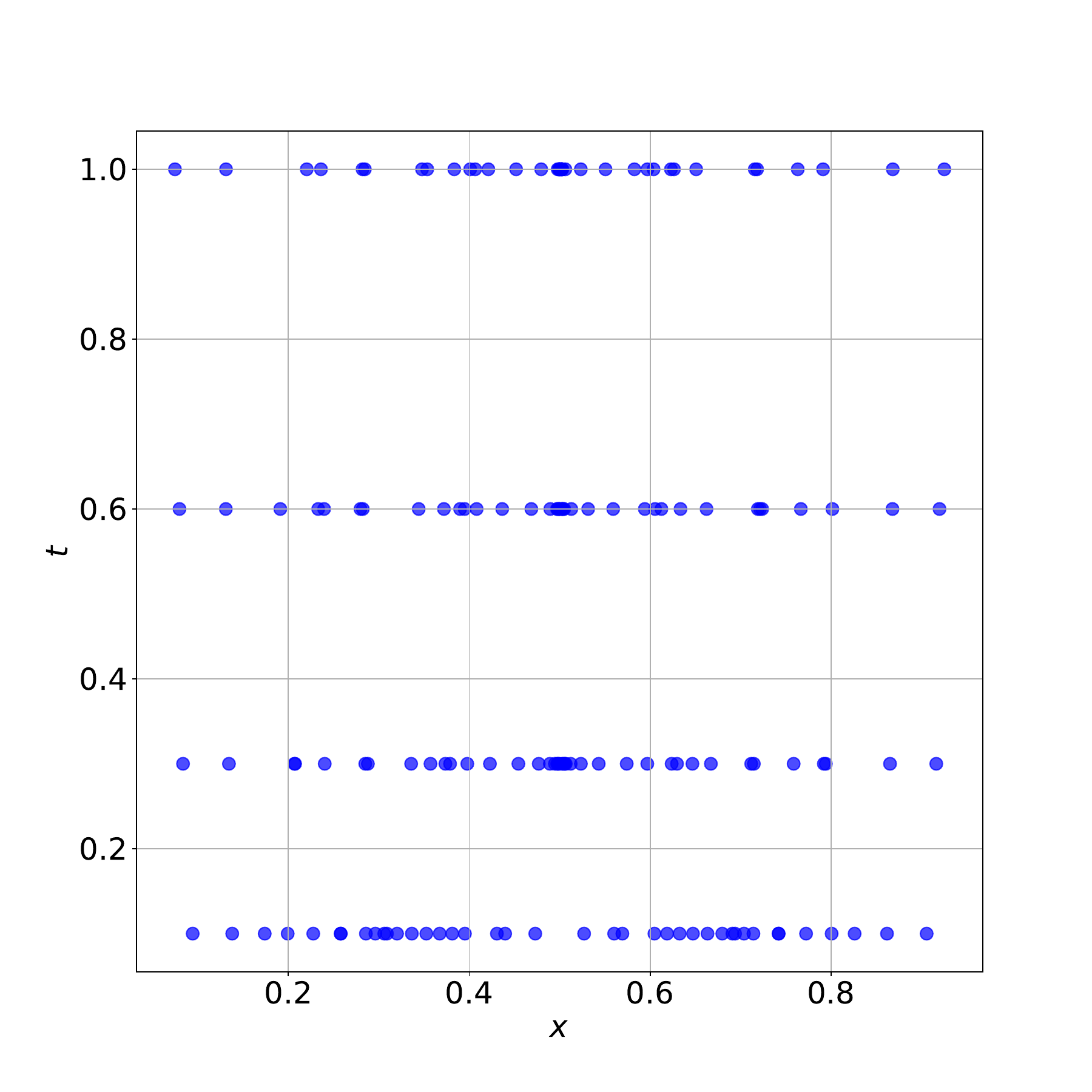}
\caption{Location of nodes for FD-SPINN}
\label{fig:burgers_nodes}
\end{subfigure}
\caption{Comparison of the implicit FD-SPINN solution of the Burgers' equation with reference solution computed using PyClaw for Case1. The location of the nodes is also shown. The shock forms at $x=0.5$; the corresponding clustering of nodes illustrates the adaptivity of the SPINN algorithm.}
\label{fig:inviscid_burgers}
\end{figure}

\begin{figure}
\begin{subfigure}{0.5\textwidth}
\includegraphics[width=\textwidth]{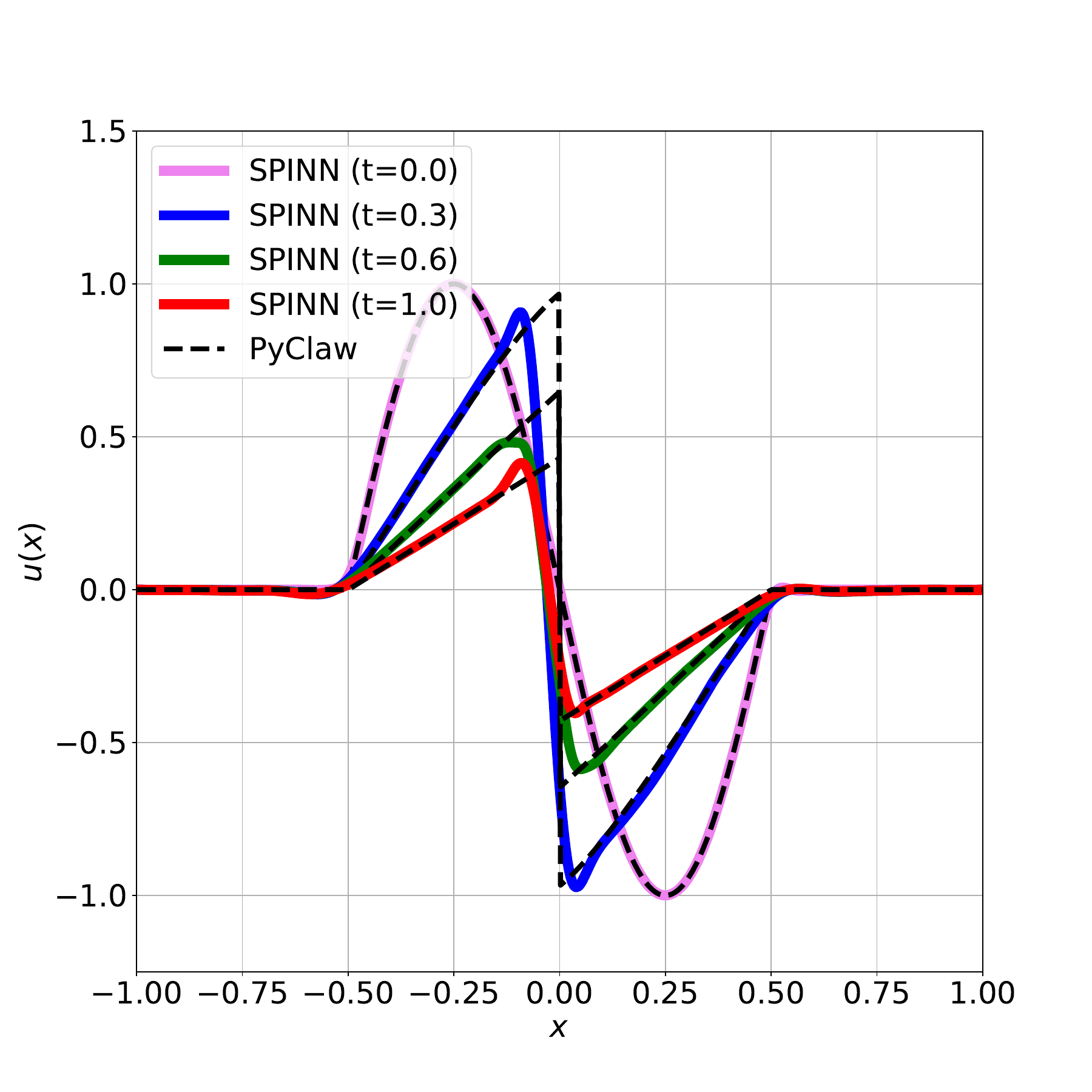}
\caption{Space-time SPINN solution for Case1}
\label{fig:burgers_st_case1}
\end{subfigure}
~
\begin{subfigure}{0.5\textwidth}
\includegraphics[width=\textwidth]{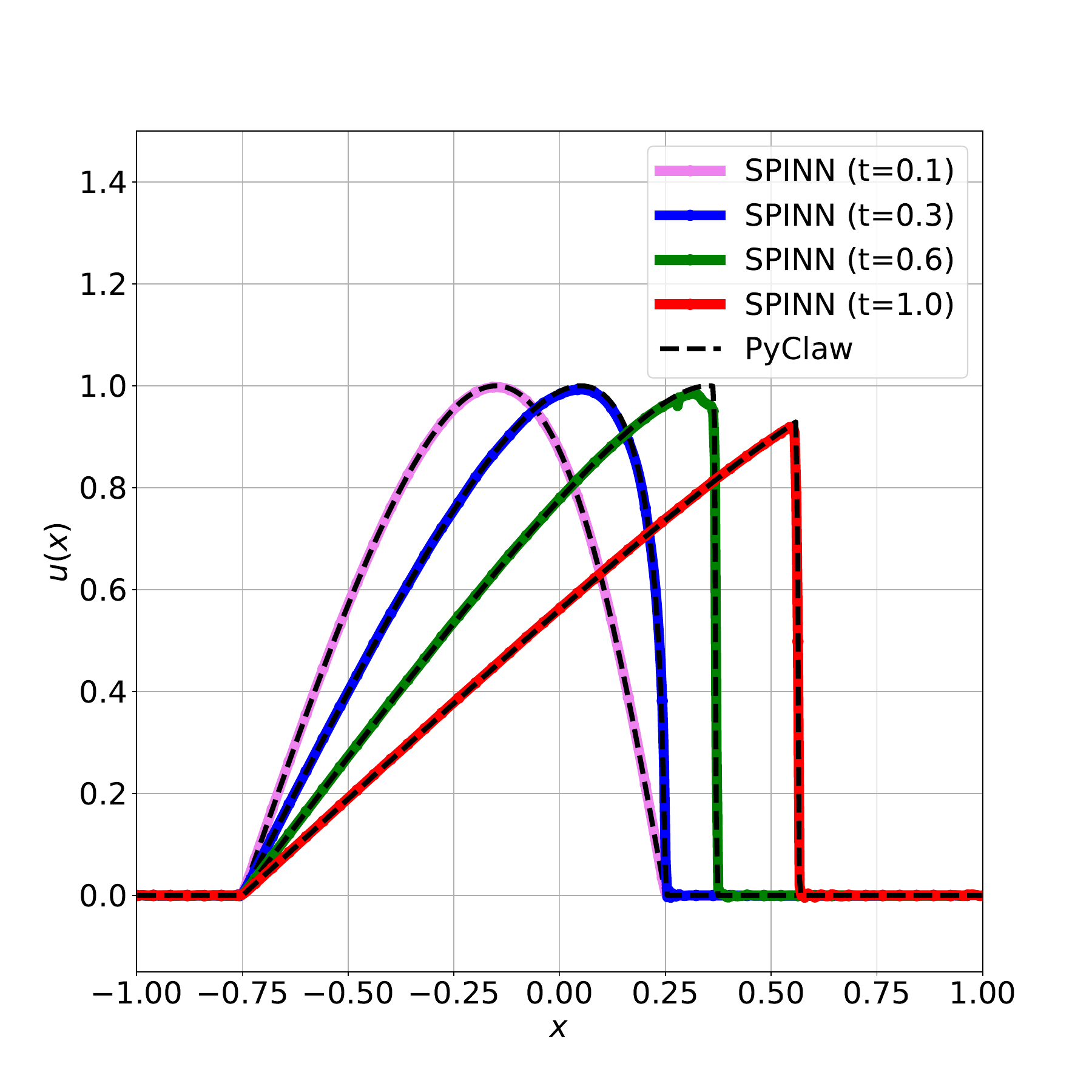}
\caption{FD-SPINN solution for Case2}
\label{fig:burgers_fd_case2}
\end{subfigure}
\caption{Comparison of the space-time SPINN solution for Case1 and the FD-SPINN solution for Case2 of the Burgers' equation with reference solution computed using PyClaw.}
\label{fig:inviscid_burgers_mix}
\end{figure}

We next consider a classic non-linear, time-dependent hyperbolic PDE in one spatial dimension, namely the inviscid Burgers' equation. The inviscid and viscous Burgers' equations are important hyperbolic PDEs that arise in many applications.  This problem is interesting because it is non-linear, and develops a shock (a discontinuity) in finite time for initially smooth solutions. The inviscid Burgers' equation along with the initial condition reads,
\begin{displaymath}
\begin{split}
\frac{\partial u}{\partial t} + u \frac{\partial u}{\partial x} &= 0,  \quad x \in [0, 1], t \in [0, T],\\
u(x, 0) &= \sin(2 \pi x),  \quad x \in [0, 1]\\
u(0, t) = u(1, t) &= 0.
\end{split}
\end{displaymath}
\rb{We call this ``Case1''.} We solve the problem using first order implicit FD-SPINN using Gaussian kernel with 40 internal nodes. \rb{We use $400$ sampling points (with full sampling) and a learning rate of $10^{-4}$. We iterate upto a loss of $10^{-6}$ or $5000$ steps for each timestep and employ a timestep of $0.01$.}  Time snapshots of the solution at different times are shown in Fig.~\ref{fig:burgers_comp} and compared against reference solution obtained using PyClaw~\cite{pyclaw}, a popular and high-quality finite volume package. A shock develops at $x=0.5$ at around $t=0.25$ which is clearly captured well by SPINN. \rR{The $L_1$ error is $4.7 \times 10^{-3}$ and this is computed with reference to the PyClaw solution.} We show in Figure~\ref{fig:burgers_nodes} the position of the nodes learnt by the FD-SPINN algorithm.  It can be seen that the nodes initially cluster around the peaks of the sine function but cluster around $x=0.5$, which is the location of the shock for $t> 0.2$ thereby demonstrating the nodal adaptivity implicit in the SPINN algorithm. We also point out that the widths of the kernel at the shock locations are much smaller than the corresponding kernels centered at nodes away from the shock. What is remarkable is that \new{even for a fully inviscid problem as we consider here}, and despite using smooth kernels, in this case the Gaussian, the FD-SPINN method is able to capture the shock accurately. This thus demonstrates (i) the adaptivity of the SPINN method; the nodes closer to the shock front have a much smaller kernel width in comparison to nodes away from the shock as expected, and (ii) the ability of SPINN to capture discontinuities in the solution. \rb{We present a space-time solution for Case1 in Fig.~\ref{fig:burgers_st_case1} where we employ $400$ internal nodes, $2000$ sampling points, a learning rate of $10^{-3}$, and train for $10^4$ iterations.  The results are not as accurate as the FD-SPINN and display some dissipation \rR{and the $L_1$ error is $1.8 \times 10^{-2}$}.  We believe that this is a consequence of an insufficient number of internal nodes.  The solution does capture the overall features quite well.  We observe that there is no viscosity in this problem; indeed, the solution does display on occasion some oscillatory behavior for both FD-SPINN and ST-SPINN.  Since FD-SPINN allows us to employ a limited number of nodes more effectively, it is much more efficient in capturing the correct solution with a smaller number of nodes.}

\begin{figure}
\centering
\begin{subfigure}{0.45\textwidth}
\includegraphics[width=\textwidth]{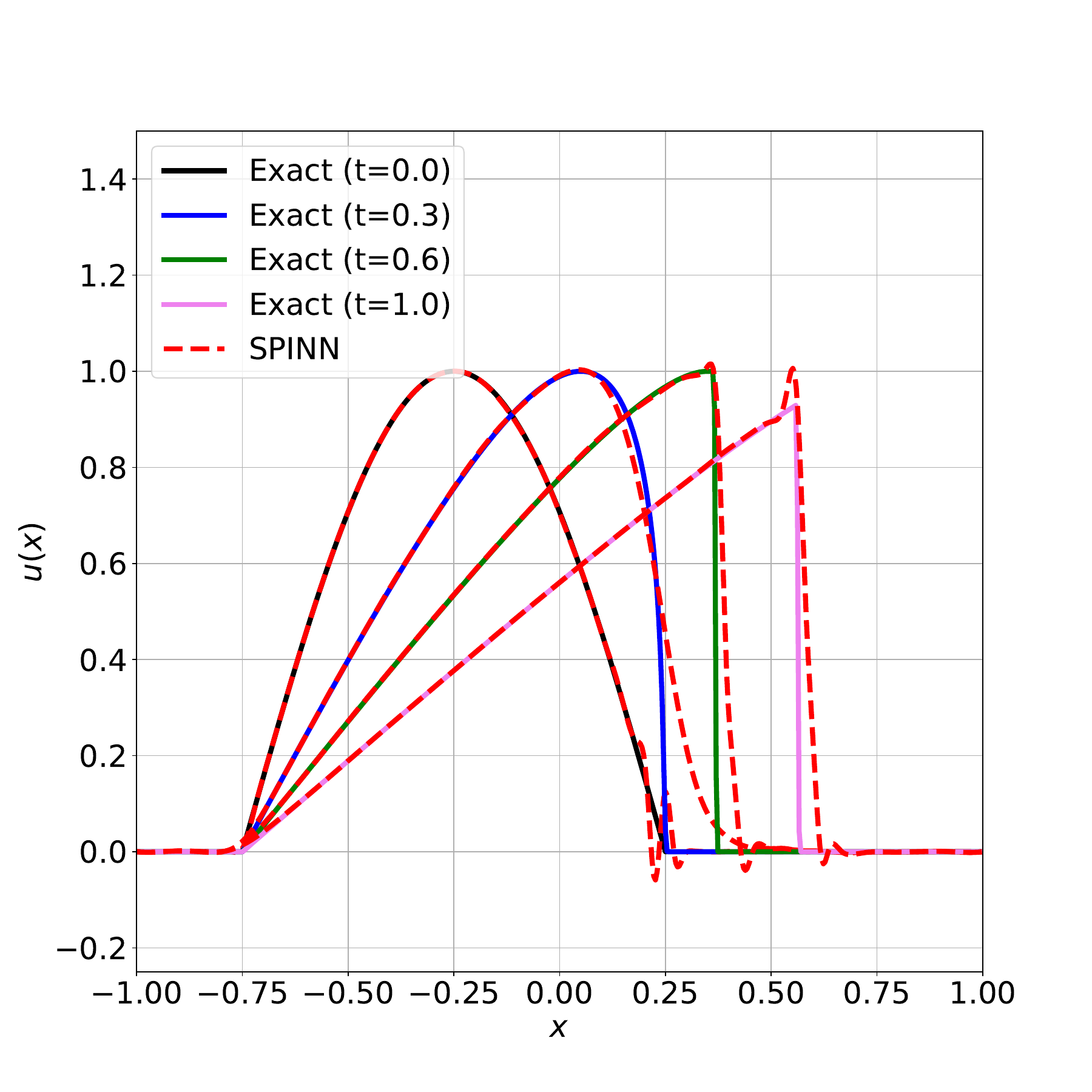}
\caption{Space-time SPINN solution}
\label{fig:spinn_burgers_a}
\end{subfigure}
~
\begin{subfigure}{0.5\textwidth}
\includegraphics[width=\textwidth]{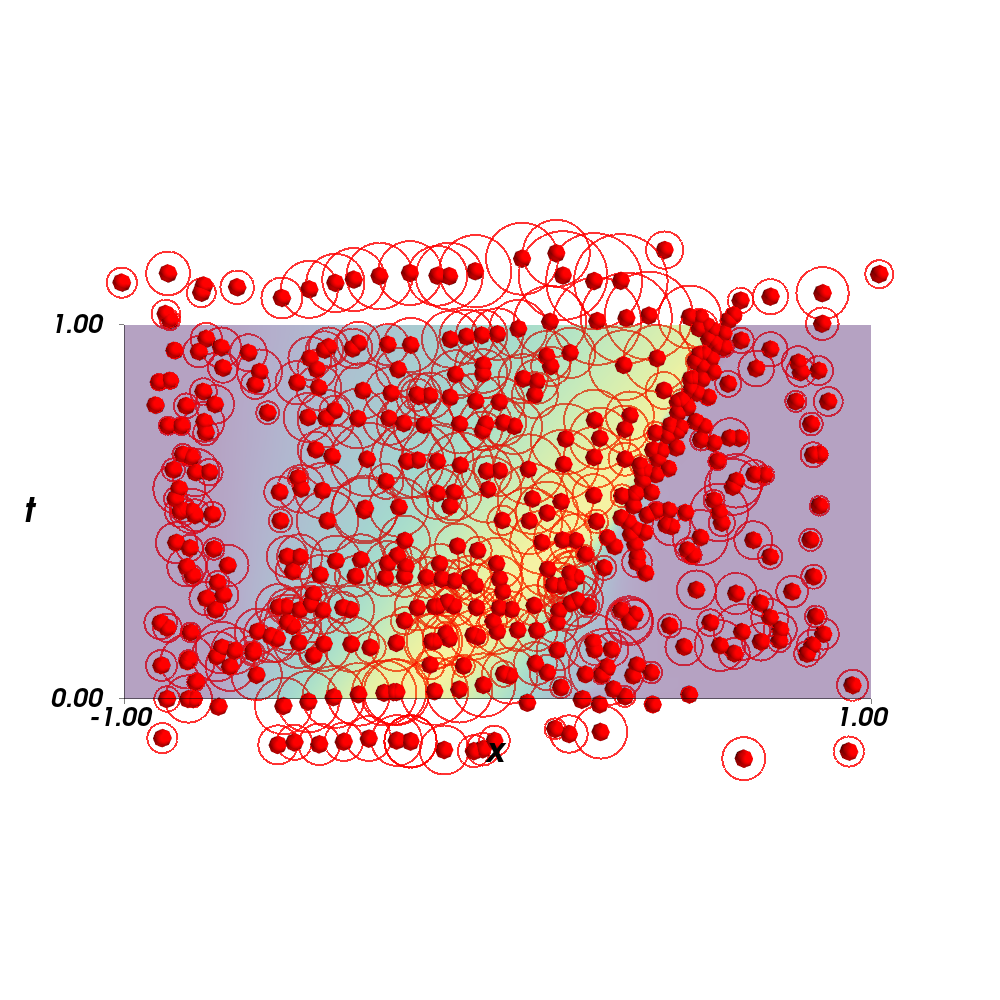}
\caption{Location and width of nodes}
\label{fig:spinn_burgers_b}
\end{subfigure}
\caption{Solution of the inviscid Burgers' equation for Case2 using space-time SPINN with Gaussian kernel and about $400$ nodes.}
\label{fig:spinn_burgers}
\end{figure}

We also present a space-time solution for the Burgers' equation over the domain $x \in [-1, 1]$ using the following initial condition:
\begin{displaymath}
u_0(x) = \begin{cases}
\sin(\pi(x + 0.75)), & -0.75 \le x \le 0.25,\\
0, & \text{otherwise}.
\end{cases}
\end{displaymath}
\rb{We call this ``Case2''.} This problem also develops a discontinuity. The solution obtained using the space-time SPINN version with a Gaussian kernel is shown in Figure~\ref{fig:spinn_burgers_a} and compared with a reference simulation again obtained using PyClaw~\cite{pyclaw}. \rb{We employ around 400 nodes in this case, with a learning rate of $2\times 10^{-3}$, 2000 sampling points, and iterate for 10000 steps.} The corresponding distribution of nodes and their widths are shown in Figure~\ref{fig:spinn_burgers_b}. \rR{The $L_1$ error is $1.2 \times 10^{-2}$.} As remarked earlier in the case of linear advection, we see that the nodes align along the shock front, and the widths of the nodes are correspondingly smaller near the shock front to ensure that the shock is captured adequately. Note also that the alignment of the nodes start after $t \simeq 0.5$, which is when the shock forms. It can also be seen that the solution captures the correct shock speed. However, despite the large number of nodes used, we note that there are some oscillations and diffuse behavior near the shock front, unlike the FD-SPINN case. We expect these problems to be solved when the number of nodes is increased, or when a small non-zero artificial viscosity is introduced; we will be investigating this in more detail in a future work. \rr{We however point out that unlike works like \cite{RPK2019} where the viscous Burgers equation is solved, albeit with a small viscosity, we solve the fully inviscid problem and still get very good results. With the addition of a small viscosity, we expect the performance of both FD-SPINN and space-time SPINN to be better.} \rb{It is important to note that even the FD-SPINN solution can display oscillations in these cases - the tolerance for iterations has to be set low enough that these are removed.  We present a FD-SPINN solution for Case2 in Fig.~\ref{fig:burgers_fd_case2}.  The results are in good agreement although at $t=0.6$ some slight oscillations can be seen.  This case was run with $150$ nodes, $600$ sampling points, a learning rate of $10^{-4}$, and trained each timestep upto a tolerance of $10^{-7}$ or 10000 iterations with a timestep of 0.005.}  \rR{The $L_1$ error is $1.9 \times 10^{-3}.$} \rb{As can be seen, the new method produces very good results even for an inviscid problem.}

\subsubsection{Allen-Cahn equation}
\label{subsec:allencahn}

\begin{figure}
\centering
\includegraphics[width=\textwidth]{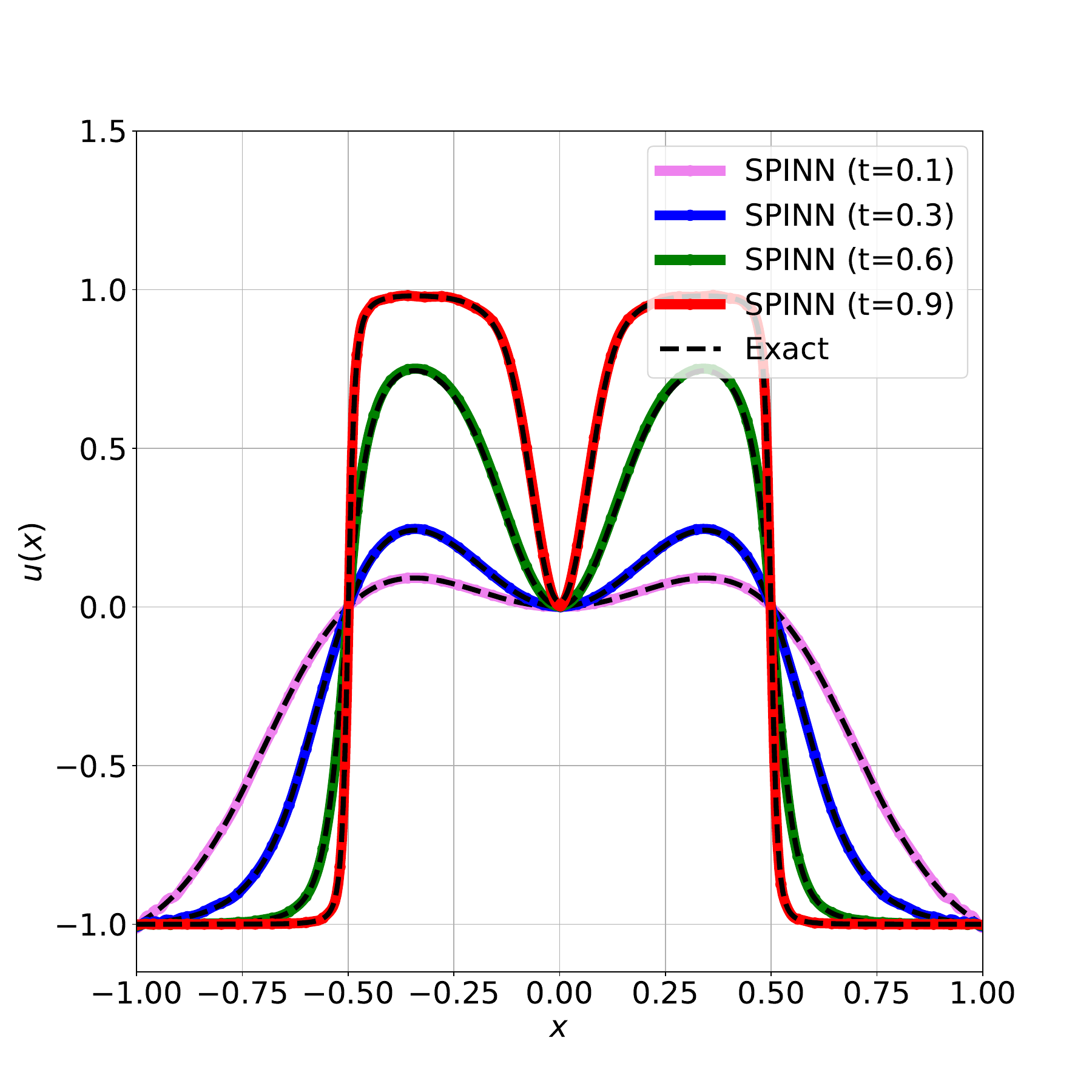}
\caption{FD SPINN solution for the Allen-Cahn problem compared with the exact solution available from \cite{RPK2019}.}
\label{fig:spinn_allen_cahn}
\end{figure}

\rr{
We present next a solution of the following Allen-Cahn equation with periodic boundary conditions:
\begin{equation} \label{eq:allen_cahn}
\begin{split}
\frac{\partial u(x,t)}{\partial t} - 0.0001\frac{\partial^2 u(x,t)}{\partial x^2} + 5u(x,t)^3 - 5u(x,t) & = 0, \quad x \in (-1,1), \; t \in (0,1),\\
u(x,0) &= x^2 \cos \pi x,\\
u(-1,t) &= u(1,t)\\
u_x(-1,t) &= u_x(1,t).
\end{split}
\end{equation}
We consider this PDE for two reasons: (i) to illustrate the fact that periodic boundary conditions can be handled easily with SPINN, and (ii) to make a comparison with the PINN solution for the same equation as carried out in \cite{RPK2019}. We solve the Allen-Cahn PDE \eqref{eq:allen_cahn} using the implicit finite difference version of SPINN. We use 100 nodes, with 500 samples, and a learning rate of $10^{-3}$, we iterate until a loss of $10^{-6}$ is reached or 5000 iterations are complete and use a timestep of 0.005. This problem completes execution on a CPU in about 4 minutes. We compare the result with the solution made available in \cite{RPK2019}. The results are shown in Fig.~\ref{fig:spinn_allen_cahn}.  The loss is consistently less than $10^{-6}$ and eventually is around $10^{-8}$.} \rR{The $L_1$ error is $4.4 \times 10^{-3}$ which we compute using the reference solution given in \cite{RPK2019}. } \rr{This shows the excellent accuracy obtained for this problem with only 100 SPINN nodes.}

\subsection{Fluid dynamics}
\label{subsec:ns}

Our final example involves solving the steady incompressible viscous Navier-Stokes equations in two spatial dimensions. We consider the classic lid-driven cavity problem~\cite{ldc:ghia}. Consider a unit square placed on the x-axis in the region $[0, 1] \times [0, 1]$ and filled with a fluid with a density $\rho=1$.   Let the Cartesian velocity components be given by $(u, v)$ and the pressure of the fluid be $p$.  The governing differential equations for the fluid when it attains a steady flow is,
\begin{displaymath}
\begin{split}
\frac{\partial u}{\partial x} + \frac{\partial u}{\partial y} &= 0, \\
u \frac{\partial u}{\partial x} + v \frac{\partial u}{\partial y} &= -\frac{1}{\rho} \frac{\partial p}{\partial x} + \nu \nabla^2 u,\\
u \frac{\partial v}{\partial x} + v \frac{\partial v}{\partial y} &= -\frac{1}{\rho} \frac{\partial p}{\partial y} + \nu \nabla^2 v,\\
\end{split}
\end{displaymath}
where $\nu$ is the kinematic viscosity of the fluid.  The first equation is the conservation of mass (also called the continuity equation) and the subsequent two are the momentum equations. The boundary conditions are given as,
\begin{displaymath}
\begin{split}
    u(x, 0) &= 0\\
    u(x, 1) &= 1\\
    u(0, y) &= 0 \\
    u(1, y) &= 0 \\
    \frac{\partial p}{\partial n} &= 0,
\end{split}
\end{displaymath}
where $n$ is the normal vector along the boundary. The velocity $v$ is zero on the boundary of the square. The walls are modeled with no-slip boundary conditions consistent with the behavior of a viscous fluid. The Reynolds number, $Re$ for this problem is defined as $Re = 1/\nu$. There is no exact solution for this problem but there are many numerical solutions available.  We compare the horizontal and vertical profile of the velocity along the center-line of the square with the classic results of \cite{ldc:ghia} at a Reynolds number of 100.

\begin{figure}
\includegraphics[width=\textwidth]{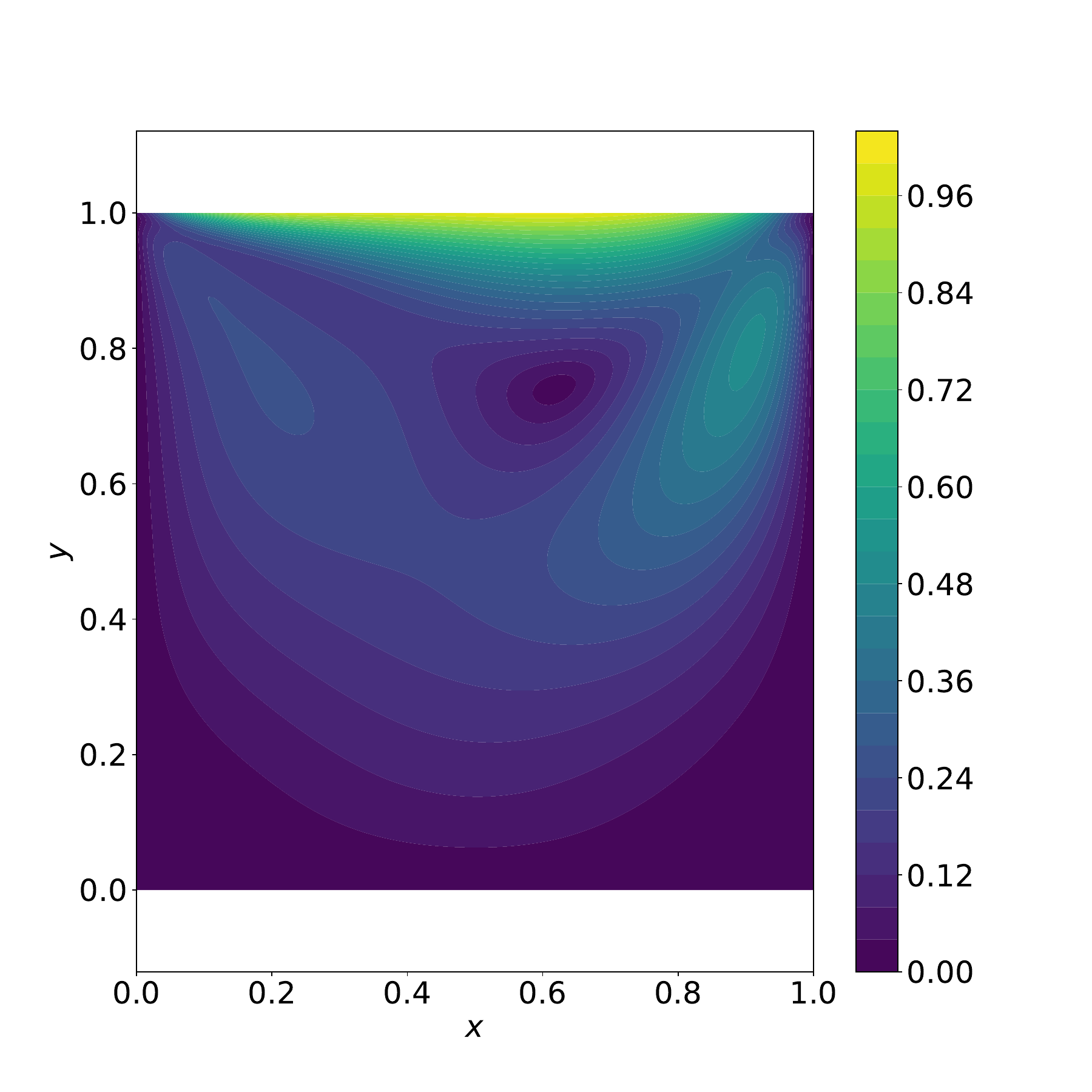}
\caption{Velocity magnitude of flow inside the driven cavity
 with a fluid having kinematic viscosity $\nu=0.01$ using SPINN with a Gaussian kernel.
}
\label{fig:ldc_velocity}
\end{figure}

\begin{figure}
\begin{subfigure}{0.49\textwidth}
\includegraphics[width=\textwidth]{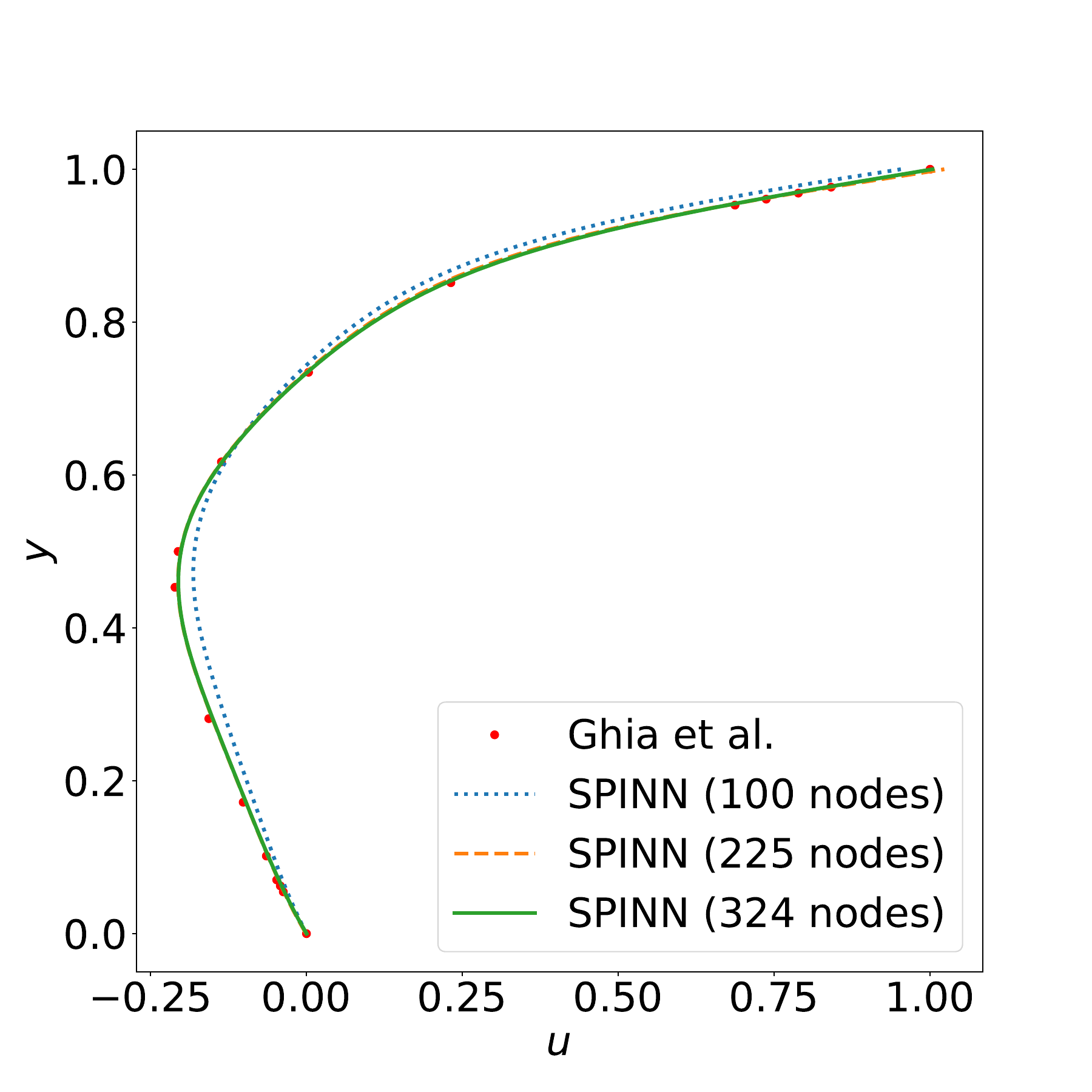}
\caption{$u$ versus $y$ obtained using SPINN.}
\label{fig:ldc_u_vs_y}
\end{subfigure}
~
\begin{subfigure}{0.49\textwidth}
\includegraphics[width=\textwidth]{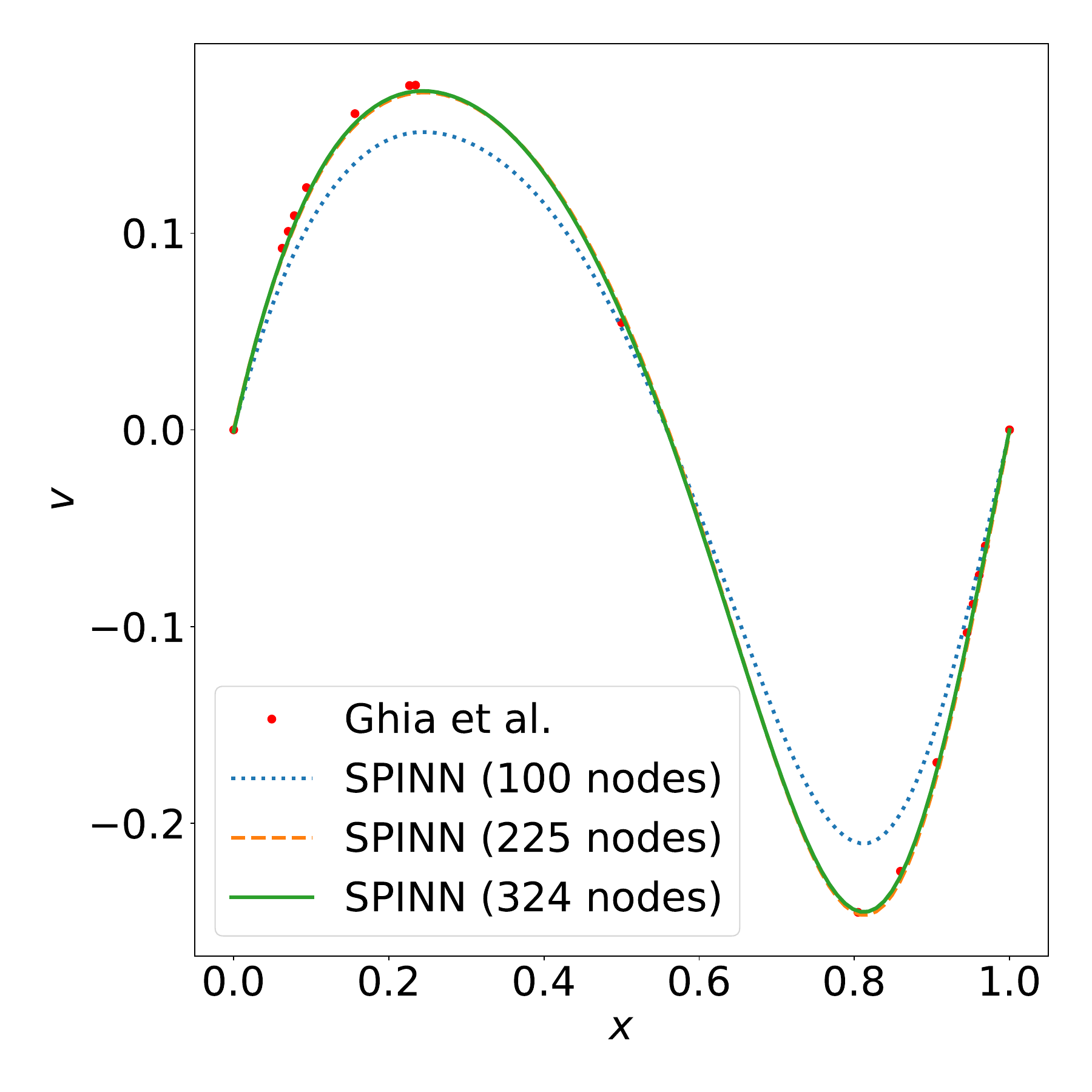}
\caption{$v$ versus $x$ obtained using SPINN.}
\label{fig:ldc_v_vs_x}
\end{subfigure}
~
\begin{subfigure}{0.49\textwidth}
\includegraphics[width=\textwidth]{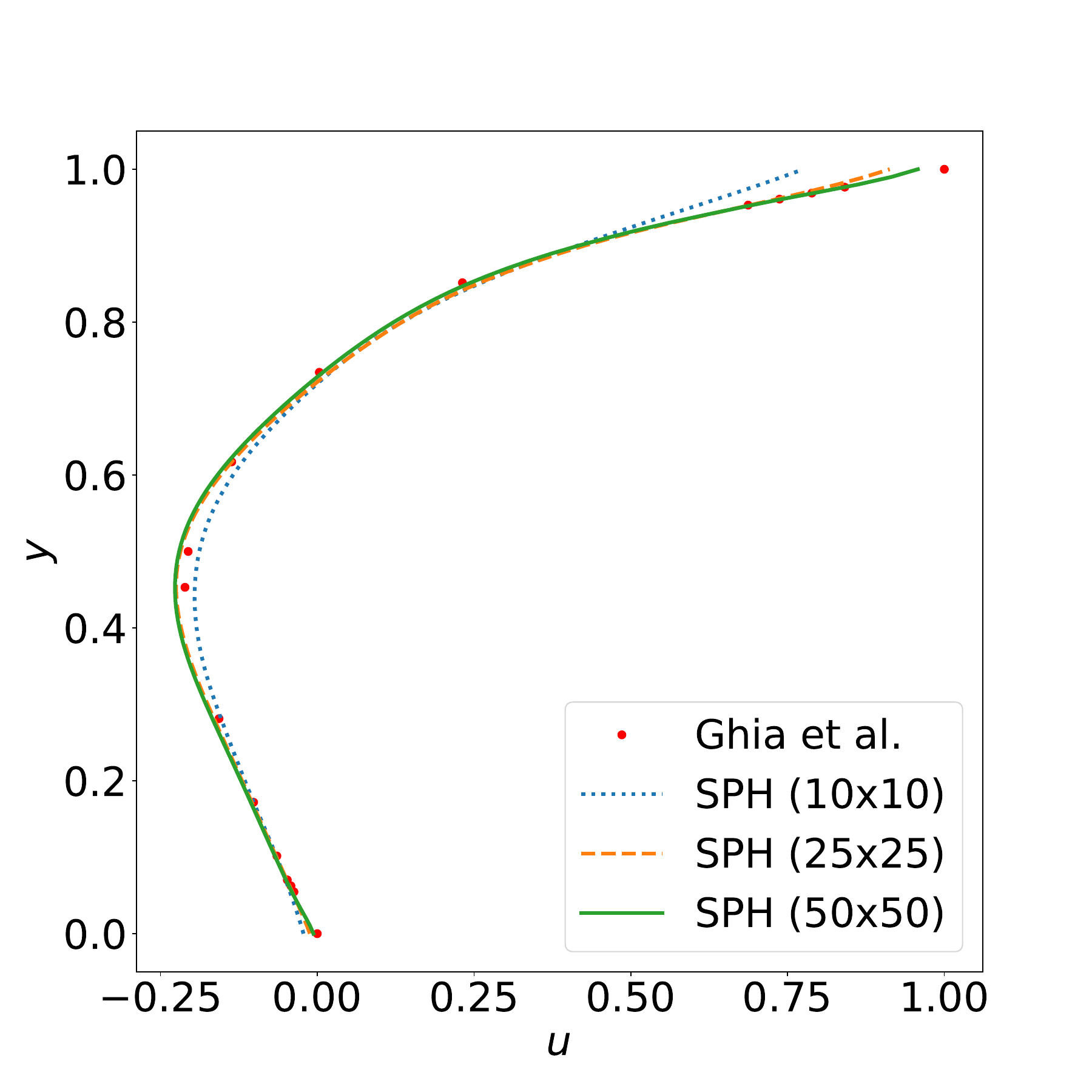}
\caption{$u$ versus $y$ obtained with an SPH scheme.}
\label{fig:ldc_u_vs_y_pysph}
\end{subfigure}
~
\begin{subfigure}{0.49\textwidth}
\includegraphics[width=\textwidth]{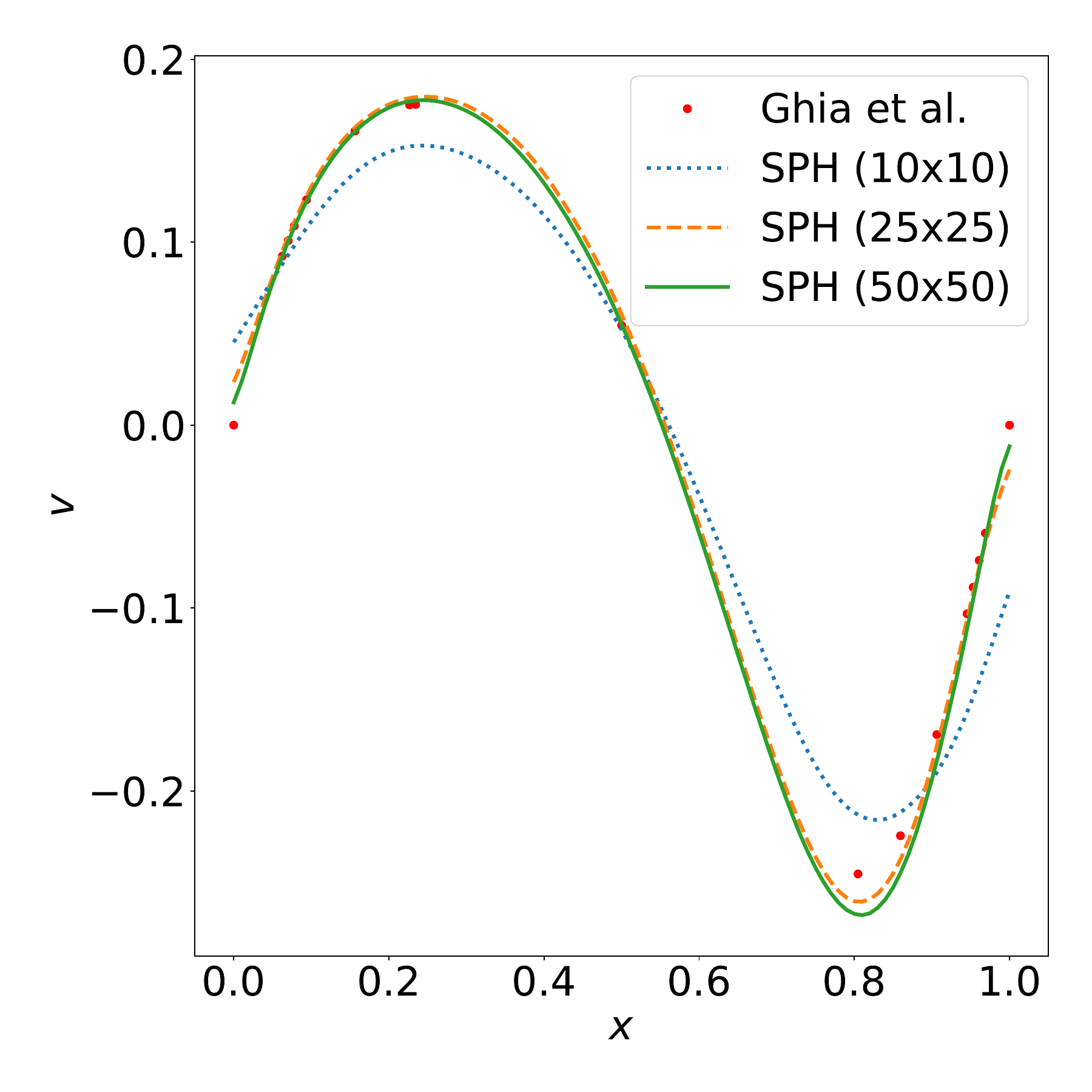}
\caption{$v$ versus $x$ obtained with an SPH scheme.}
\label{fig:ldc_v_vs_x_pysph}
\end{subfigure}

\caption{\rb{Comparison of the centerline velocity obtained for the lid-driven cavity problem using SPINN and with an SPH scheme as compared with \cite{ldc:ghia}.}}
\label{fig:ldc}
\end{figure}

The solution of this problem requires that SPINN returns a vector representing the solution $(u, v, p)$ at each point where the solution is desired, unlike the previously studied scalar differential equations. The magnitude of the velocity computed using SPINN \rb{with a Gaussian kernel} is shown in Fig.~\ref{fig:ldc_velocity}.  In Figs.~\ref{fig:ldc_u_vs_y} and ~\ref{fig:ldc_v_vs_x} the velocity profile along the center-lines are compared with those of \cite{ldc:ghia} as the number of internal nodes is varied.  We obtain good results with around 225 internal nodes for this problem.  \rb{The optimizer is run for 10000 iterations with a learning rate of $5 \times 10^{-4}$.} This example was chosen to demonstrate the ability of SPINN to model a non-linear system of PDEs that frequently arise in the modeling of physical systems. \rb{In Figs.~\ref{fig:ldc_u_vs_y_pysph} and ~\ref{fig:ldc_v_vs_x_pysph} the velocity profiles are computed using an SPH simulation using an initial  $10 \times 10, 25 \times 25$ and  $50\times 50$ grid of particles for the fluid and compared with the results of \cite{ldc:ghia}.  These simulations were made using the standard cavity example available as part of the open source PySPH package~\cite{pysph2020}.  The lid driven cavity problem has been simulated using the EDAC-SPH scheme~\cite{edac-sph:cf:2019} where the NS equations are evolved in time and driven to a steady state.
As can be seen by a close inspection of the results, the solution obtained using SPINN with 225 nodes is more accurate than that obtained by the SPH simulation using 2500 particles.  This suggests that SPINN behaves like a higher-order method.}

\rr{We report the $L_{\infty}$ error by using the data provided in \cite{ldc:ghia} as a reference.  We compute the $u$ and $v$ velocities along the centerline from the SPINN and PySPH solutions and then compute the values at the corresponding location from \cite{ldc:ghia}.  We compute the $L_{\infty}$ error using these values.  We show the computed values in Table~\ref{tab:cavity_error}.  As can be clearly seen, the results of SPINN are very accurate even with a very small number of nodes.}

\begin{table}
\centering
\begin{tabular}{lrr}
\toprule
  Method      &   $L_{\infty}$ error:  $u$ &   $L_\infty$ error: $v$ \\
\midrule
  SPINN 100 nodes & 0.073718 & 0.035425 \\
  SPINN 200 nodes  & 0.022839 & 0.006325 \\
  SPINN 300 nodes & 0.012010 & 0.004969 \\
PySPH 50x50 particles & 0.042251 & 0.022198 \\
\bottomrule
\end{tabular}
\caption{Comparison of $L_\infty$ error of SPINN and PySPH solution of centerline velocity components with respect to the solution reporeted in \cite{ldc:ghia}.}
\label{tab:cavity_error}
\end{table}

\rb{The algorithms used for SPINN and SPH are considerably different, however, a comparison of the execution times provides some insight into the relative computational performance of the methods. We note that PySPH employs a highly optimized implementation whereas the SPINN implementation is largely a proof of concept and in this particular case inefficiently computes the influence of all nodes at a given point.  The SPINN case with 225 nodes and 10000 iterations takes around 825 seconds to complete.  The SPH simulation with 2500 particles for 33000 iterations takes 517 seconds of computational time on a single core of an Intel i7 CPU running at 2.9GHz. This shows that for a similar accuracy, SPINN is competitive with a traditional meshless method.
}

\section{Discussion}

The primary purpose of the various examples presented earlier is to provide a proof-of-concept demonstration that the SPINN model works well for a large class of ODEs and PDEs. There are however many aspects which can be improved. We discuss some of these here; we will be investigating these systematically in forthcoming publications.

\subsection{Interpretability of SPINN}
\rr{In this section, we further clarify the precise sense in which we say that SPINN is interpretable. Once a specific kernel has been chosen, either in the form of basis functions like RBF, softplus hat, neural network kernel, etc., the SPINN approximation reduces to the ansatz \eqref{eq:pde_meshless_approx}.  The interpretability of this ansatz when choosing simple basis functions like RBF and softplus hat is evident. In the case when the kernel network is chosen as an entire neural network, we note that we do not provide a detailed interpretation of the weights and biases of the kernel network. Such an intepretation, however, is not necessary in the current setting since all that we require is that the kernel network, which is typically much smaller than a full DNN that would be employed in methods like PINN, provides a particular choice of basis $\varphi$ in \eqref{eq:pde_meshless_approx}. This is to be compared with traditional approximations, like Daubechies wavelets for instance, where the precise shape of the mother wavelet is not \emph{interpretable} in the classical sense, but the wavelet transform itself as a whole is interpretable. This is the sense in which we consider SPINN to be interpretable when using neural network kernels - given a particular choice of basis function, which is what the kernel network encodes, the specific way in which shifted and scaled versions of this basis function are linearly combined to provide the SPINN approximation provides a representation that is interpretable in exactly the same sense that traditional representations are.  The difference between our interpretation of SPINN and basis function intepretations of PINN as in \cite{CGPPT20} is that by confining all non-interpretability to a much smaller neural network kernel, we are able to \emph{peer} into the workings of the SPINN algorithms in a manner that can be done with traditional meshless methods. Indeed, as we have demonstrated with various examples, we can post-process the SPINN solution to extract nodal positions and local widths of the basis functions. This in turn permits us to highlight the precise sense in which the SPINN algorithm handles regions of sharp gradients or shocks by locally adapting a single parent basis function. As seen in ODE \eqref{eq:ode4} which has a highly oscillatory solution, there is no natural means to design the architecture of PINNs. In sharp contrast, in SPINN we choose the number of nodes in a manner that is similar to how the discretization is chosen based on a Nyquist sampling condition in traditional meshless methods. Such a detailed analysis and network architectural choices involving local information is not easily accomplished when employing methods like PINN, even when interpreted as global basis functions.}

\subsection{Boundary conditions}
For all the variants of SPINN presented in this work, the loss is defined as a weighted sum of the interior and boundary losses. The boundary loss, in particular, is enforced as a penalty term that is proportional to the square of the predicted and imposed boundary values. The constant of proportionality, however, is arbitrarily chosen and varies for each problem. This is a well known limitation of penalty based techniques to enforce constraints. While Dirichlet boundary conditions are relatively easy to enforce, capturing Neumann boundary conditions require careful choice of the fixed and free nodes. For instance, a fixed node at a Neumann boundary when using Gaussian kernels will lead to an infinite indeterminacy of the corresponding coefficient since the slope of the Gaussian kernel is zero at its center. This translates to convergence issues with the loss minimization algorithm. As a simple solution, we use fixed nodes only on the Dirichlet boundaries. We observe that the nodes at times move outside the domain to enforce the boundary condition properly. However, we point out that there are other means to enforce boundary conditions like having a fixed layer of nodes along the boundary, as is done in some particle based methods like Smoothed Particle Hydrodynamics (SPH)~\cite{ye2019sph}. Alternatively, one could \emph{build in} the boundary conditions directly into the solution, as is done in works such as \cite{BN2018}.

\subsection{Time dependent PDEs}
In the context of FD-SPINN algorithms for time dependent PDEs, we used a first order scheme for time marching. However it is in principle straightforward to implement higher order schemes to control the time discretization errors using methods such as those proposed in \cite{SCL2020pre}. We also point out that the FD-SPINN algorithms presented here are different from the corresponding finite difference schemes since the spatial derivatives are handled \emph{exactly} using automatic differentiation. Automatic differentiation has been used in the context of finite element discretization problems in both static \cite{TRRB2002} and dynamic \cite{RG2014} PDEs, but the implicit finite difference schemes we propose here provide a systematic means to develop a variety of efficient time marching schemes.

\subsection{Computational efficiency}
The current implementation of SPINN has been designed as a proof-of-concept.  While the problems considered in this work can be solved with very little computational effort, a naive implementation will not scale well for larger problems. This is because the method currently evaluates the effect of \emph{all} the nodes on a given sample point making the problem $O(NM)$ given $N$ nodes and $M$ samples.  We have investigated performing spatial convolutions using the \texttt{pytorch-geometric} package~\cite{pytorch_geometric} to accelerate this computation by restricting the interaction of sampling points only with their nearest nodes, thereby reducing the problem to $O(nM)$ where $n$ is the typical number of neighbors.  While preliminary results are encouraging and allow us to use more nodes, there are some limitations and constraints that need to be investigated further.

\rr{We note however that SPINN is more efficient than PINN as we have demonstrated in some of the results here. Many applications involving PINN that we find in the literature use DNNs with many hidden layers and $O(100)$ neurons in each layer. In contrast, we note that the size of the SPINN architecture, even when using kernel networks, is significantly smaller in comparison. A simple estimate is provided here to illustrate this difference better: for the SPINN architecture as shown in Figure~\ref{fig:meshless_nn_detailed} the number of connections is $O(N)$, where $N$ is the number of nodes in the meshless representation, while for a fully connected network of the same size the corresponding number of connections is $O(N^2)$. \footnote{The actual numbers for the case shown in Figure~\ref{fig:meshless_nn_detailed} are $2dN + (3 + 12 + 12 + 3) + N = (2d + 1)N + 30$, while the number of connections in the corresponding DNN is $(Nd^2 + dN^2 + (3N^2 + 12N^2 + 12N^2 + 3N^2 + N) = (30 + d)N^2 + (d^2 + 1)N$. Here $d$ is the physical dimension of the problem.} We remark that though PINNs need not necessarily have the same size of the corresponding SPINN architecture, as noted above, the size of typical DNNs used in PINN is much larger than what we use in SPINN.} This highlights the sparsity of the SPINN model, and hence the corresponding gain in computational efficiency. Over and above this, we note that by choosing the kernel to be compact functions, SPINN can \rb{in principle} be made as efficient as a traditional meshless method. This is a significant computational advantage.

\rb{It is difficult to perform a direct performance comparison between a traditional meshless method and SPINN as it is currently implemented.  The algorithms are very different and this makes a direct comparision not very meaningful. However, as we have done in section~\ref{subsec:ns}, the SPINN solution is more accurate than an optimized SPH solution using an order of magnitude more computational elements (225 versus 2500) and requires about 1.5 times the computational time on the same computational hardware.  This is quite a remarkable result.

As discussed earlier, our current implementation is not optimized and relies on \texttt{pytorch}.  Many open source meshless frameworks, like PySPH~\cite{pysph2020} that was used in our comparison, are highly optimized.  On the other hand, it is easy to see that the function approximation using SPINN is of the same computational order as a conventional meshless method. Based on some preliminary computations using the geometric convolutions from \texttt{pytorch-geometric}, we observe that the evaluation of the solution using the neural network for the interior and boundary points takes about 20\% of the entire computational cost.  The remainder of the time is taken by the computation of the derivatives, losses, and optimizer.  Therefore if we are able to improve the computational performance of the forward operation and the resultant derivatives, which is certainly possible, we can obtain a performance that is comparable with that of a conventional meshless method. This along with the fact that neural networks are easily implemented on GPUs makes for a compelling case for SPINN.
}

\subsection{Finite element based extensions of SPINN}
We have used meshless approximations to motivate the development of the SPINN architecture in this work. However, conforming mesh based representations like finite elements could also be used to develop the corresponding neural network generalizations, although the connection is not straightforward. The difficulty arises because of the fact that enforcing mesh conformity places more constraints on the architecture. There are theoretical results elaborating the relation between finite element models and DNNs. For instance in \cite{HLXZ2020} the authors show how every piecewise linear finite element model can be mapped to a ReLU DNN, while in \cite{OPS19} higher order finite elements have been investigated along similar lines.  The advantage in using meshless representations over conforming mesh representations based on the finite elements is that the corresponding DNN has more flexibility in how the nodes move, and how the kernel widths adapt. In addition it also allows for a variety of generalizations like the Fourier and wavelet generalizations of SPINN.

\section{Conclusion}
To conclude, we have presented a new class of sparse DNNs which we call SPINN - Sparse, Physics-based and \rR{partially} Interpretable Neural Networks - which naturally generalize classical meshless methods while retaining the advantages of DNNs. The key features  of SPINN are:
\begin{enumerate}[(i)]
\item It is interpretable (\rR{in the specific sense discussed earlier and summarized next}), in sharp contrast to DNN based methods.
\item \rR{All the non-intepretability is reduced to a much smaller kernel network in contrast to a full DNN, thereby permitting an interpretation of the trained network in a manner similar to meshless methods.}
\item It is efficient in comparison with a DNN with the same number of neurons on account of its sparsity.
\item It is physics-based since the loss function depends directly on the strong form of a PDE or its variational form.
\item It implicitly encodes mesh adaptivity as part of its training process.
\item It is able to resolve non-smoothness in solutions, as in the case of shocks.
\item It suggests new classes of numerical algorithms for solving PDEs. For instance, the Fourier-SPINN extension was illustrated in this work.
\end{enumerate}
We have demonstrated these aspects of SPINN using a variety of ODEs and PDEs, and thus present SPINN as a novel numerical method that seamlessly bridges traditional meshless methods and modern DNN-based algorithms for solving PDEs.  Recognizing this link opens fresh avenues for developing new numerical algorithms that blend the best of both worlds. Even a mere re-expression of meshless algorithms as a SPINN model makes it easier to enhance traditional meshless methods along the lines of differentiable programming. \new{We conclude by emphasizing that the strength of SPINN lies in its interpretability, its backwards-compatibility with traditional discretizations, and forward-compatibility with DNN based methods.}

\section*{Author contributions}
Both authors contributed equally to the conceptualization, formulation of the problem, developing the codes, performing the analyses and writing the manuscript.

\bibliographystyle{plain}
\bibliography{spinn}

\newpage
\section*{Appendix}
\appendix

\section{ReLU networks and piecewise linear finite element approximation} \label{app:relu_fem_1d}
In this section, we illustrate the connection between SPINN and DNN represenations for PDEs that have been been previously studied. To keep the discussion concrete we focus on the special case of one spatial dimension. Consider the problem of minimizing the functional $I:H^1_0([0,1]) \to \mathbb{R}$,
\begin{equation} \label{eq:fnl_1d}
I(u) = \frac{1}{2}\int_0^1 \left(\frac{du(x)}{dx}\right)^2 \,dx - \int_0^1 f(x)u(x) \,dx,
\end{equation}
where $f \in L^2([0,1])$. A standard argument shows that the Euler-Lagrange equations corresponding to the minimization of the functional \eqref{eq:fnl_1d} is the second order ODE:
\begin{equation} \label{eq:ode_2ndorder_1d}
\begin{split}
\frac{d^2u(x)}{dx^2} + f(x) = 0, & \quad x \in [0,1],\\
u(0) = 0, & \quad u(1) = 0.
\end{split}
\end{equation}
To illustrate this, we consider the special case of a piecewise linear finite element approximation of the solution of \eqref{eq:ode_2ndorder_1d}. The basis functions for a piecewise linear finite element approximation can be equivalently thought of as ReLU network with one hidden layer. For this case, a convenient SPINN architecture is the following ReLU network with one hidden layer:
\begin{equation} \label{eq:spinn_1d}
u(x) = \sum_{i=0}^N w_i \text{ReLU}(x - x_i),
\end{equation}
Following \cite{HLXZ2020}, we outline the relation between a piecewise linear finite element representation of a function $u:[a,b] \to \mathbb{R}$ and neural networks with ReLU activation functions. Letting $(x_i)_{i=0}^N$ be a partition of $[a,b]$, such that $x_0 = a$, $x_N = b$, and $x_i < x_{i+1}$ for $0 \le i < N$, the piecewise linear basis function $N_i(x)$, for $1 < i < N$, is given by
\begin{equation} \label{eq:hat_function_fem_1d}
N_i(x) = \begin{cases}
\frac{x_i - x}{x_i - x_{i-1}}, & x \in [x_{i-1},x_i],\\
\frac{x - x_i}{x_{i+1} - x_i}, & x \in [x_i, x_{i+1}].
\end{cases}
\end{equation}
An important observation that connects the finite element approximation of $u$ with ReLU neural networks is the fact that the basis function in \eqref{eq:hat_function_fem_1d} can be written as a linear combination of ReLU functions in the following form:
\begin{equation} \label{eq:hat_1d_relu_repr}
N_i(x) = \frac{1}{h_i}\text{ReLU}(x - x_{i-1}) - \left(\frac{1}{h_i} + \frac{1}{h_{i+1}}\right)\text{ReLU}(x - x_i) + \frac{1}{h_{i+1}}\text{ReLU}(x - x_{i+1}),
\end{equation}
where we use the symbols $h_k = x_{k} - x_{k-1}$, $1 < k < N$, to denote the lengths of the various elements in a given partition of $[a,b]$. Using \eqref{eq:hat_function_fem_1d} and \eqref{eq:hat_1d_relu_repr}, we can write the piecewise linear finite element approximation of $u$, namely
\begin{equation} \label{eq:fem_approx_1d}
u(x) = \sum_{i=0}^N N_i(x) U_i,
\end{equation}
as
\begin{equation} \label{eq:fem_1d_ReLU}
u(x) = \sum_{i=0}^{N - 1} (\theta_{i+1} - \theta_i) \text{ReLU}(x - x_i),
\end{equation}
where $\theta_i = (U_i - U_{i-1})/h_i$. The representation \eqref{eq:fem_1d_ReLU} informs us that a piecewise linear finite element approximation of 1d functions is equivalent to a DNN with one hidden layer with weights and biases consistent with \eqref{eq:fem_1d_ReLU}.

In \cite{HLXZ2020}, the authors proposed a hybrid method wherein they use the representation \eqref{eq:fem_1d_ReLU}, with the weights $(\theta_i)$ computed using the standard finite element method holding the mesh fixed, and the biases $(x_i)$ are computed by minimizing the loss $I$ as a function of the biases $(x_i)$ for fixed values of the weights $(\theta_i)$. This staggered approach, however, does not take full advantage of the variety of stochastic gradient algorithms that have been developed for DNNs. In contrast, the SPINN architecture which we propose in this work does not use a staggered approach, and is more efficient.

Even in this simple case, we note the following features: (i) the weights connecting the input layer, which just takes in $x$, and the hidden layer is $1$ uniformly, (ii) the biases of the hidden layer are directly interpretable as the position of the nodes of the finite element discretization, and (iii) the weights connecting the hidden layer and the output layer are interpretable in terms of the nodal values of the corresponding finite element solution. We also see that the number of neurons in the hidden layer is just the number of interior nodes in the finite element discretization. This is in sharp contrast to the approach followed in Physics Informed Neural Networks (PINN) \cite{RPK2019}, or the Deep-Ritz method \cite{EYu2018}, which employ dense neural networks, and hence are not easily interpretable.

\section{Code design}

\begin{sloppypar}
The source code for SPINN is freely available at \url{https://github.com/nn4pde/SPINN}.  We use the Python programming language, the PyTorch~\cite{pytorch} library, and NumPy~\cite{numpy}.  In addition the code employs the following libraries for visualization and plotting, \verb|matplotlib|~\cite{mpl} is used for the simpler plots and Mayavi~\cite{mayavi} is used for more complex three-dimensional plots.  We use the \verb|pytorch-geometric|~\cite{pytorch_geometric} package to demonstrate the use of geometric convolutions as a means to accelerate the performance, however this is an optional requirement.  Finally, every plot shown in this manuscript is completely automated using the \verb|automan|~\cite{automan:2018} package.
\end{sloppypar}

Our code follows a simple object-oriented design employing the following objects:
\begin{enumerate}[(i)]
\item The \verb|PDE| base class and its sub-classes provide the common methods that one would need to override to define a particular set of ordinary/partial differential equations to solve along with their boundary conditions.
\item Subclasses of \verb|torch.nn.Module| manage the SPINN models.
\item The \verb|Optimizer| class manages the loss optimization.
\item The \verb|Plotter| base class provides routines for live plotting and storing the results.
\item The \verb|App| base class manages the creation and execution of the objects mentioned above to solve a particular problem.
\end{enumerate}

Each problem demonstrated has a separate script which can be executed standalone along with the ability to show the results live. The code is written in a manner such that every important parameter can be configured through the command line.  This is used in our automation script, \verb|automate.py|, which uses the \verb|automan| framework to automate the creation of every figure we present in this work.  We provide the name of the specific scripts for the different differential equations in the sections that follow; these can be found in the \verb|code| sub-directory of the repository. Further, the individual parameters used to obtain the various plots can be found by perusing \verb|automate.py|.  It bears emphasis that every single result presented here is fully reproducible and can be generated by running a single command.

\end{document}